\documentclass{ieeeaccess}
\usepackage{cite}
\usepackage{amsmath,amssymb,amsfonts}
\usepackage{algorithmic}
\usepackage{graphicx}
\usepackage{textcomp}

\usepackage{bm}
\makeatletter
\AtBeginDocument{\DeclareMathVersion{bold}
\SetSymbolFont{operators}{bold}{T1}{times}{b}{n}
\SetSymbolFont{NewLetters}{bold}{T1}{times}{b}{it}
\SetMathAlphabet{\mathrm}{bold}{T1}{times}{b}{n}
\SetMathAlphabet{\mathit}{bold}{T1}{times}{b}{it}
\SetMathAlphabet{\mathbf}{bold}{T1}{times}{b}{n}
\SetMathAlphabet{\mathtt}{bold}{OT1}{pcr}{b}{n}
\SetSymbolFont{symbols}{bold}{OMS}{cmsy}{b}{n}
\renewcommand\boldmath{\@nomath\boldmath\mathversion{bold}}}
\makeatother

\def\BibTeX{{\rm B\kern-.05em{\sc i\kern-.025em b}\kern-.08em
    T\kern-.1667em\lower.7ex\hbox{E}\kern-.125emX}}

\usepackage{mathtools}
\usepackage{bm}
\usepackage{multirow}
\usepackage{makecell}
\usepackage[subrefformat=parens]{subcaption}
\usepackage{booktabs}
\usepackage[ruled,vlined]{algorithm2e}
\usepackage[table]{xcolor}

\begin{document}
\doi{10.1109/ACCESS.2026.3713604}

\title{
Design and Experimental Validation of Sensorless 4-Channel Bilateral Teleoperation for Low-Cost Manipulators
}
\author{
\uppercase{Koki Yamane}\authorrefmark{1},
\IEEEmembership{Student Member, IEEE},
\uppercase{Yunhan Li}\authorrefmark{1},
\uppercase{Masashi Konosu}\authorrefmark{1},
\uppercase{Koki Inami}\authorrefmark{1},
\IEEEmembership{Student Member, IEEE},
\uppercase{Junji Oaki}\authorrefmark{2},
\IEEEmembership{Member, IEEE},
\uppercase{Toshiaki Tsuji}\authorrefmark{3},
\IEEEmembership{Senior Member, IEEE},
and \uppercase{Sho Sakaino}\authorrefmark{2},
\IEEEmembership{Member, IEEE},
}

\address[1]{Degree Programs in Intelligent and Mechanical Interaction Systems,
University of Tsukuba, 1-1-1 Tennodai, Tsukuba, Ibaraki 305-8573, Japan}
\address[2]{Faculty of Engineering, Information and Systems,
University of Tsukuba, 1-1-1 Tennodai, Tsukuba, Ibaraki 305-8573, Japan}
\address[3]{Department of Electrical Engineering, Electronics, and Applied Physics,
Saitama University, 255 Shimo-Ookubo, Sakura-ku, Saitama 338-8570, Japan}
\tfootnote{
This work was supported by JSPS KAKENHI Grant Number 24K00905, 24KJ0503, JST PRESTO Grant Number JPMJPR24T3, and JST ALCA-Next Japan, Grant Number JPMJAN24F1.
This study was based on the results obtained from the JPNP20004 project subsidized by the New Energy and Industrial Technology Development Organization (NEDO).\\
This work has been published in IEEE Access, vol. 14, pp. 109407–109428, 2026, doi: 10.1109/ACCESS.2026.3713604. A correction to the published article is available at doi: 10.1109/ACCESS.2026.3739427.
}

\corresp{Corresponding author: Koki Yamane (e-mail: yamane.koki.td@alumni.tsukuba.ac.jp).}

\begin{abstract}
Teleoperation of low-cost manipulators is attracting increasing attention as a practical means of collecting demonstration data for imitation learning. 
However, most existing low-cost systems rely on unilateral position control without force feedback, while implementing force-feedback bilateral teleoperation is difficult because low-cost manipulators typically have low-resolution encoders and no joint torque sensors. 
This paper presents a sensorless 4-channel bilateral teleoperation framework that integrates identified nonlinear dynamics compensation with a disturbance-observer-based velocity and external-force estimation scheme. 
By interpreting the observer structure in the frequency domain, we clarify the coupling between the velocity- and external-force-estimation bandwidths and derive practical tuning guidelines based on the damping ratio and a single cutoff frequency. 
Real-robot experiments, including force-sensor comparison and teleoperation tasks, demonstrate that the proposed framework provides practically useful force estimates and enables stable teleoperation in high-speed and contact-rich scenarios under low-cost hardware constraints. 
As an application, imitation-learning experiments demonstrate that incorporating estimated force information into demonstrations improves task success rates in the tested contact-rich manipulation tasks.
\end{abstract}

\titlepgskip=-21pt

\maketitle

\section{Introduction}
\label{sec:introduction}

\PARstart{I}{n}
recent years, influenced by the remarkable achievements of Deep Learning in fields such as computer vision and natural language processing, research into the application of Deep Learning to robot motion generation has attracted attention.
Among these approaches, imitation learning (IL), also known as learning from demonstration (LfD)~\cite{schaal1999imitation,calinon2009robot}, which uses human demonstrations as supervisory signals, is attracting significant interest because it enables high-sample-efficiency learning of robot motion, which is costly to collect on actual robots. 
More recently, large-scale neural network models that have been trained using imitation learning with extremely large amounts of human motion data~\cite{black2026pi0visionlanguageactionflowmodel,gr00tn1_2025} are called ``Vision-Language-Action (VLA) models'' or ``Large Behavior Models (LBM)'', drawing an analogy to large pre-trained models in natural language processing.

Along this trend, there is an increasing number of low-cost hardware devices used to collect teaching data for robot learning, such as the ALOHA series~\cite{zhao2023learning} and GELLO~\cite{wu2024gello}.
This trend is primarily driven by the need to deploy a large number of robots to acquire massive datasets while keeping the cost per robot low.
In addition, policies learned through imitation learning may occasionally generate undesirable behaviors, and using compact, low-cost hardware helps mitigate the risk of damage or safety hazards.

Most of these low-cost teleoperation systems use unilateral control, which only transmits target position values from the leader robot to the follower robot~\cite{zhao2023learning,wu2024gello}.
While this method is easy to implement and provides good operability for low-contact operations, it lacks force feedback and has difficulty with contact-rich tasks.
Recently, bilateral teleoperation has been increasingly adopted, enabling a two-way exchange of information between leader and follower robots as well as force feedback.
There are various types of bilateral teleoperation.
Among them, position--force architectures, such as force-reflection type bilateral teleoperation, are often used, in which external forces from followers are generated in a feedforward manner on the leader side~\cite{michel2024passivity,liu2025factr}.
However, these structures cannot track position and force simultaneously because the generated force depends on the positional error~\cite{lawrence1993stability,yokokohji1994bilateral}.
Recent teleoperation studies have investigated structured control frameworks for contact-rich manipulation, including unified contact modeling with hybrid motion/force control in unknown environments~\cite{huang2024unified} and shared control with constrained optimization and guiding force feedback for underwater electric manipulators~\cite{huang2024shared}.
In contrast, this study focuses on low-cost force-sensorless manipulators, for which realizing 4-channel bilateral teleoperation requires simultaneous estimation of joint velocity and external force from low-resolution encoder measurements and input torque.

4-channel bilateral teleoperation is a method in which both the leader and the follower transmit their positions and forces to each other and perform position and force control in parallel.
Theoretically, a 4-channel architecture can achieve the ideal response of synchronization of position and force perfectly~\cite {lawrence1993stability,yokokohji1994bilateral}.
However, to simultaneously perform position and force control with opposing objectives in a single robot, ensuring non-interference requires accurate dynamics models and observation of position, velocity, and external forces.
Because the dynamics of manipulators with six or more degrees of freedom involve many parameters and extremely complex functions, accurately estimating the parameters and performing real-time calculations have long been impractical.

As a practical implementation of 4-channel bilateral teleoperation, an accelerated control system was proposed that combines a simplified dynamics model for each joint with disturbance estimation (including external forces and model errors) using a disturbance observer (DOB)~\cite{ohnishi1996motion,shimada2023disturbance} and compensation based on the estimated disturbance.
This enables high performance even when a certain amount of model error exists~\cite{iida2004reproducibility,tsuji2006controller,sakaino2011multi-DOF,suzuki2018development,yilmaz2023sensorless}.
This acceleration control-based 4-channel bilateral teleoperation system has also been studied for application to imitation learning.
Sakaino~\textit{et al.} trained a Long-Short-Term Memory (LSTM)~\cite{hochreiter1997long} using data collected with four-channel bilateral teleoperation and conducted experiments to perform contact-rich tasks~\cite{adachi2018imitation,hayashi2022independently,sakaino2022imitation}.
Kobayashi~\textit{et al.} published the Alpha- $\alpha$ and Bi-ACT series~\cite{buamanee2024bi}, which is a combination of a low-cost 4-channel bilateral teleoperation system, Alpha-$\alpha$, and the method for training Action Chunking with Transformer (ACT)~\cite{zhao2023learning} using data collected by 4-channel bilateral teleoperation.

In practice, when using low-cost hardware, factors such as slow control cycles and insufficient resolution of rotary encoders prevent sufficiently fast compensation for model errors, thereby limiting the effectiveness of per-joint linear approximation models.
Furthermore, although modeling errors arising from such linear approximations have been effectively mitigated by using actuators with high gear reduction ratios, this approach becomes less effective for direct-drive or pseudo-direct-drive motors, which have been increasingly adopted in recent years.

With recent advances in identification algorithms, accurate dynamic parameter identification with reduced reliance on high-resolution rotary encoders, such as the direct and inverse dynamic identification model (DIDIM)~\cite{gautier2012new,leboutet2021inertial} or closed-loop input error (CLIE)~\cite{perrusquia2022stable,oaki2026reproducible}, has become practical.
At the same time, increased computational resources and improved algorithms have enabled real-time computation of multi-degree-of-freedom manipulator dynamics using software tools such as OpenSYMORO~\cite{khalil2014OpenSYMORO} and Pinocchio~\cite{carpentier2019pinocchio}.

In this study, we focused on 4-channel bilateral teleoperation combining a parameter-identified, accurate nonlinear manipulator dynamics model with a disturbance observer to achieve high tracking performance and operability with a low-cost manipulator that cannot use high feedback gains.
We compared the proposed method with other teleoperation techniques, such as unilateral control, position-symmetric control, and force-feedback control, as well as with conventional 4-channel bilateral teleoperation with simplified dynamics models or phase-lagged velocity estimation, in high-speed motions, and confirmed the improvement in control performance achieved by the proposed method.
Additionally, we confirmed that including estimated force information in demonstration data improves the task success rate of imitation learning policies without additional sensors.

The main contributions of this paper are summarized as follows:
\begin{itemize}
  \item We present a practical 4-channel bilateral teleoperation system for low-cost manipulators, integrating an identified nonlinear dynamics model with a disturbance-observer-based velocity and external force estimation.
  \item By interpreting the velocity and disturbance observer in the frequency domain, we explicitly clarify the design coupling between velocity and external force estimation bandwidths, and provide practical parameter constraints and systematic tuning guidelines that reduce the effective tuning freedom of two observer gains to a single cutoff frequency.
  \item We validate the proposed design choices through bilateral teleoperation experiments on a low-cost manipulator, demonstrating stable contact-rich demonstrations and improved imitation learning performance without force sensors.
\end{itemize}

\begin{figure*}[t]
    \centering
    \includegraphics[width=\linewidth]{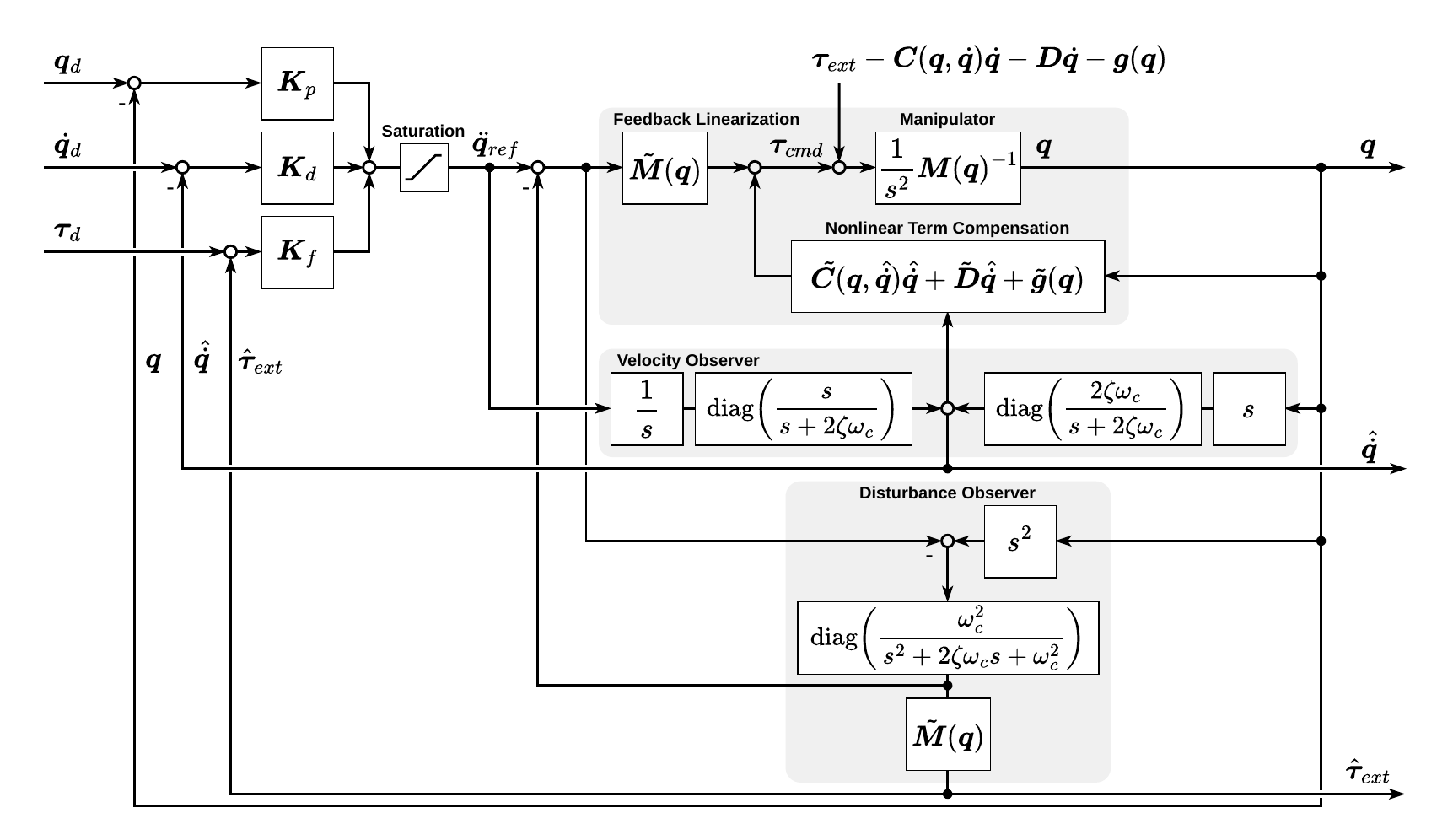}
    \caption{Block diagram of controller and observer}
    \label{fig:block_diagram}
\end{figure*}

\begin{figure}[t]
    \centering
    \includegraphics[width=\linewidth]{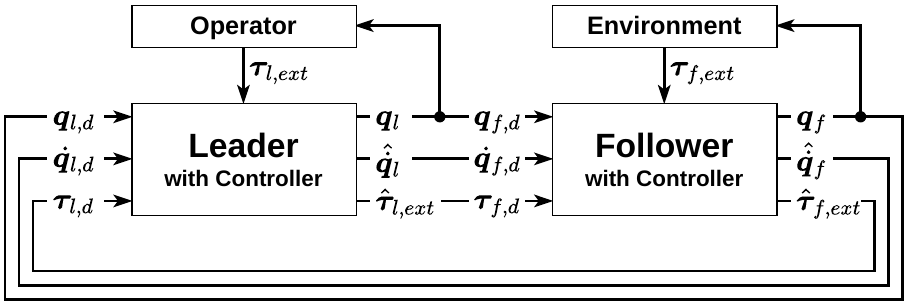}
    \caption{Block diagram of 4ch bilateral teleoperation}
    \label{fig:block_diagram_4ch}
\end{figure}

\section{Methodology}

\subsection{4-channel bilateral teleoperation}
We describe a 4-channel bilateral teleoperation system that uses a rigid serial link model.
The block diagram of each manipulator is shown in Fig.~\ref{fig:block_diagram} and the block diagram of 4-channel bilateral teleoperation is shown in Fig.~\ref{fig:block_diagram_4ch}.
Although many studies on 4-channel bilateral teleoperation have been published in the past, the major feature of the control system used in this study is that it uses parameter identification of the manipulator's rigid-body model's nonlinear dynamics and incorporates compensation for the nonlinear term, along with state estimation based on this.

\subsubsection{Manipulator Dynamics}
In general, assuming that each link is a rigid body, the dynamics of the manipulator can be expressed as a rigid serial link model.
\begin{align}
\bm{M}(\bm{q})\ddot{\bm{q}} + \bm{C}(\bm{q}, \dot{\bm{q}})\dot{\bm{q}} + \bm{D}\dot{\bm{q}} + \bm{g}(\bm{q}) = \bm{\tau}_{cmd} + \bm{\tau}_{ext}
\end{align}
where
$\bm{M}(\bm{q})$ is the inertial matrix,
$\bm{C}(\bm{q}, \dot{\bm{q}})\dot{\bm{q}}$ is the centrifugal and Coriolis forces,
$\bm{D}$ is the viscous friction coefficient,
$\bm{g}(\bm{q})$ is the gravitational forces
and $\bm{\tau}_{cmd}, \bm{\tau}_{ext}$ are the command and external torques of each joint, respectively.
For simplicity, the input torque after compensation is defined as $\bm{\tau}_{u}$ as follows:
\begin{align}
 \bm{\tau}_{u}
&\coloneqq
\bm{\tau}_{cmd}
 - \tilde{\bm{C}}(\bm{q}, \hat{\dot{\bm{q}}}) \hat{\dot{\bm{q}}} - \tilde{\bm{D}} \hat{\dot{\bm{q}}} - \tilde{\bm{g}}(\bm{q}) \label{eq:compensation} \\
\bm{\tau}_{u} &= \bm{M}(\bm{q})\ddot{\bm{q}} - \bm{\tau}_{ext} - \Delta_{comp}\label{eq:tau_u}
\end{align}
where
$\Delta_{comp}$ is the error of the nonlinear term compensation, $\tilde{\bigcirc}$ denotes parameter-identified manipulator dynamics functions, and $\hat{\bigcirc}$ denotes values estimated by an observer, which will be explained later.

\subsubsection{Transparency}

In bilateral teleoperation, a desirable control objective is to transmit the operator motion to the follower while displaying the environment-side interaction to the operator.
This property is often discussed in terms of transparency, i.e., how accurately the environment impedance is displayed to the operator~\cite{lawrence1993stability}.
Following this impedance-display viewpoint, we relate transparency to the leader--follower impedance of the control system and formulate an impedance-shaping objective for 4-channel bilateral teleoperation.

The control-system impedance expressed in the leader--follower coordinates is written in the general coupled form as
\begin{equation}
\begin{bmatrix}
\bm\tau_l \\
\bm\tau_f
\end{bmatrix}_{ext}
=
\underbrace{
\begin{bmatrix}
\bm Z_{ll} & \bm Z_{lf} \\
\bm Z_{fl} & \bm Z_{ff}
\end{bmatrix}
}_{\bm Z_{\mathrm{LF}}}
\begin{bmatrix}
\dot{\bm q}_l \\
\dot{\bm q}_f
\end{bmatrix}
.
\label{eq:lf_control_impedance}
\end{equation}
Here, $\bm{q}_l$ and $\bm{q}_f$ denote the joint-angle vectors of the leader and follower manipulators, respectively, and $\bm{\tau}_{l,ext}$ and $\bm{\tau}_{f,ext}$ denote the corresponding external joint torque vectors.
Here and below, the argument $(s)$ is omitted from the impedance matrices for compactness.

When the follower side is connected to an environment impedance $\bm Z_{env}(s)$, the boundary condition is
\begin{equation}
-\bm\tau_{f,ext}(s)
=
\bm Z_{env}(s)\dot{\bm q}_f(s).
\label{eq:follower_environment_impedance}
\end{equation}
Adding the environment-impedance term at the follower port,
$[\bm 0^{\top},(\bm Z_{env}\dot{\bm q}_f)^{\top}]^{\top}$,
to both sides of \eqref{eq:lf_control_impedance} gives
\begin{equation}
\begin{bmatrix}
\bm\tau_{l,ext} \\
\bm\tau_{f,ext}+\bm Z_{env}\dot{\bm q}_f
\end{bmatrix}
=
\left(
\bm Z_{\mathrm{LF}}
+
\begin{bmatrix}
\bm 0 & \bm 0 \\
\bm 0 & \bm Z_{env}
\end{bmatrix}
\right)
\begin{bmatrix}
\dot{\bm q}_l \\
\dot{\bm q}_f
\end{bmatrix}.
\label{eq:environment_added_lf_system}
\end{equation}
Using the boundary condition in \eqref{eq:follower_environment_impedance}, the environment-terminated system becomes
\begin{equation}
\begin{bmatrix}
\bm\tau_{l,ext} \\
\bm 0
\end{bmatrix}
=
\begin{bmatrix}
\bm Z_{ll} & \bm Z_{lf} \\
\bm Z_{fl} & \bm Z_{ff}+\bm Z_{env}
\end{bmatrix}
\begin{bmatrix}
\dot{\bm q}_l \\
\dot{\bm q}_f
\end{bmatrix}.
\label{eq:environment_terminated_lf_system}
\end{equation}
The first row of \eqref{eq:environment_terminated_lf_system} represents the torque displayed at the leader side, whereas the second row represents the follower-side environment termination.

Eliminating $\dot{\bm q}_f$ from the second row gives
\begin{equation}
\dot{\bm q}_f
=
-
\left(
\bm Z_{ff}
+
\bm Z_{env}
\right)^{-1}
\bm Z_{fl}
\dot{\bm q}_l.
\label{eq:qf_elimination_lf}
\end{equation}
Substituting \eqref{eq:qf_elimination_lf} into the first row gives Schur-complement form as follows:
\begin{equation}
\bm\tau_{l,ext}
=
\underbrace{
\left[
\bm Z_{ll}
-
\bm Z_{lf}
\left(
\bm Z_{ff}
+
\bm Z_{env}
\right)^{-1}
\bm Z_{fl}
\right]
}_{\bm Z_{l\gets f}}
\dot{\bm q}_l.
\label{eq:leader_torque_schur_lf}
\end{equation}
The ideal environment-to-operator impedance display is
\begin{equation}
\bm Z_{l\gets f}
=
\bm Z_{env}.
\label{eq:ideal_environment_impedance_display}
\end{equation}
Therefore, the displayed-impedance error can be written as
\begin{equation}
\bm Z_{l\gets f}
=
\bm Z_{env}
+
\Delta \bm Z_{l\gets f},
\label{eq:leader_displayed_impedance_error_decomposition}
\end{equation}
where
\begin{equation}
\Delta \bm Z_{l\gets f}
=
-
\bm Z_{env}
+
\bm Z_{ll}
-
\bm Z_{lf}
\left(
\bm Z_{ff}
+
\bm Z_{env}
\right)^{-1}
\bm Z_{fl}.
\label{eq:leader_displayed_impedance_error_general}
\end{equation}
Equation~\eqref{eq:leader_displayed_impedance_error_general} shows that the control objective can be interpreted as shaping the leader--follower impedance matrix $\bm Z_{\mathrm{LF}}$ so that $\Delta \bm Z_{l\gets f}$ becomes small for the connected environment impedance.
The same argument can be applied to the opposite direction by connecting an operator-side impedance to the leader port and evaluating the equivalent impedance displayed at the follower side.

\subsubsection{Coordinate Transformation}

The general expression in \eqref{eq:leader_displayed_impedance_error_general} shows the desired impedance relation.
The remaining question is how the impedance matrix $\bm Z_{\mathrm{LF}}$ should be shaped to reduce the displayed impedance error.
To make this impedance-design objective explicit, we introduce relative-motion and common-motion coordinates:
\begin{equation}
\begin{bmatrix}
\dot{\bm q}_- \\
\dot{\bm q}_+
\end{bmatrix}
=
\bm J_{\pm}
\begin{bmatrix}
\dot{\bm q}_l \\
\dot{\bm q}_f
\end{bmatrix},
\qquad
\bm J_{\pm}
:=
\begin{bmatrix}
\bm I & -\bm I \\
\frac{1}{2}\bm I & \frac{1}{2}\bm I
\end{bmatrix}
.
\label{eq:velocity_coordinate_transformation}
\end{equation}
Here, $\dot{\bm q}_-$ and $\dot{\bm q}_+$ denote the relative and common velocities, respectively.
The corresponding external torque transformation is determined from power invariance as
\begin{equation}
\begin{bmatrix}
\bm\tau_- \\
\bm\tau_+
\end{bmatrix}_{ext}
=
\bm J_{\pm}^{-\top}
\begin{bmatrix}
\bm\tau_l \\
\bm\tau_f
\end{bmatrix}_{ext}
=
\begin{bmatrix}
\frac{1}{2}\bm I & -\frac{1}{2}\bm I \\
\bm I & \bm I \\
\end{bmatrix}
\begin{bmatrix}
\bm\tau_l \\
\bm\tau_f
\end{bmatrix}_{ext}
.
\label{eq:torque_coordinate_transformation}
\end{equation}

In the transformed coordinates, the control-system impedance is written as
\begin{equation}
\begin{bmatrix}
\bm\tau_- \\
\bm\tau_+
\end{bmatrix}_{ext}
=
\underbrace{
\begin{bmatrix}
\bm Z_{--} & \bm Z_{-+} \\
\bm Z_{+-} & \bm Z_{++}
\end{bmatrix}
}_{\bm Z_{\pm}}
\begin{bmatrix}
\dot{\bm q}_- \\
\dot{\bm q}_+
\end{bmatrix}.
\label{eq:transformed_control_impedance}
\end{equation}
Using \eqref{eq:velocity_coordinate_transformation} and \eqref{eq:torque_coordinate_transformation}, the impedance matrix in the leader--follower coordinates is obtained from the modal impedance matrix as
\begin{equation}
\bm Z_{\mathrm{LF}}
=
\bm J_{\pm}^{\top}
\bm Z_{\pm}
\bm J_{\pm}
.
\label{eq:lf_impedance_from_pm}
\end{equation}

To make the effect of each transformed impedance element explicit, consider the scalar single-axis case.
Substituting the scalar form of \eqref{eq:lf_impedance_from_pm} into the Schur-complement expression in \eqref{eq:leader_displayed_impedance_error_general} gives
\begin{equation}
\Delta Z_{l\gets f}
=
\frac{N_{\Delta}}{D_{\Delta}},
\label{eq:leader_displayed_impedance_general_modal_error}
\end{equation}
with
\begin{equation}
\begin{aligned}
N_{\Delta}
&=
Z_{--}Z_{++}
-
Z_{env}^2
+
\underbrace{
Z_{env}\left(
Z_{-+}+Z_{+-}
\right)
-
Z_{-+}Z_{+-}
}_{\text{modal coupling}}
,
\\
D_{\Delta}
&=
Z_{--}
+
\frac{1}{4}Z_{++}
+
Z_{env}
-
\underbrace{
\frac{1}{2}
\left(
Z_{-+}+Z_{+-}
\right)
}_{\text{modal coupling}}
.
\end{aligned}
\label{eq:leader_displayed_impedance_general_modal_error_terms}
\end{equation}
Equation~\eqref{eq:leader_displayed_impedance_general_modal_error_terms} shows that the off-diagonal modal impedances affect the displayed-impedance error.
A large relative-mode impedance $Z_{--}$ attenuates the influence of these terms because it becomes the dominant term. However, this attenuation does not imply that the off-diagonal terms can be ignored.
In practical systems, $Z_{--}$ is finite, especially in the high-frequency range.
Moreover, the residual coupling term depends on the environment impedance $Z_{env}$, and thus its influence can become significant in contact with a stiff environment.
Therefore, the effect of the off-diagonal modal impedances is not negligible in general, and suppressing $Z_{-+}$ and $Z_{+-}$ is desirable for accurate impedance display.
This modal-decoupling condition is expressed as
\begin{equation}
Z_{-+}(s)\simeq 0,
\qquad
Z_{+-}(s)\simeq 0.
\label{eq:modal_decoupling_condition_scalar}
\end{equation}

Under the modal-decoupling condition in \eqref{eq:modal_decoupling_condition_scalar}, \eqref{eq:leader_displayed_impedance_general_modal_error_terms} reduces to
\begin{equation}
\Delta Z_{l\gets f}
=
\frac{
Z_{--}Z_{++}
-
Z_{env}^2
}{
Z_{--}
+
Z_{env}
+
\frac{1}{4}Z_{++}
}.
\label{eq:leader_displayed_impedance_noninterference}
\end{equation}
To gain insight, when the relative-mode impedance dominates the denominator, i.e., \(|Z_{--}| \gg |Z_{env} + Z_{++}/4|\), the displayed-impedance error can be approximated as
\begin{equation}
\Delta Z_{l\gets f}
\simeq
Z_{++}
-
\frac{Z_{env}^2}{Z_{--}}.
\label{eq:leader_displayed_impedance_approx}
\end{equation}
Consistent with this interpretation, a sufficient limiting modal-impedance objective is
\begin{equation}
\begin{aligned}
    Z_{--}(s)&\to\infty,
    &
    Z_{++}(s)&\to 0
    \quad
    \Rightarrow
    \quad
    \Delta Z_{l\gets f}(s)\to 0,
\end{aligned}
\label{eq:impedance_display_limit}
\end{equation}
or equivalently,
\begin{equation}
Z_{l\gets f}(s)\to Z_{env}(s).
\end{equation}
Thus, high impedance in the relative-motion mode and low impedance in the common-motion mode directly improve the environment-to-operator impedance display.
In the limiting case, this impedance design yields the ideal response of bilateral teleoperation,
\begin{equation}
    \dot{\bm q}_-=\dot{\bm q}_l-\dot{\bm q}_f=\bm 0,
    \qquad
    \bm\tau_{+,ext}=\bm\tau_{l,ext}+\bm\tau_{f,ext}=\bm 0.
    \label{eq:ideal_unscaled_response}
\end{equation}
These relations are consequences of the ideal impedance design, not the definition of the control objective itself.

This modal interpretation also clarifies the role of 4-channel bilateral control.
Because position and force information are exchanged in both directions, 4-channel control can shape the relative-motion and common-motion impedances while suppressing their cross-coupling.
In contrast, bilateral controllers with fewer information channels generally impose structural constraints on $\bm Z_{\pm}$, making it difficult to simultaneously suppress $Z_{-+}$ and $Z_{+-}$ while assigning high impedance to the relative-motion mode and low impedance to the common-motion mode.

\subsubsection{Controller}
The relative-motion mode should have high impedance to suppress the position difference, which motivates assigning large virtual inertia, damping, and stiffness.
In contrast, the common-motion mode should have low impedance, and its desired impedance should therefore be minimized by removing virtual stiffness and assigning small virtual inertia and damping within practical stability and noise constraints.
Accordingly, the controller gains are designed as the parameters of desired mass--damper--spring systems in the transformed coordinates.
This corresponds to impedance control~\cite{hogan1984impedance} applied after the modal transformation~\cite{tsuji2006controller,kutsuzawa2025_acceleration_bilateral_control}, as follows:
\begin{equation}
\begin{aligned}
\begin{bmatrix}
\bm{\tau}_- \\
\bm{\tau}_+ \\
\end{bmatrix}_{ext}
&=
\begin{bmatrix}
\bm{M}_{-} & \bm{0} \\
\bm{0} & \bm{M}_{+} \\
\end{bmatrix}
\begin{bmatrix}
\ddot{\bm{q}}_- \\
\ddot{\bm{q}}_+ \\
\end{bmatrix}
+
\begin{bmatrix}
\bm{D}_{-} & \bm{0} \\
\bm{0} & \bm{0} \\
\end{bmatrix}
\begin{bmatrix}
\dot{\bm{q}}_- \\
\dot{\bm{q}}_+ \\
\end{bmatrix}
\\ &\qquad 
+
\begin{bmatrix}
\bm{K}_{-} & \bm{0} \\
\bm{0} & \bm{0} \\
\end{bmatrix}
\begin{bmatrix}
\bm{q}_- \\
\bm{q}_+ \\
\end{bmatrix}
.
\end{aligned}
\end{equation}
Rearranging the desired dynamics with respect to the transformed acceleration gives
\begin{equation}
\begin{aligned}
\begin{bmatrix}
\ddot{\bm{q}}_- \\
\ddot{\bm{q}}_+ \\
\end{bmatrix}
&=
\begin{bmatrix}
\bm{M}_{-} & \bm{0} \\
\bm{0} & \bm{M}_{+} \\
\end{bmatrix}^{-1}
\Bigg(
-
\begin{bmatrix}
\bm{D}_{-} & \bm{0} \\
\bm{0} & \bm{0} \\
\end{bmatrix}
\begin{bmatrix}
\dot{\bm{q}}_- \\
\dot{\bm{q}}_+ \\
\end{bmatrix}
\\ &\qquad 
-
\begin{bmatrix}
\bm{K}_{-} & \bm{0} \\
\bm{0} & \bm{0} \\
\end{bmatrix}
\begin{bmatrix}
\bm{q}_- \\
\bm{q}_+ \\
\end{bmatrix}
+
\begin{bmatrix}
\frac{1}{2} \bm{I} & -\frac{1}{2} \bm{I} \\
\bm{I} & \bm{I} \\
\end{bmatrix}
\begin{bmatrix}
\bm{\tau}_l \\
\bm{\tau}_f \\
\end{bmatrix}_{ext}
\Bigg)
.
\label{eq:desired_impedance}
\end{aligned}
\end{equation}
This expression can be decomposed into proportional--derivative (PD) control of the relative-motion mode and force-driven motion of the common-motion mode, leading to the following acceleration reference:
\begin{align}
\begin{bmatrix}
\ddot{\bm{q}}_- \\
\ddot{\bm{q}}_+ \\
\end{bmatrix}_{ref}
&=
\begin{bmatrix}
- 2\bm{K}_{p}\bm{q}_- - 2\bm{K}_{d}\hat{\dot{\bm{q}}}_- \\
\bm{K}_{f}\hat{\bm{\tau}}_+ \\
\end{bmatrix}
\label{eq:acc_ref_-+}
\end{align}
where $\bm{K}_{p}$, $\bm{K}_{d}$, $\bm{K}_{f}$ are the controller gains, and \(\hat{\bm\tau}_{+}:=\hat{\bm\tau}_{l,ext}+\hat{\bm\tau}_{f,ext}\).
The coefficient $2$ in the position control is because this acceleration reference $\ddot{\bm{q}}_{-ref}$ is expressed in terms of the difference in joint angles, and the difference moves twice as fast as the joint of each robot because both the leader and follower are moving.
Comparing the equation~\eqref{eq:desired_impedance} and \eqref{eq:acc_ref_-+}, the relationships between the controller gains and the desired dynamics parameters are as follows: $\bm{K}_{p}=\frac{1}{2}\bm{M}_{-}^{-1}\bm{K}_{-}$, $\bm{K}_{d}=\frac{1}{2}\bm{M}_{-}^{-1}\bm{D}_{-}$, and $\bm{K}_{f}=\bm{M}_{+}^{-1}$.

Since the coordinate transformation matrix $\bm{J}_{\pm}$ is time-invariant, the coordinate transformation of the angular acceleration can be described by the same matrix.
By using the inverse matrix of the coordinate transformation matrix $\bm{J}_{\pm}$, the acceleration reference for each manipulator can be expressed as follows:
\begin{align}
\begin{bmatrix}
\ddot{\bm{q}}_l \\
\ddot{\bm{q}}_f \\
\end{bmatrix}_{ref}
&=
\bm{J}_{\pm}^{-1}
\begin{bmatrix}
\ddot{\bm{q}}_- \\
\ddot{\bm{q}}_+ \\
\end{bmatrix}_{ref}
=
\begin{bmatrix}
\frac{1}{2}\bm{I} & \bm{I} \\
-\frac{1}{2}\bm{I} & \bm{I} \\
\end{bmatrix}
\begin{bmatrix}
\ddot{\bm{q}}_- \\
\ddot{\bm{q}}_+ \\
\end{bmatrix}_{ref}.
\label{eq:acc_ref_lf}
\end{align}
Finally, to reproduce the acceleration reference using the identified inertia matrix $\tilde{\bm{M}}(\bm{q})$ and estimated external force $\hat{\bm{\tau}}_{ext}$, the input torque $\bm{\tau}_{u}$ is set as follows:
\begin{align}
\bm{\tau}_{u} &= \tilde{\bm{M}}(\bm{q})\ddot{\bm{q}}_{ref} - \hat{\bm{\tau}}_{ext}\label{eq:tau_u2}.
\end{align}
Substituting (\ref{eq:acc_ref_lf}) in equation (\ref{eq:tau_u2}), and adding compensation according to equation (\ref{eq:compensation}), the control law for each manipulator is set as follows:
\begin{align}
\bm{\tau}_{cmd}  &=
\tilde{\bm{M}}(\bm{q})
\left \{
\bm{K}_{p}(\bm{q}_{d} - \bm{q})
+ \bm{K}_{d}(\dot{\bm{q}}_{d} -\hat{\dot{\bm{q}}})
+ \bm{K}_{f}(\bm{\tau}_{d} + \hat{\bm{\tau}}_{ext})
\right \}
\notag \\
&\qquad - \hat{\bm{\tau}}_{ext}
 + \tilde{\bm{C}}(\bm{q}, \hat{\dot{\bm{q}}}) \hat{\dot{\bm{q}}} + \tilde{\bm{D}} \hat{\dot{\bm{q}}} + \tilde{\bm{g}}(\bm{q})
\end{align}
where the desired states $\bm{q}_{d}, \dot{\bm{q}}_{d}, \bm{\tau}_{d}$ are opposite side states between leader and follower.

Note that the position control without force feedback in equation~\eqref{eq:acc_ref_-+} can be interpreted as assigning infinite inertia to the desired dynamics.
In an ideal system with infinite inertia, the position error would not change even in the absence of spring and damper terms. In practice, however, the achievable bandwidth of the desired inertia is limited, and inertia alone cannot eliminate existing position errors.
Therefore, the position control gains $\bm{K}_{p}, \bm{K}_{d}$, which correspond to the stiffness-to-inertia and damping-to-inertia ratios, are introduced to suppress drift caused by high-frequency external disturbances.
The force gain $\bm{K}_{f}$ represents the inverse inertia of the averaged joint motion, and a higher gain provides a lighter perceived impedance for the operator. In practice, the force gain is limited by phase lag and noise in the force observation; therefore, choosing a value corresponding to the same or a larger inertia than the actual system's inertia is considered a safe tuning strategy.

\begin{figure*}[t]
    \centering
    \includegraphics[width=\linewidth]{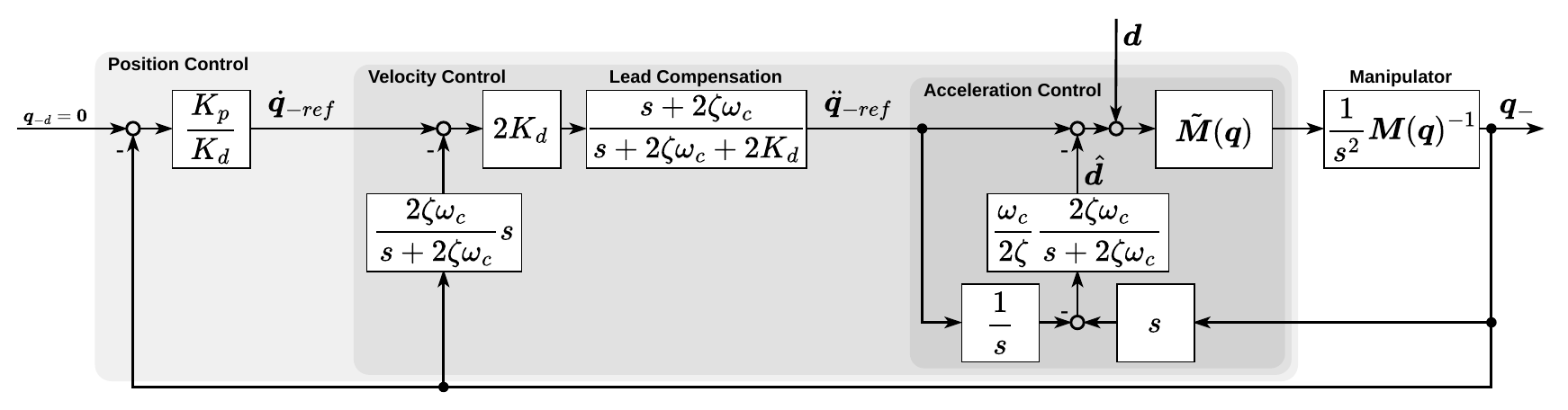}
    \caption{
        \textbf{Block diagram of the position controller of joint angle difference interpreted as cascade control.} \\
        $\bm{q}_{-d}$ is the desired value for position, and $\dot{\bm{q}}_{-ref}$, $\ddot{\bm{q}}_{-ref}$ are the reference values of velocity, acceleration control of joint angle difference, respectively.
        Lead compensation is implemented through the estimated angular velocity feedback, and integral control is implemented through the estimated external torque feedback.
        Note that this block diagram only illustrates the dynamics of the joint angle difference and does not include force control on the joint angle average.
    }
    \label{fig:block_diagram_cascade}
\end{figure*}

\subsection{Velocity and External Force Estimation}
To implement the 4-channel bilateral teleoperation described above, joint angles, joint angular velocities, and external forces of the leader and follower are required.
However, the manipulator CRANE-X7 used in this study has only 12-bit rotary encoders on the output side of the reduction gear and lacks torque sensors.
Therefore, the joint angular velocities and external forces must be estimated by an observer.
Although numerous methods have been proposed for estimating joint velocity~\cite{liu2022velocity,arteaga2022local} or external forces~\cite{haddadin2017robot,garofalo2019sliding,han2021toward}, this study adopts an approach based on a disturbance observer (DOB)~\cite{ohnishi1996motion,murakami1993torque,shimada2023disturbance} or an extended state observer (ESO)~\cite{han2009pid,chen2015disturbance} to simultaneously estimate both velocity and external torque using only position measurements and input torque.
By interpreting the observer structure in the frequency domain, the proposed formulation explicitly reveals the design coupling between the bandwidths of the velocity and external force estimators, enabling practical parameter constraints and systematic tuning of the two observer gains via a single cutoff frequency.

\subsubsection{State-Space Model with Disturbance}
Following the convention for disturbance observers~\cite{shimada2023disturbance}, a state-space model of the manipulator dynamics that includes the disturbance as a state can be formulated.
First, the acceleration-domain disturbance term $\bm{d}$ is defined as
\begin{align}
\bm{d} \coloneqq \bm{M}(\bm{q})^{-1}\left(\bm{\tau}_{ext} + \Delta_{comp}\right).\label{eq:disturbance}
\end{align}
The disturbance term $\bm{d}$ therefore includes the acceleration-domain effect of the environmental interaction force, residual dynamics-compensation errors, and unmodeled effects such as unknown payloads.
In this study, $\bm{d}$ is modeled as a zeroth-order disturbance, i.e., a constant disturbance over the target observer bandwidth.
Although higher-order disturbance models and high-order observer designs can improve the estimation of rapidly varying interaction forces by introducing additional disturbance states~\cite{komada2002control,han2021toward}, they also require additional observer gains.
These additional gains can make tuning more difficult and may increase overshoot, peaking, and noise sensitivity when the estimated disturbance is fed back to the plant~\cite{shimada2023disturbance}.
Because the estimated disturbance is directly used for force feedback in 4-channel bilateral teleoperation, this study prioritizes a low-order observer structure with an explicit frequency-domain interpretation over high-order disturbance modeling, thereby reducing the risk of overshoot and noise amplification in force feedback.
Assuming that the disturbance varies slowly compared with the observer bandwidth, i.e., $\dot{\bm{d}} \approx \bm{0}$ within the target frequency range, the manipulator dynamics can be approximated by the following state-space model:
\begin{align}
\frac{d}{dt}
\begin{bmatrix}
\bm{q} \\
\dot{\bm{q}} \\
\bm{d} \\
\end{bmatrix}
&=
\begin{bmatrix}
\bm{0} & \bm{I} & \bm{0} \\
\bm{0} & \bm{0} & \bm{I} \\
\bm{0} & \bm{0} & \bm{0} \\
\end{bmatrix}
\begin{bmatrix}
\bm{q} \\
\dot{\bm{q}} \\
\bm{d} \\
\end{bmatrix}
+
\begin{bmatrix}
\bm{0} \\
\bm{I} \\
\bm{0} \\
\end{bmatrix}
\bm{M}(\bm{q})^{-1}
\bm{\tau}_{u}
\\
\bm{y}&=
\begin{bmatrix}
\bm{I} & \bm{0} & \bm{0} \\
\end{bmatrix}
\begin{bmatrix}
\bm{q} \\
\dot{\bm{q}} \\
\bm{d} \\
\end{bmatrix}
.
\end{align}
Although the constant-disturbance assumption is not strictly satisfied in contact-rich manipulation, it provides a practical predictive model when reliable prior information about the external force is unavailable.
Time-varying force components can be tracked only within the observer bandwidth, and increasing this bandwidth is ultimately limited by encoder quantization noise.

\subsubsection{Minimal-Order Observer}
To reduce the phase lag introduced by the low-pass filter, a minimal-order observer is used instead of a full-order observer.
Although several methods have been proposed for constructing minimal-order observers~\cite{gopinath1971control}, here we explain a specific, simplified method.
First, treating the velocity as an observation, we will construct a full-order observer using a state-space model with only velocity and external force.
\begin{equation}
\begin{aligned}
\frac{d}{dt}
\begin{bmatrix}
\hat{\dot{\bm{q}}} \\
\hat{\bm{d}} \\
\end{bmatrix}
&=
\begin{bmatrix}
\bm{0} & \bm{I} \\
\bm{0} & \bm{0} \\
\end{bmatrix}
\begin{bmatrix}
\hat{\dot{\bm{q}}} \\
\hat{\bm{d}} \\
\end{bmatrix}
+
\begin{bmatrix}
\bm{I} \\
\bm{0} \\
\end{bmatrix}
\tilde{\bm{M}}(\bm{q})^{-1}
\bm{\tau}_{u}
\\ & \qquad
+
\begin{bmatrix}
2\bm{Z}\bm{\Omega}_{c} \\
\bm{\Omega}_{c}\bm{\Omega}_{c} \\
\end{bmatrix}
(\frac{d}{dt}\bm{q}-\hat{\dot{\bm{q}}})
\label{eq:vfob} \\
\end{aligned}
\end{equation}
\begin{equation}
\hat{\bm{\tau}}_{ext} \coloneqq \tilde{\bm{M}}(\bm{q}) \hat{\bm{d}}
\end{equation}
Here, the observer gain is parameterized by the cutoff frequency parameter 
$\bm{\Omega}_c \coloneqq \mathrm{diag}\!\left(\omega_{c,1}, \ldots, \omega_{c,n}\right)$ 
and the damping ratio 
$\bm{Z} \coloneqq \mathrm{diag}\!\left(\zeta_{1}, \ldots, \zeta_{n}\right)$.

\subsubsection{Frequency-Domain Interpretation and Tuning}
For simplicity, we consider the case where the parameters are identical for all joints, i.e., $\bm{\Omega}_c=\omega_c\bm{I}, \bm{Z}=\zeta\bm{I}$.
In this case, the transfer-function representation is given as follows.
\begin{align}
\hat{\dot{\bm{q}}} &= \underbrace{\frac{s}{s+2\zeta\omega_c}}_{\text{First-Order HPF}}\frac{1}{s}\ddot{\bm{q}}_{ref} + \underbrace{\frac{2\zeta\omega_c}{s+2\zeta\omega_c}}_{\text{First-Order LPF}}s\bm{q}
\label{eq:vel_obs}
\\
&\qquad\qquad\qquad \left(\ddot{\bm{q}}_{ref} = \tilde{\bm{M}}(\bm{q})^{-1}(\bm{\tau}_{u}+\hat{\bm{\tau}}_{ext})\right) \label{eq:dd_theta_ref}
\\
\hat{\bm{\tau}}_{ext} &= \tilde{\bm{M}}(\bm{q}) \underbrace{\frac{\omega_c^2}{s^2 + 2\zeta\omega_c s + \omega_c^2}}_{\text{Second-Order LPF}} (s^2 \bm{q} - \tilde{\bm{M}}(\bm{q})^{-1}\bm{\tau}_{u})
\label{eq:tau_ext_obs}
\end{align}

Regarding the equation~\eqref{eq:vel_obs}, the velocity estimation can be interpreted as a first-order complementary filter that incorporates the estimated external force, despite the observer's second-order structure. The first term predicts the velocity based on the input torque and the nominal dynamics model.
The second term of the equation~\eqref{eq:vel_obs} is a feedback term using a derivative of angle measurement values with a first-order low-pass filter, and the cutoff frequency is set to the same as the first-order high-pass filter of the first term.

Equation~\eqref{eq:tau_ext_obs} shows that the external force estimate is obtained from the difference between the observed acceleration and the acceleration predicted from the input torque, filtered through a second-order low-pass filter.
This implies that the observer gain settings correspond to the design of a second-order low-pass filter.
Moreover, the cutoff frequency used in the velocity estimation described above is also determined using these filter parameters.
In external force estimation, prior information about external force is typically unavailable, making it difficult to apply statistical approaches such as Kalman filtering.
Therefore, a cutoff-frequency-based parameter tuning is a practical and commonly adopted approach.
The cutoff frequency $\omega_c$ is selected based on the sensor's noise characteristics.
For a second-order low-pass filter, the damping ratio $\zeta$ is typically chosen from a narrow range, and its tuning freedom is limited.
In this study, the damping ratio is fixed to critical damping ($\zeta=1$), which yields the fastest response among overshoot-free configurations.

\begin{figure*}[t]
    \centering
    \includegraphics[width=\linewidth]{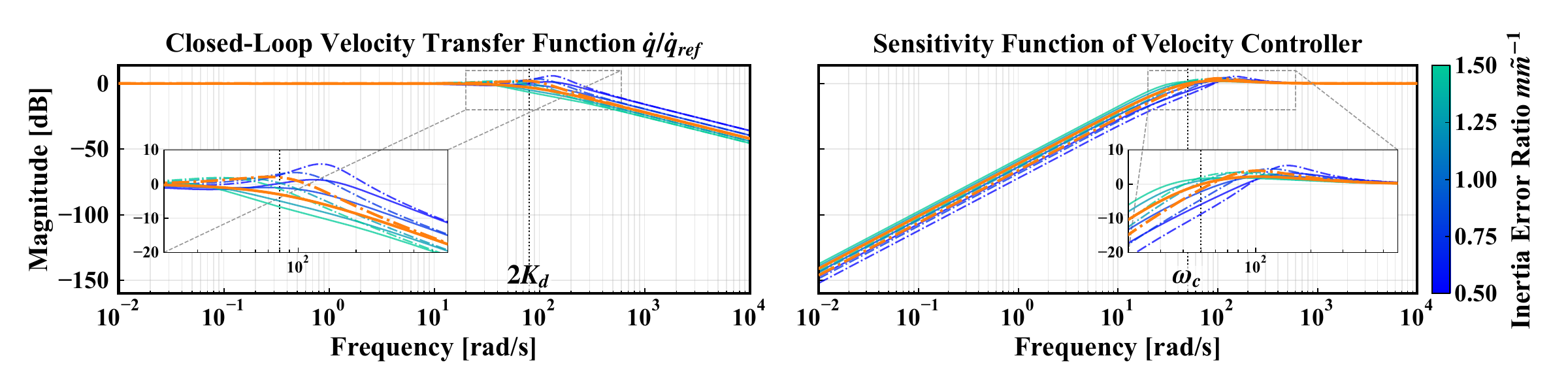}
    \caption{
        \textbf{Frequency response analysis of the velocity control layer under inertia variations.} \\
The left panel shows the tracking transfer function from the velocity reference $\dot{q}_{ref}$ to the actual velocity $\dot{q}$, and the right panel illustrates the sensitivity function of the velocity controller.
In both plots, the solid lines represent the case with lead compensation, and the dashed lines show the case without lead compensation for comparison.
The nominal case ($m\tilde{m}^{-1} = 1.0$) is highlighted in orange to emphasize its ideal response.
The inset plots provide a magnified view of the response around the cutoff frequency region and demonstrate the suppression of peaking, even under significant model uncertainties.
    }
    \label{fig:bode_plot_vdob}
\end{figure*}

\subsection{Effect of Inertia Variations}
A key issue in applying disturbance-observer-based control to manipulators is that the inertia is configuration-dependent. If a constant nominal inertia is used, the inertia mismatch is lumped into the estimated disturbance and can only be compensated within the observer bandwidth. This limitation is particularly important for low-cost manipulators, where the observer cutoff frequency must be kept relatively low because of encoder quantization noise. Therefore, using an identified configuration-dependent inertia model reduces the disturbance component that must be compensated by the observer and improves the achievable control performance.
The identified inertia model also justifies the use of model-based velocity prediction. Conventional velocity estimation based on numerical differentiation with a first-order low-pass filter ($\hat{\dot{\bm{q}}} = \frac{\omega_c}{s+\omega_c}s\bm{q}$) suffers from phase lag, whereas the velocity observer combines joint-angle measurements with model-based predictions from the input torque. This reduces estimation delay and enables higher position-control gains.
Although the actual inertia varies with configuration, the following analysis provides a local frequency-domain interpretation by approximating the inertia as constant over a small motion range and treating its deviation from the nominal model as an inertia mismatch.

\subsubsection{Acceleration Layer as Velocity FF-I-P Control}
The angle difference position controller can be interpreted as a cascade controller for acceleration, velocity, and position control.
The block diagram as a cascade controller is shown in Fig.~\ref{fig:block_diagram_cascade}.
From the equation~\eqref{eq:dd_theta_ref}, 
the acceleration reference $\ddot{\bm{q}}_{ref}$
can be written as follows:
\begin{equation}
\begin{aligned}
\ddot{\bm{q}}_{ref} &= \tilde{\bm{M}}(\bm{q})^{-1}(\bm{\tau}_{u}+\hat{\bm{\tau}}_{ext})
\\
&= \frac{s^2 + 2\zeta\omega_c s}{s^2 + 2\zeta\omega_c s + \omega_c^2}\tilde{\bm{M}}(\bm{q})^{-1}\bm{\tau}_{u} + \frac{\omega_c^2}{s^2 + 2\zeta\omega_c s + \omega_c^2}s^2\bm{q} \label{eq:acc_ref_cascade}
.
\end{aligned}
\end{equation}
Expanding the equation~\eqref{eq:acc_ref_cascade} for the input torque $\bm{\tau}_{u}$,
\begin{align}
\bm{\tau}_{u} &= \tilde{\bm{M}}(\bm{q})\frac{s^2 + 2\zeta\omega_c s + \omega_c^2}{s^2 + 2\zeta\omega_c s} \left( \ddot{\bm{q}}_{ref} - \frac{\omega_c^2}{s^2 + 2\zeta\omega_c s + \omega_c^2}s^2\bm{q} \right) \notag \\
&= \tilde{\bm{M}}(\bm{q})\Bigg\{\ddot{\bm{q}}_{ref} - \underbrace{\frac{\omega_c}{2\zeta}\frac{2\zeta\omega_c}{s + 2\zeta\omega_c} \left( s\bm{q}-\frac{1}{s}\ddot{\bm{q}}_{ref} \right)}_{\text{disturbance estimate }\hat{\bm{d}}=\tilde{\bm{M}}(\bm{q})^{-1}\hat{\bm{\tau}}_{ext}} \Bigg\} \label{eq:acceleration_integral_control}
.
\end{align}
Equation~\eqref{eq:acceleration_integral_control} shows that the acceleration layer can be interpreted as a velocity feedforward--integral--proportional (FF-I-P) controller with a first-order low-pass filter.
Because this structure relies on velocity feedback rather than direct acceleration feedback, it can be implemented using only position measurements from a low-resolution rotary encoder.
In the implementation, the internal state is maintained as an estimate of the disturbance rather than as an integral of the control error.
The observer state is updated using the saturated control input actually applied to the actuator, which naturally prevents integral windup.
Qualitatively, this acceleration control mechanism can be interpreted as follows: the predicted velocity, obtained by integrating the acceleration reference, is compared with the velocity obtained from differentiating the position measurement.
The velocity prediction error filtered through a first-order low-pass filter is fed back to compensate for the torque deficit relative to the acceleration reference.

This FF-I-P controller interpretation also explains why a constant-disturbance model is used in this study.
The constant-disturbance model introduces an integral-like compensation mechanism in the acceleration layer, which rejects low-frequency disturbances and constant or low-frequency model errors.
Although higher-order disturbance models can improve the estimation of rapidly varying disturbances by adding disturbance states, these additional states behave as extra integrator-like dynamics when the estimated disturbance is fed back to the plant.
As a result, the closed-loop order increases, and overshoot, peaking, and noise sensitivity may become more pronounced.
Because the estimated disturbance is directly used for force feedback and reflected to the operator in bilateral teleoperation, this study deliberately adopts the zeroth-order disturbance model and tunes the resulting second-order observer through the cutoff frequency and damping ratio.

From equations ~\eqref{eq:tau_u},~\eqref{eq:disturbance}, and~\eqref{eq:acceleration_integral_control}, the transfer function of the acceleration control layer is as follows:
\begin{align}
\bm{\ddot{q}} &= m^{-1}\bm{\tau}_u + \bm{d} \notag \\
&= \frac{s^2+2\zeta \omega_c s+\omega_c^2}{m\tilde{m}^{-1}(s^2+2\zeta \omega_c s)+\omega_c^2}\bm{\ddot{q}}_{ref} 
\notag \\
&\qquad
+\frac{m\tilde{m}^{-1}(s^2+2\zeta \omega_c s)}{m\tilde{m}^{-1}(s^2+2\zeta \omega_c s) + \omega_c^2}\bm{d}\label{eq:transfer_function_acceleration_s}\\
&\coloneqq G_a(s)\bm{\ddot{q}}_{ref}+S_a(s)\bm{d}
\label{eq:transfer_function_acceleration_general}
\end{align}
where, for simplicity, the inertia matrix $\bm{M}(\bm{\theta})$ is approximated as a constant and isotropic matrix, i.e., $\bm{M}(\bm{\theta}) = m\bm{I}$.
In particular, when there is no inertial error ($m\tilde{m}^{-1}=1$),
\begin{align}
\bm{\ddot{q}} &= \bm{\ddot{q}}_{ref} + \underbrace{\frac{s^2+2\zeta \omega_c s}{s^2+2\zeta \omega_c s + \omega_c^2}}_{\text{Second-Order HPF}} \bm{d}.\label{eq:transfer_function_acceleration_nominal}
\end{align}
Here, $\bm{d}$ denotes the equivalent disturbance in the acceleration domain, defined as the external joint torque disturbance normalized by the nominal inertia, defined as equation~\eqref{eq:disturbance}.
Equation~\eqref{eq:transfer_function_acceleration_nominal} shows that the transfer function from the acceleration reference to the actual acceleration ideally becomes unity, while disturbance rejection is shaped as a second-order high-pass filter, complementary to the second-order low-pass filter used in the external force estimation.
In the absence of model errors and external disturbances, the velocity feedback cancels the integral effect, so that the closed-loop acceleration dynamics coincide with the nominal acceleration tracking design. This structure can be regarded as a two-degree-of-freedom servo with nominal integral cancellation~\cite{fujisaki1992two}.
From equation~\eqref{eq:transfer_function_acceleration_s}, even when the nominal inertia differs from the actual plant inertia, the controller forces the low-frequency closed-loop behavior to match the nominal inertia. Although the bandwidth of this inertia nominalization is limited by the cutoff frequency $\omega_c$, high-frequency components of the acceleration reference are directly transmitted to the actual acceleration, with a gain determined by the inertia mismatch.
Consequently, the acceleration layer does not necessarily require a higher bandwidth than the velocity layer, unlike typical cascade control structures composed of stacked first-order subsystems.

\subsubsection{Velocity Layer as Lead-Compensated P Control}
Furthermore, substituting equation~\eqref{eq:vel_obs} into the PD position control of the angle difference in equation~\eqref{eq:acc_ref_-+}, the acceleration reference of the joint angle difference is:
\begin{equation}
\begin{aligned}
\ddot{\bm{q}}_{-ref} &= -2\bm{K}_{p}\bm{q}_{-}
- 2\bm{K}_{d}\hat{\dot{\bm{q}}}_{-} 
\\
&= -2\bm{K}_{p}\bm{q}_{-} -2\bm{K}_{d}\left(\frac{s}{s+2\zeta\omega_c}\frac{1}{s}\ddot{\bm{q}}_{-ref} + \frac{2\zeta\omega_c}{s+2\zeta\omega_c}s\bm{q}_{-} \right) \label{eq:acc_ref_-_with_vel_obs}
.
\end{aligned}
\end{equation}
For simplicity, the gains are assumed to be identical across all axes, i.e., $\bm{K}_p = K_p\bm{I}$ and $\bm{K}_d = K_d\bm{I}$, whereby (\ref{eq:acc_ref_-_with_vel_obs}) can be expanded as follows:
\begin{align}
\ddot{\bm{q}}_{-ref} &= \underbrace{\frac{s+2\zeta\omega_c}{s+2\zeta\omega_c+2K_d}}_{\text{Lead Compensation}} 2K_d \Bigg( -\frac{K_p}{K_d}\bm{q}_- -\underbrace{\frac{2\zeta\omega_c}{s+2\zeta\omega_c}}_{\text{First-Order LPF}}s\bm{q}_- \Bigg)
.\label{eq:lead_compensation}
\end{align}
Equation~\eqref{eq:lead_compensation} indicates that the estimated angular velocity feedback functions as a lead compensation.
This reduces phase lag and increases phase margin, thereby improving stability and responsiveness.
From equations~\eqref{eq:transfer_function_acceleration_general} and~\eqref{eq:lead_compensation}, the transfer function of the velocity control layer is as follows:
\begin{equation}
\begin{aligned}
\bm{\dot{q}}_{-}&= \frac{G_a(s)2K_d\frac{s+2\zeta\omega_c}{s+2\zeta\omega_c+2K_d}}{s+G_a(s)2K_d\frac{s+2\zeta\omega_c}{s+2\zeta\omega_c+2K_d}\frac{2\zeta\omega_c}{s+2\zeta\omega_c}}\bm{\dot{q}}_{-ref} 
\\
&\qquad+\frac{S_a(s)}{s+G_a(s)2K_d\frac{s+2\zeta\omega_c}{s+2\zeta\omega_c+2K_d}\frac{2\zeta\omega_c}{s+2\zeta\omega_c}}\bm{d}_{-}
\end{aligned}
\end{equation}
where $\bm{d}_{-} = \bm{d}_{l} - \bm{d}_{f}$.
In particular, when there is no inertial error ($m\tilde{m}^{-1}=1$),
\begin{equation}
\begin{aligned}
\bm{\dot{q}}_{-}&= \underbrace{\frac{s+2\zeta\omega_c}{s+2\zeta\omega_c}}_{\text{1}}\underbrace{\frac{2K_d}{s+2K_d}}_{\text{First-Order LPF}}\bm{\dot{q}}_{-ref} 
\\
&\qquad+ \underbrace{\frac{s(s+2\zeta\omega_c+2K_d)}{(s+2K_d)(s+2\zeta\omega_c)}}_{\text{Second-Order HPF}} \underbrace{\frac{s^2+2\zeta \omega_c s}{s^2+2\zeta \omega_c s + \omega_c^2}}_{\text{Second-Order HPF}} \frac{1}{s}\bm{d}_{-}
.\label{eq:transfer_function_velocity}
\end{aligned}
\end{equation}
As shown in \eqref{eq:transfer_function_velocity}, in the absence of modeling errors, the closed-loop velocity dynamics become first-order with a time constant determined by the velocity gain $K_d$ due to pole-zero cancellation.
Furthermore, the sensitivity function, which is the disturbance-to-acceleration transfer function, behaves as a fourth-order high-pass filter, resulting from the cascade of the second-order disturbance rejection dynamics in the acceleration layer and the second-order closed-loop characteristics of the outer velocity loop.

\subsubsection{Frequency Response under Inertia Mismatch}
Fig.~\ref{fig:bode_plot_vdob} presents the frequency response of the velocity control layer under variations in inertia. In the case of a velocity observer, the effective phase lead introduced around the cutoff frequency $2\zeta\omega_c$ mitigates the amplification of the resonant peak caused by inertia mismatch. As a result, the magnitude of the disturbance-to-acceleration transfer function remains bounded and does not exhibit significant resonant amplification over a realistic range of inertia variations, while the transfer from the velocity reference to the actual velocity remains close to an ideal first-order response.

\begin{figure*}[t]
    \begin{minipage}[b]{0.99\linewidth}
        \centering
        \includegraphics[width=\linewidth]{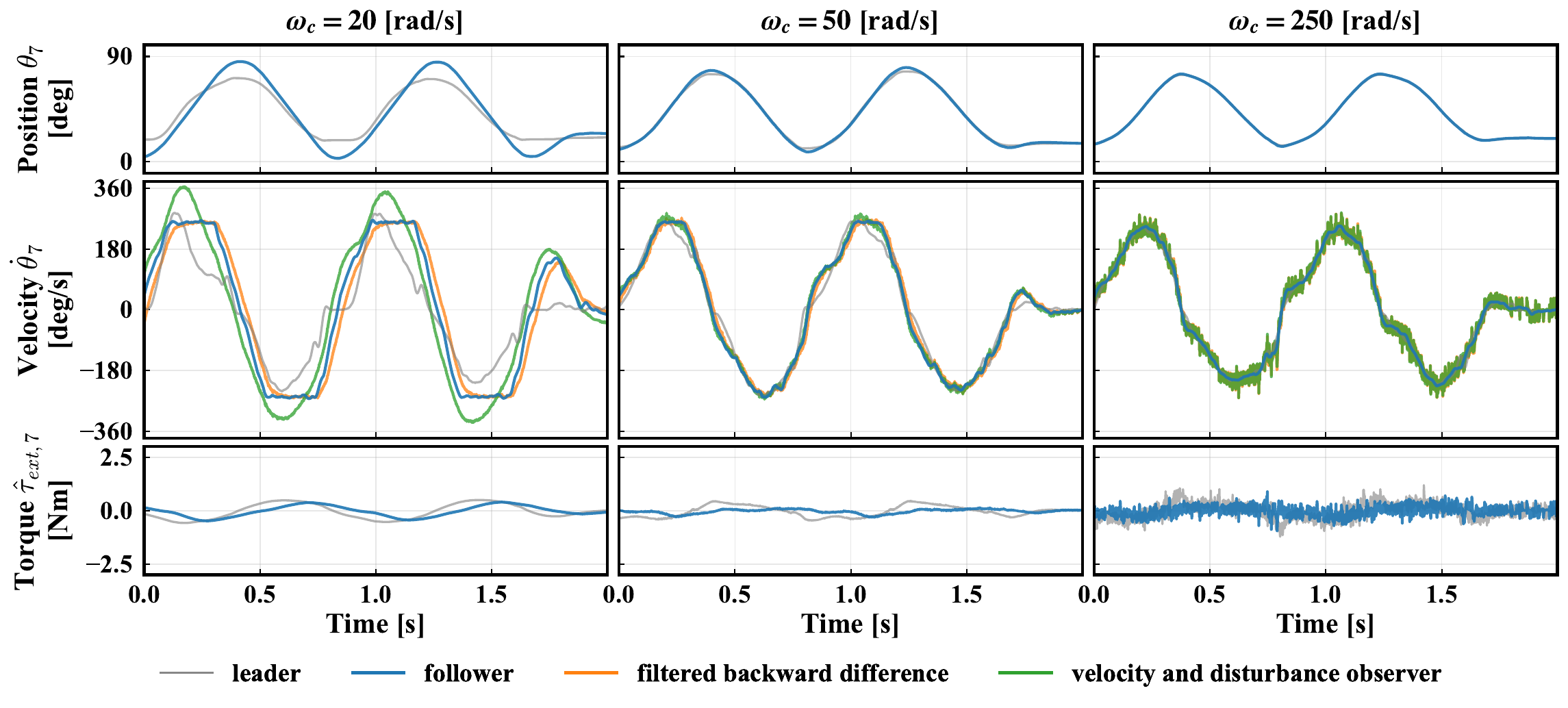}
    \end{minipage}
    \begin{minipage}[b]{0.99\linewidth}
        \centering
        \includegraphics[width=\linewidth]{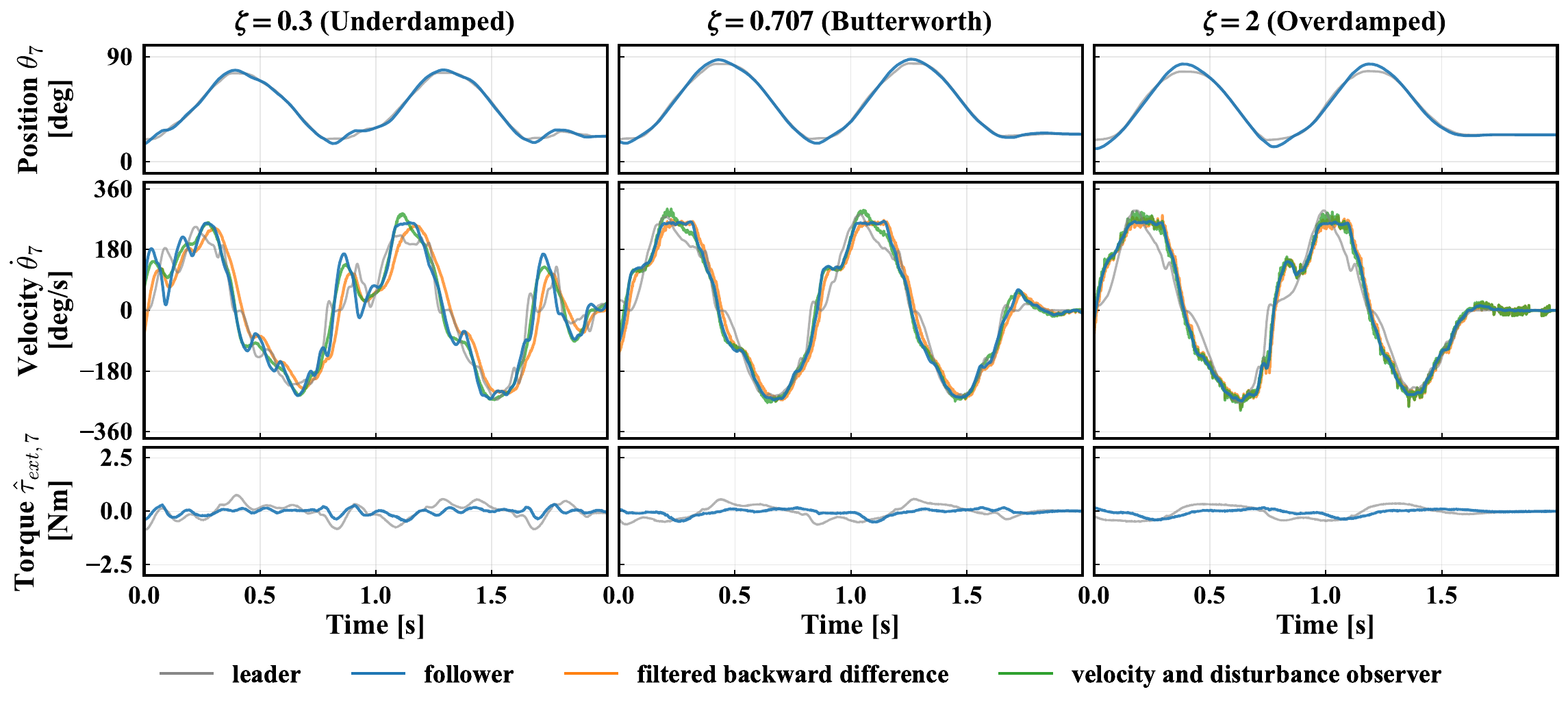}
    \end{minipage}
    \caption{
    \textbf{Comparison of cutoff frequency and damping ratio during free motion.} \\
    In the cutoff frequency comparison, the damping ratio $\zeta$ of the observer is set to 1. In the damping ratio comparison, the cutoff frequency is set to 50 rad/s.
    The leader and follower velocities were computed by applying Savitzky–Golay differentiation to the position signals (window length = 51, polynomial order = 3) and are used as near-zero-phase reference signals for comparison.
    }
    \label{fig:obs_param_comp_free_motion}
\end{figure*}

\begin{figure*}[t]
    \begin{minipage}[b]{0.99\linewidth}
        \centering
        \includegraphics[width=\linewidth]{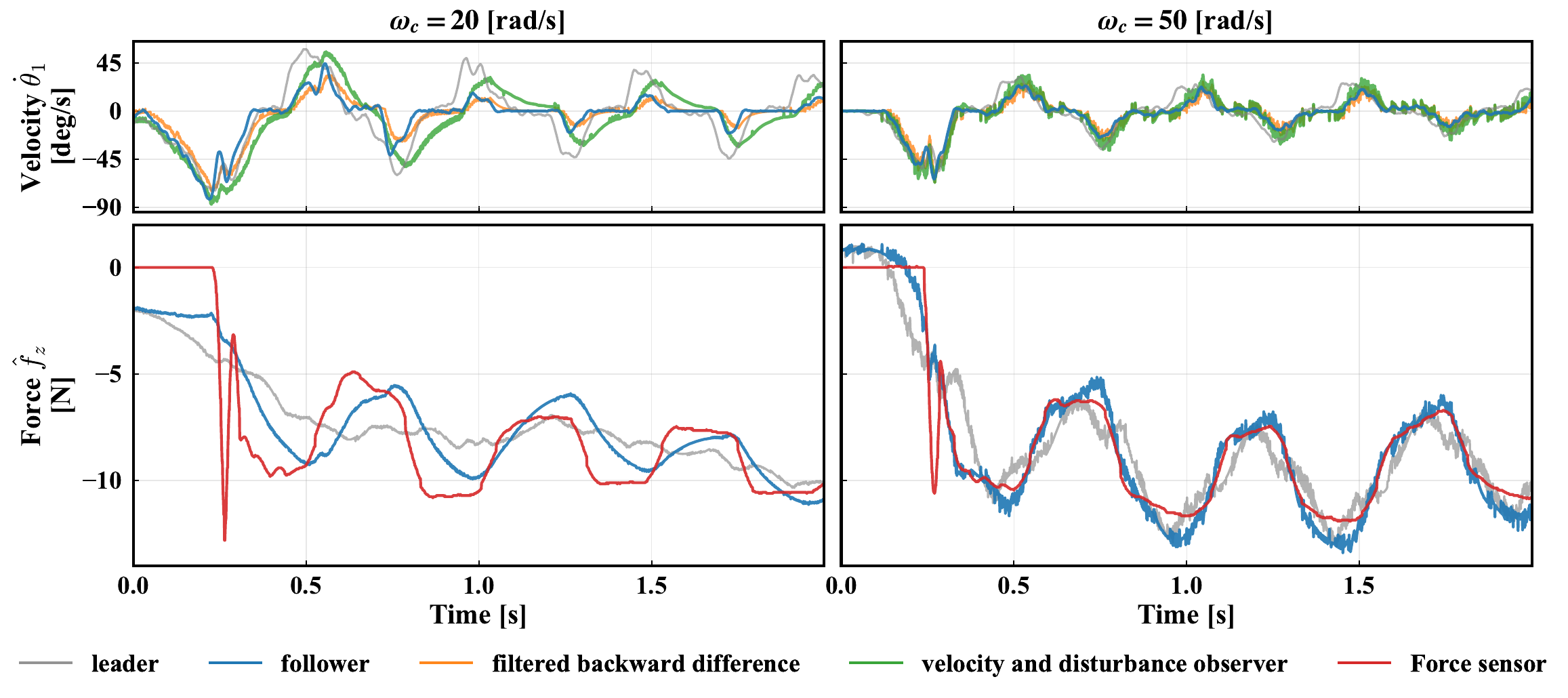}
    \end{minipage}
    \begin{minipage}[b]{0.99\linewidth}
        \centering
        \includegraphics[width=\linewidth]{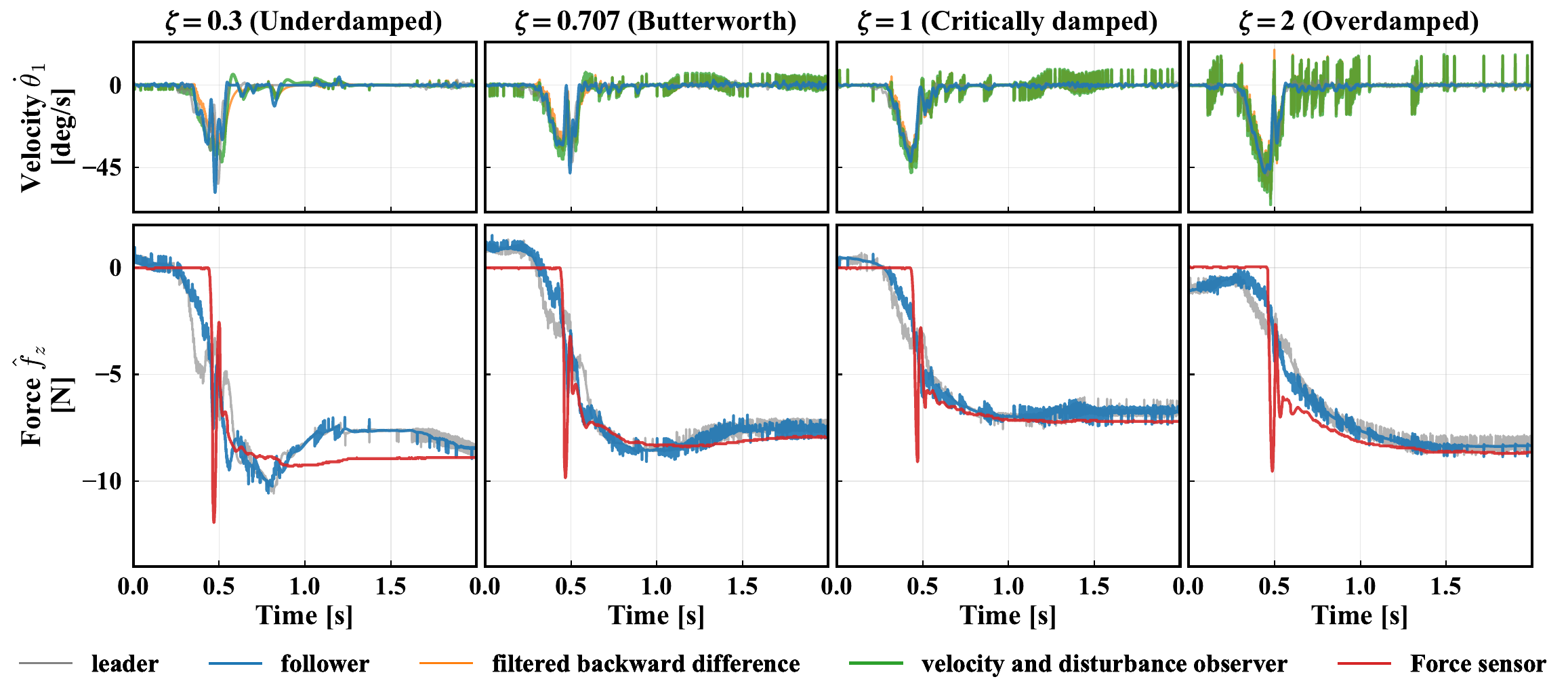}
    \end{minipage}
    \caption{
    \textbf{Comparison of cutoff frequency and damping ratio with force sensor measurements.} \\
    In the cutoff frequency comparison, the damping ratio $\zeta$ of the observer is set to 1. In the damping ratio comparison, the cutoff frequency is set to 50 rad/s.
    The leader and follower velocities were computed by applying Savitzky–Golay differentiation to the position signals (window length = 51, polynomial order = 3) and are used as near-zero-phase reference signals for comparison.
    The force sensor measurements include an internal low-pass filter with a cutoff frequency of 10 Hz ($\simeq 62.83$ rad/s).
    }
    \label{fig:obs_param_comp_force_sensor}
\end{figure*}

\begin{figure*}[t]
    \centering
    \begin{minipage}[b]{0.99\linewidth}
        \centering
        \begin{minipage}[b]{0.24\linewidth}
            \centering
            \includegraphics[width=\linewidth]{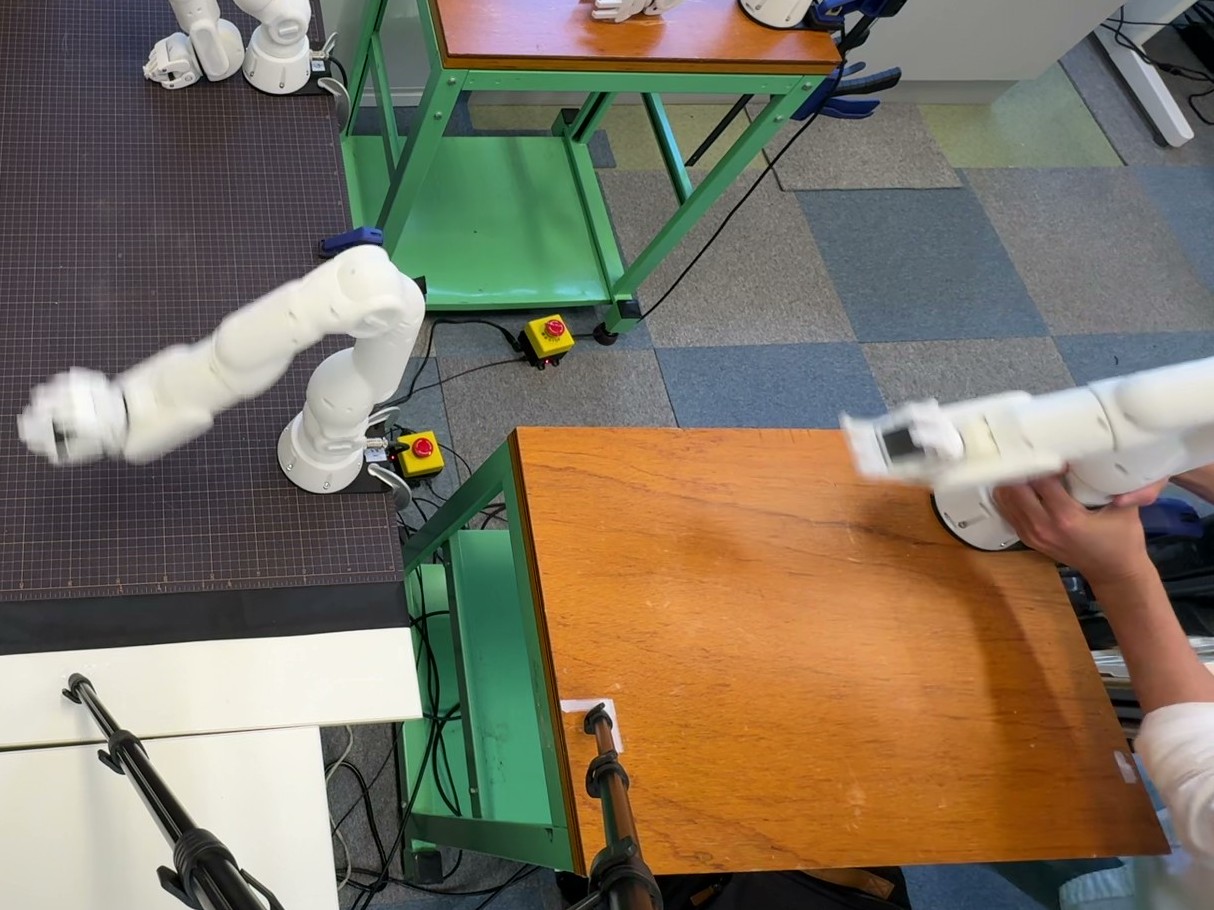}
        \end{minipage}
        \begin{minipage}[b]{0.24\linewidth}
            \centering
            \includegraphics[width=\linewidth]{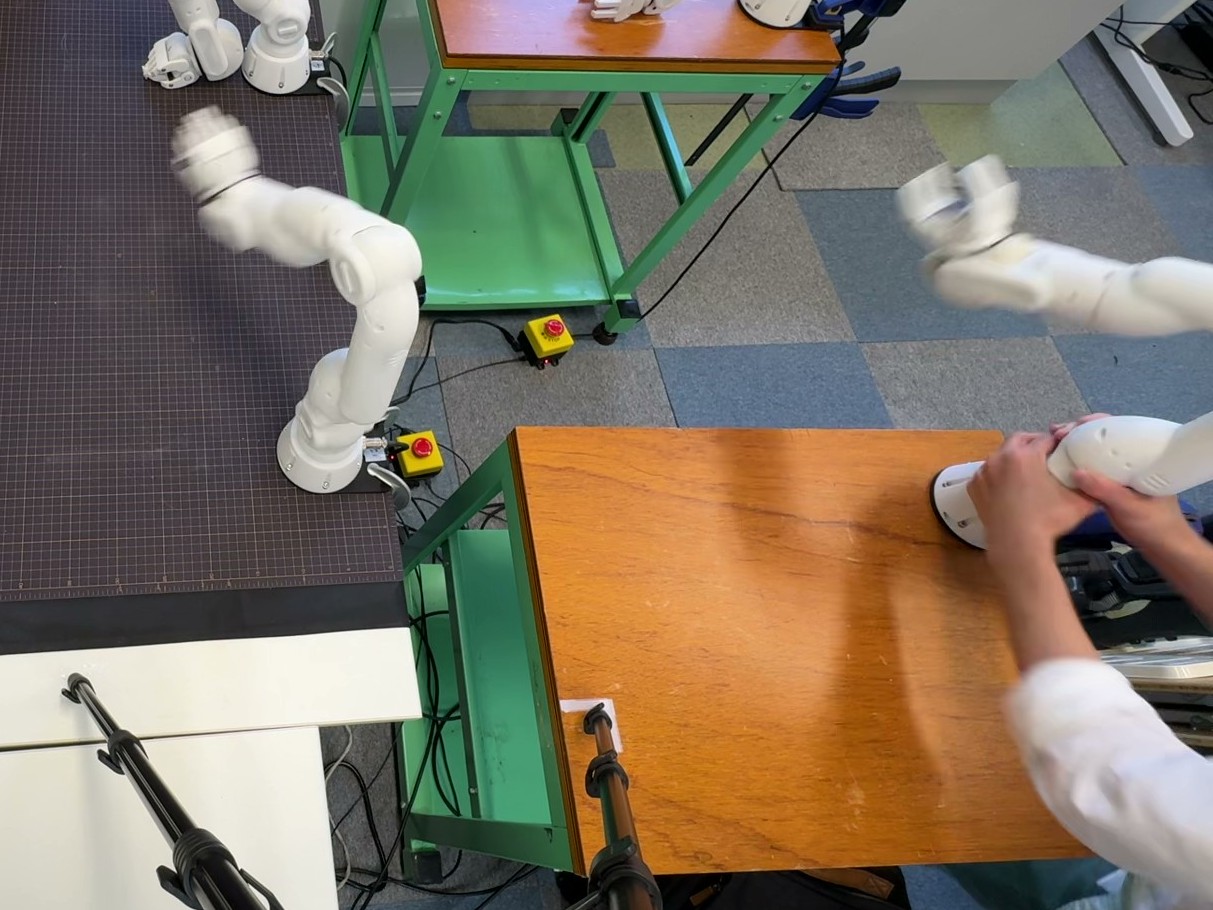}
        \end{minipage}
        \begin{minipage}[b]{0.24\linewidth}
            \centering
            \includegraphics[width=\linewidth]{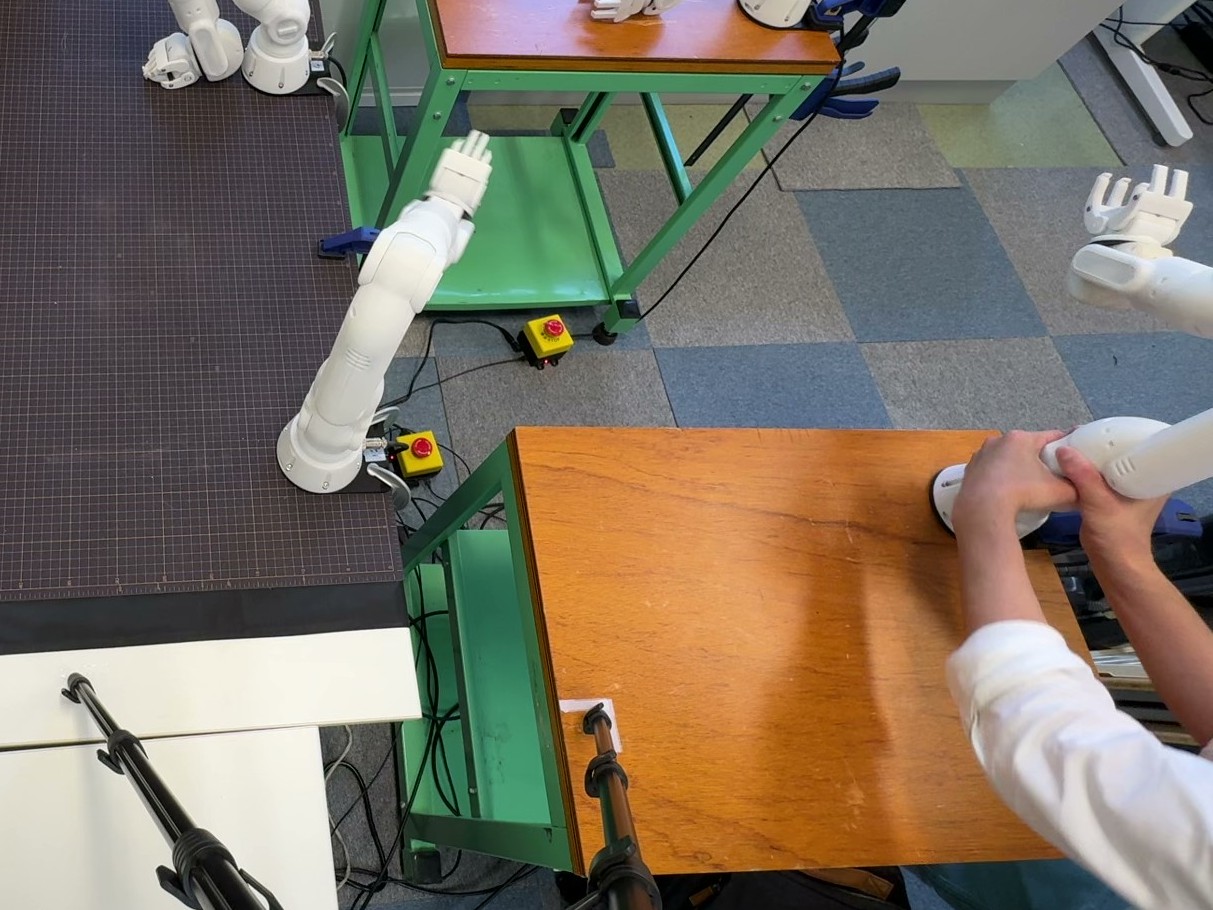}
        \end{minipage}
        \begin{minipage}[b]{0.24\linewidth}
            \centering
            \includegraphics[width=\linewidth]{figures/snapshot/teleoperation_comparison/teleoperation_comparison_free_motion_2_croped.jpg}
        \end{minipage}
        \subcaption{snapshot of free motion task}
        \label{fig:snapshot_free_motion}
    \end{minipage}
    
    \begin{minipage}[t]{0.99\linewidth}
        \includegraphics[width=\linewidth]{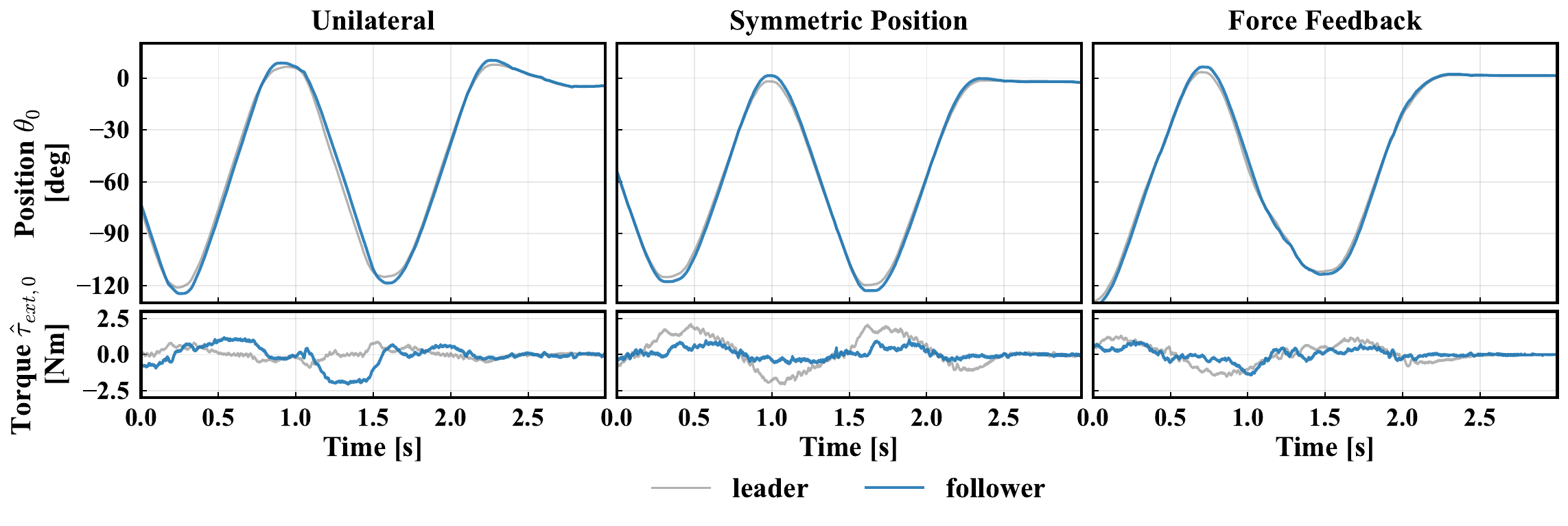}
        \subcaption{Other teleoperation methods}
    \end{minipage}
    
    \begin{minipage}[t]{0.99\linewidth}
        \includegraphics[width=\linewidth]{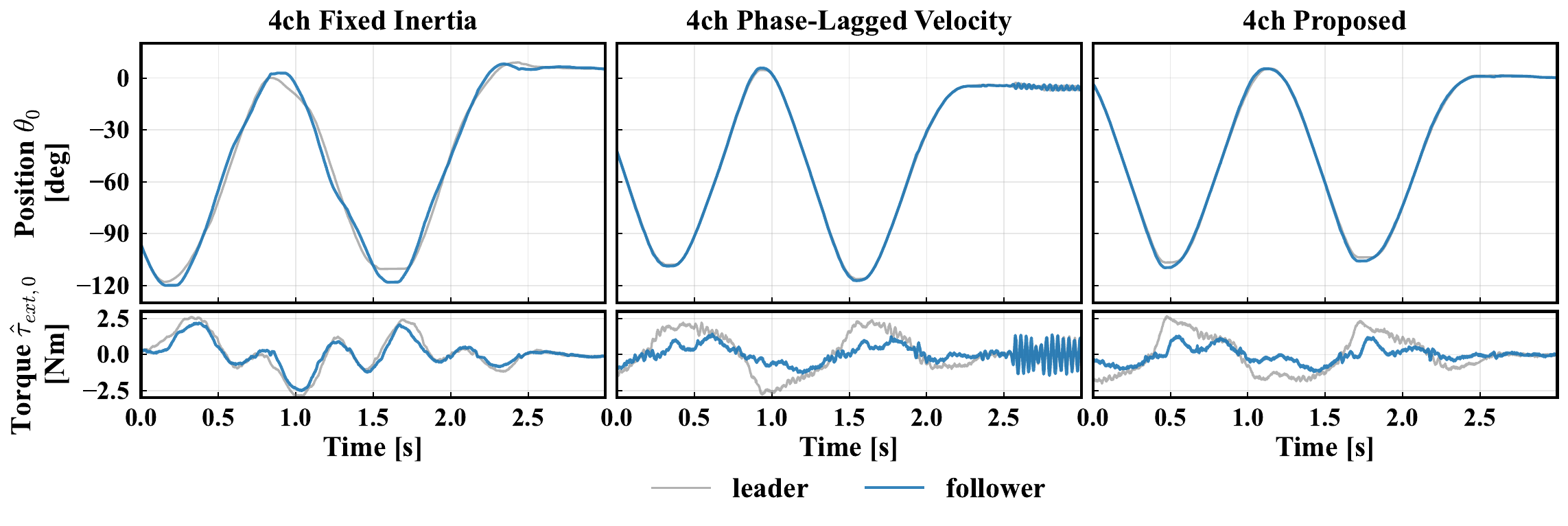}
        \subcaption{Ablation study}
    \end{minipage}
    \caption{
        Results of teleoperation comparison for free motion
    }
    \label{fig:teleoperation_comparison}
\end{figure*}

\begin{figure*}[t]
    \centering
    \begin{minipage}[t]{0.99\linewidth}
        \centering
        \begin{minipage}[b]{0.24\linewidth}
            \centering
            \includegraphics[width=\linewidth]{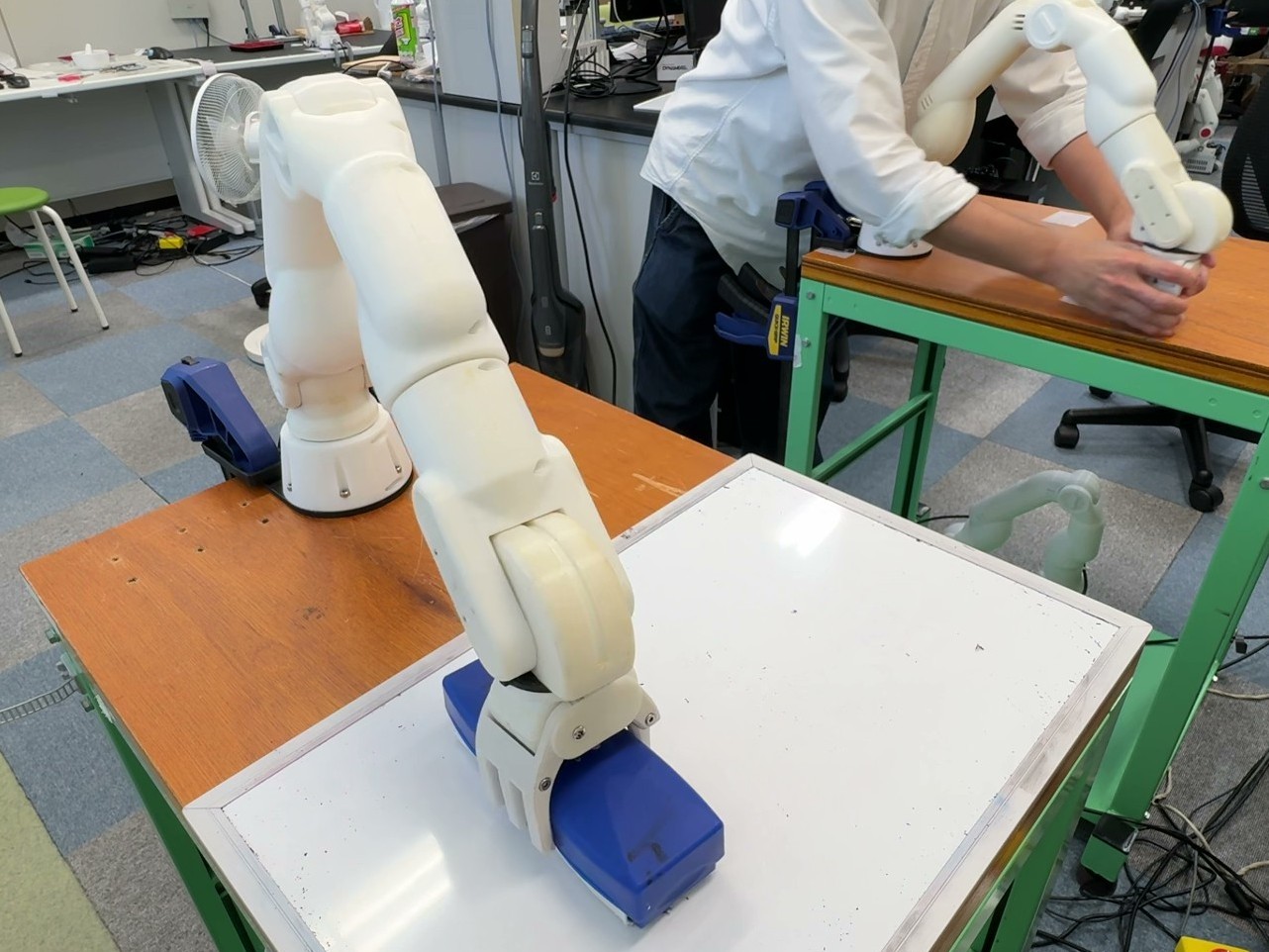}
        \end{minipage}
        \begin{minipage}[b]{0.24\linewidth}
            \centering
            \includegraphics[width=\linewidth]{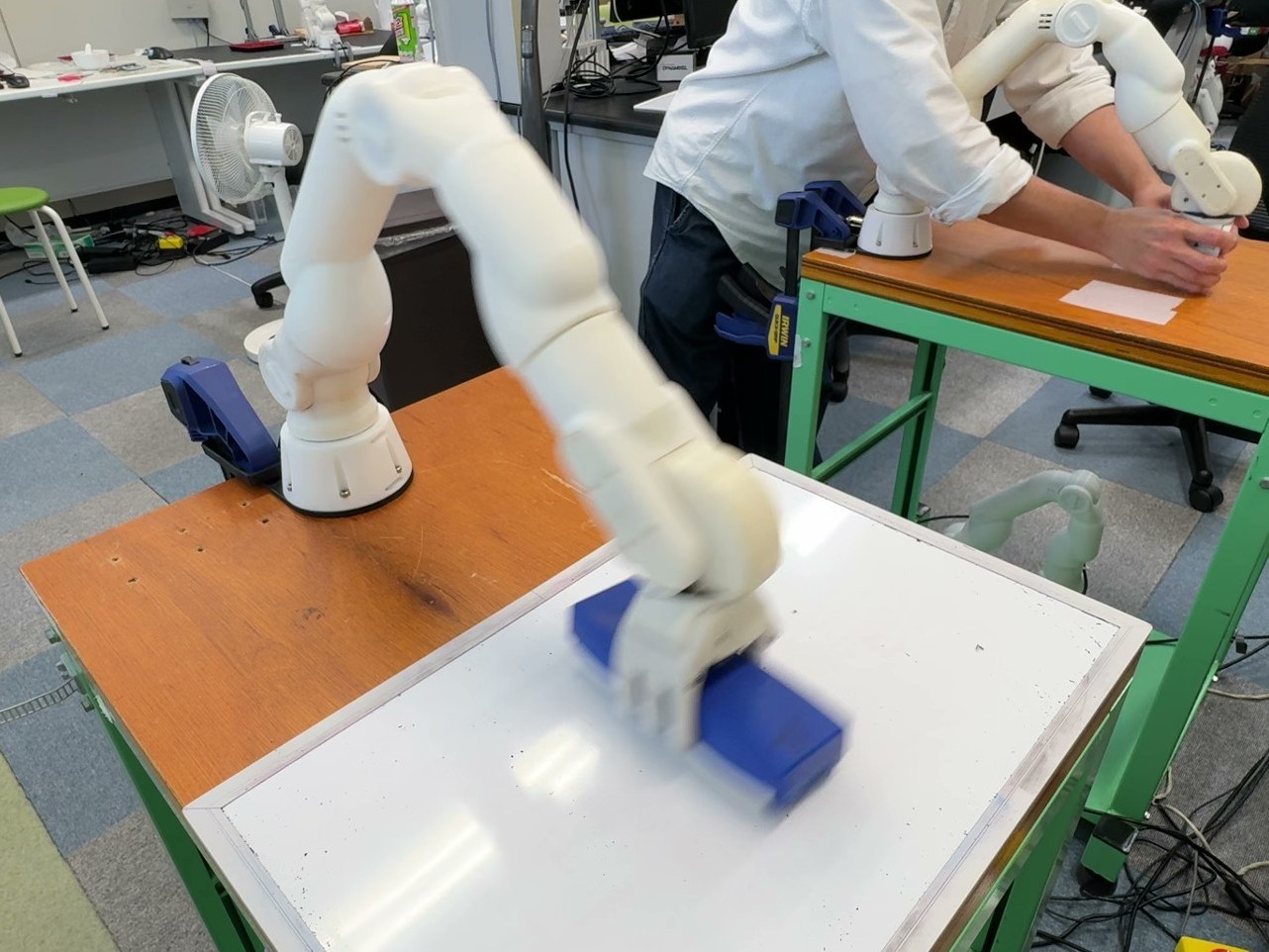}
        \end{minipage}
        \begin{minipage}[b]{0.24\linewidth}
            \centering
            \includegraphics[width=\linewidth]{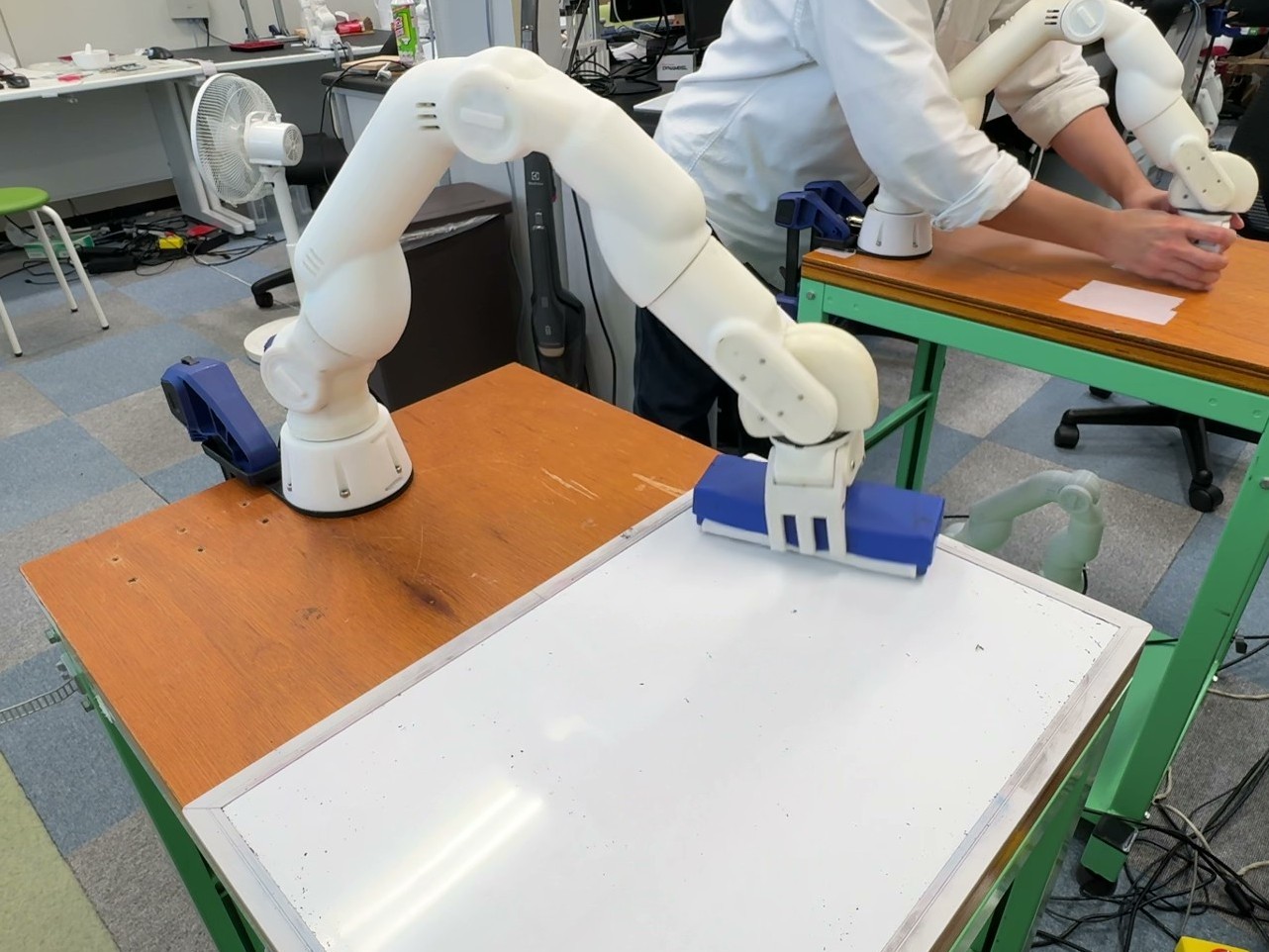}
        \end{minipage}
        \begin{minipage}[b]{0.24\linewidth}
            \centering
            \includegraphics[width=\linewidth]{figures/snapshot/teleoperation_comparison/teleoperation_comparison_wiping_2_croped.jpg}
        \end{minipage}
        \subcaption{Snapshot of wiping task}
        \label{fig:snapshot_wiping}
    \end{minipage}
    
    \begin{minipage}[t]{0.99\linewidth}
        \includegraphics[width=\linewidth]{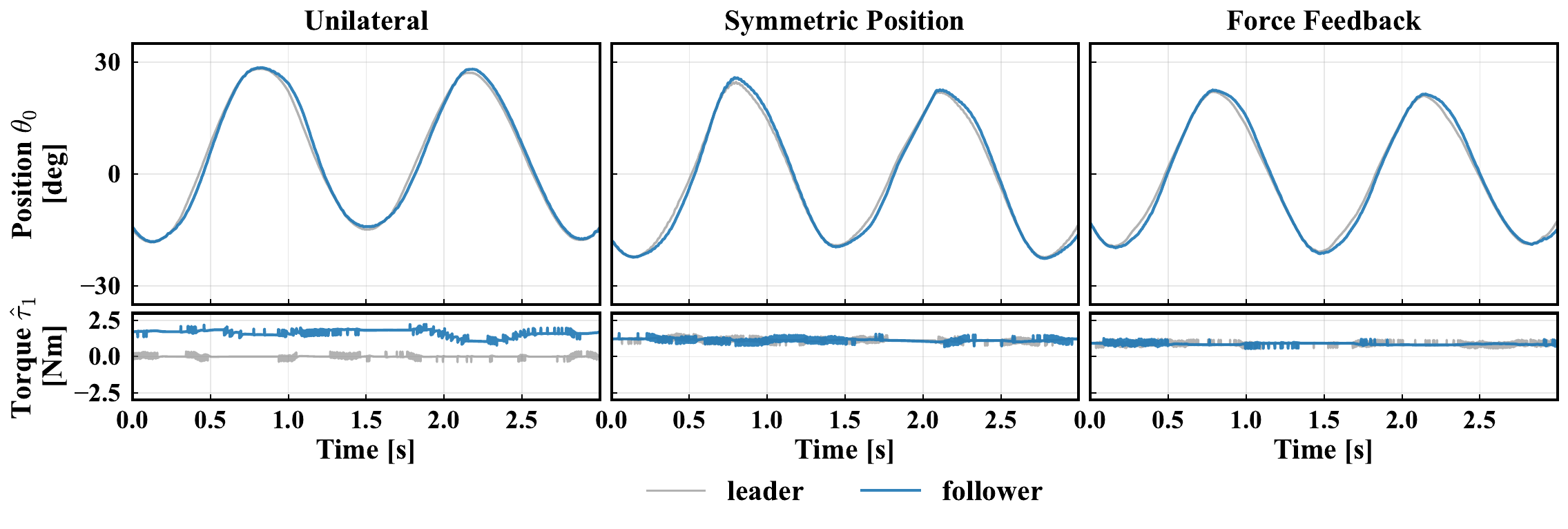}
        \subcaption{Other teleoperation methods}
    \end{minipage}
    
    \begin{minipage}[t]{0.99\linewidth}
        \includegraphics[width=\linewidth]{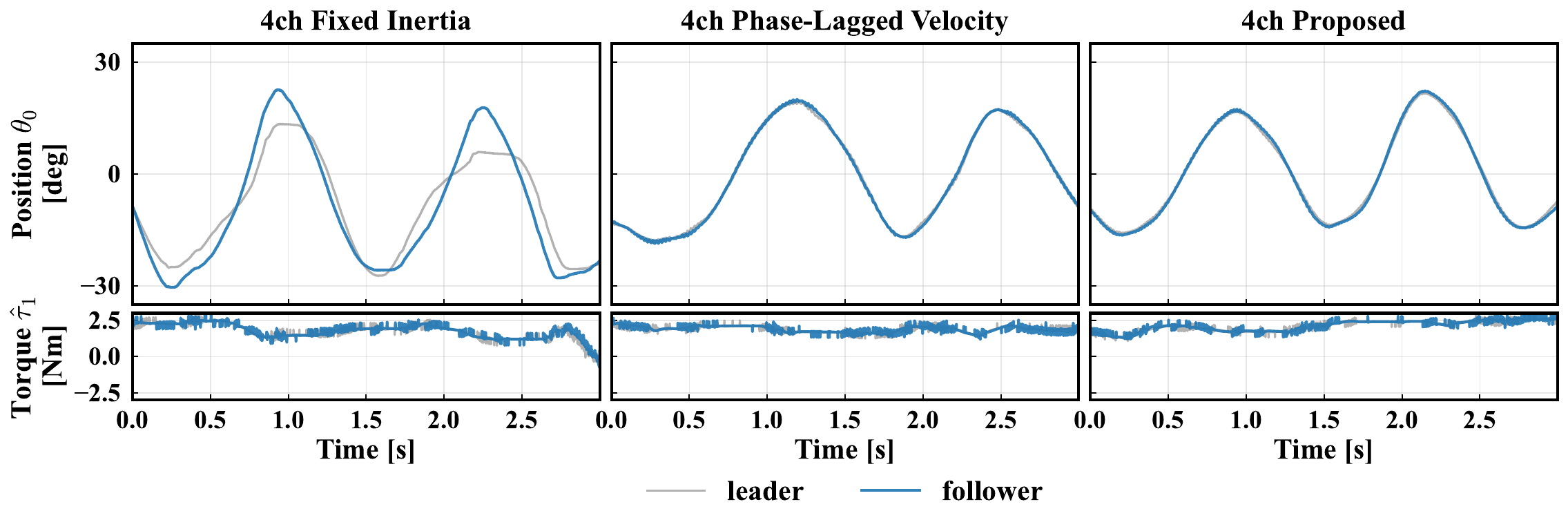}
        \subcaption{Ablation study}
    \end{minipage}
    \caption{
        Results of teleoperation comparison for wiping task
    }
    \label{fig:teleoperation_comparison_wiping}
\end{figure*}

\begin{figure*}[t]
    \begin{minipage}[b]{0.99\linewidth}
        \centering
        \begin{minipage}[b]{0.24\linewidth}
            \centering
            \includegraphics[width=\linewidth]{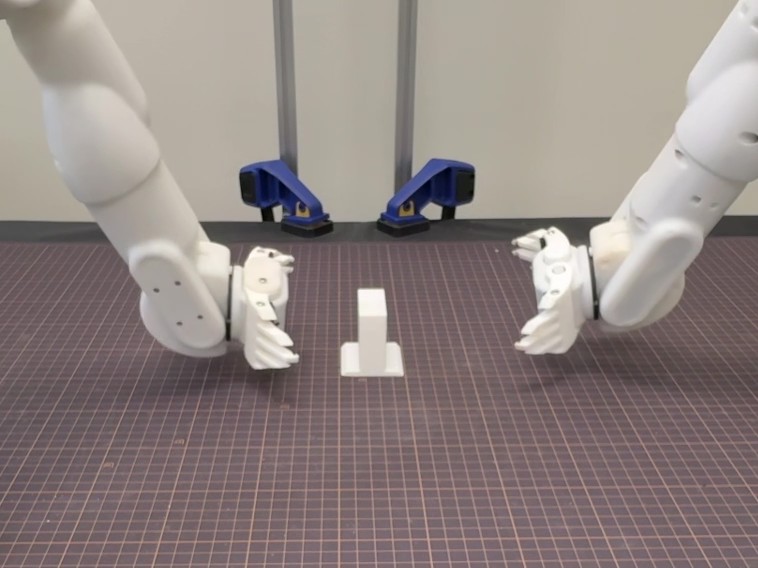}
        \end{minipage}
        \begin{minipage}[b]{0.24\linewidth}
            \centering
            \includegraphics[width=\linewidth]{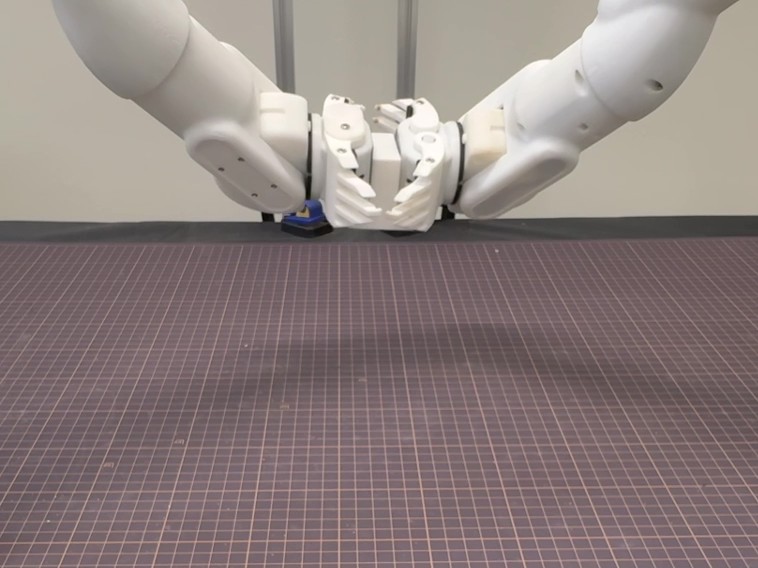}
        \end{minipage}
        \begin{minipage}[b]{0.24\linewidth}
            \centering
            \includegraphics[width=\linewidth]{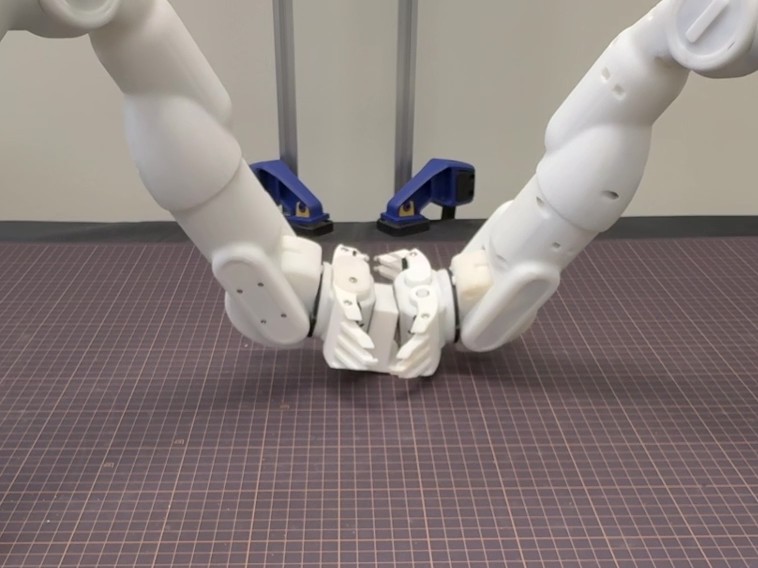}
        \end{minipage}
        \begin{minipage}[b]{0.24\linewidth}
            \centering
            \includegraphics[width=\linewidth]{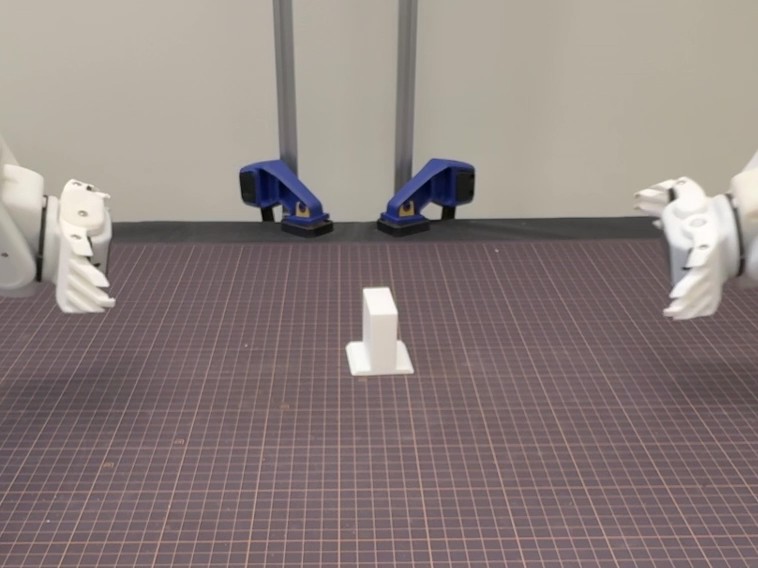}
        \end{minipage}
        \subcaption{Dual-Arm Pick-and-Place}
        \label{fig:snapshot_dual_arm_pick_and_place}
    \end{minipage}
    \begin{minipage}[b]{0.99\linewidth}
        \centering
        \begin{minipage}[b]{0.24\linewidth}
            \centering
            \includegraphics[width=\linewidth]{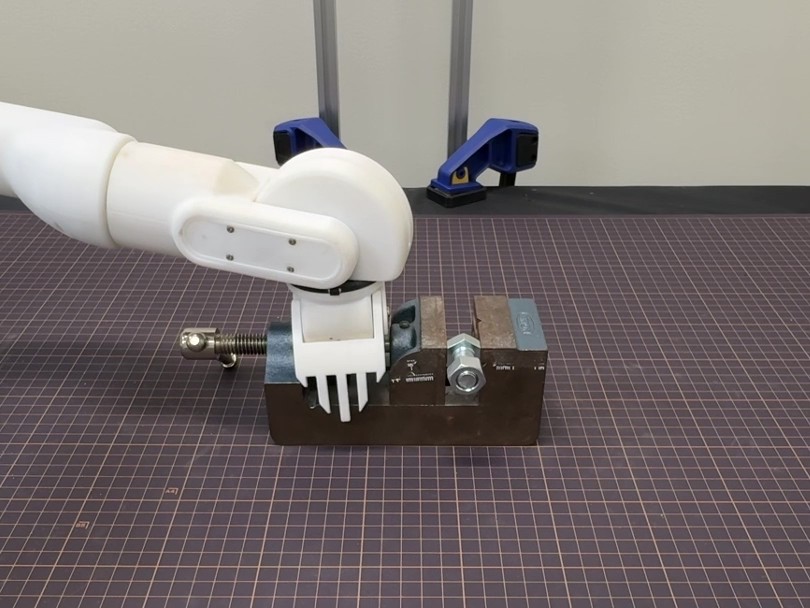}
        \end{minipage}
        \begin{minipage}[b]{0.24\linewidth}
            \centering
            \includegraphics[width=\linewidth]{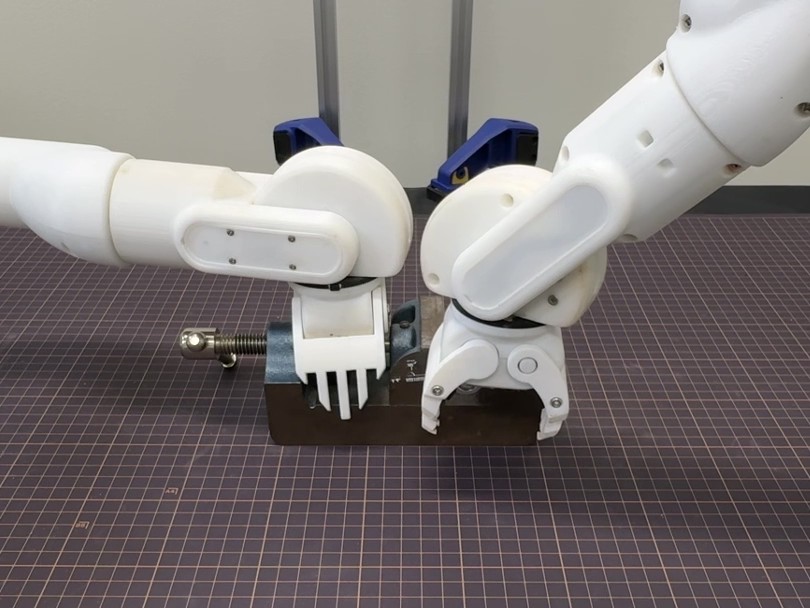}
        \end{minipage}
        \begin{minipage}[b]{0.24\linewidth}
            \centering
            \includegraphics[width=\linewidth]{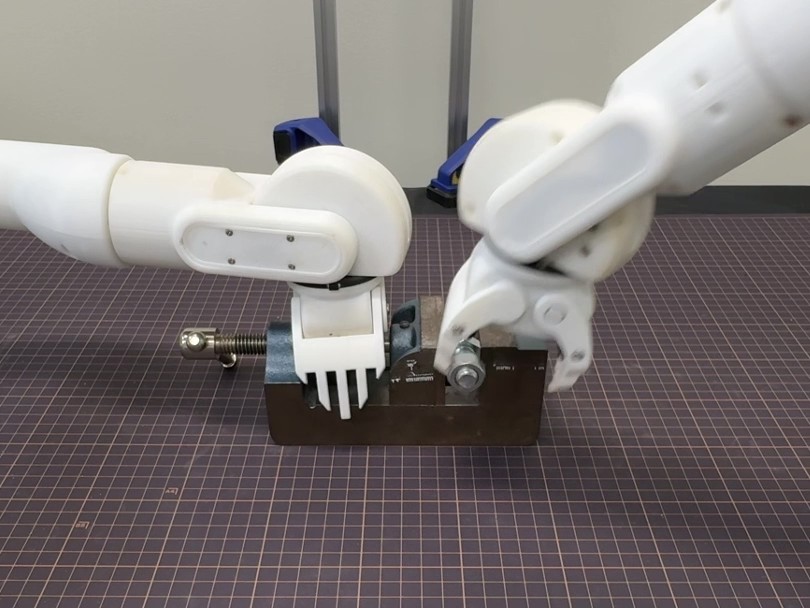}
        \end{minipage}
        \begin{minipage}[b]{0.24\linewidth}
            \centering
            \includegraphics[width=\linewidth]{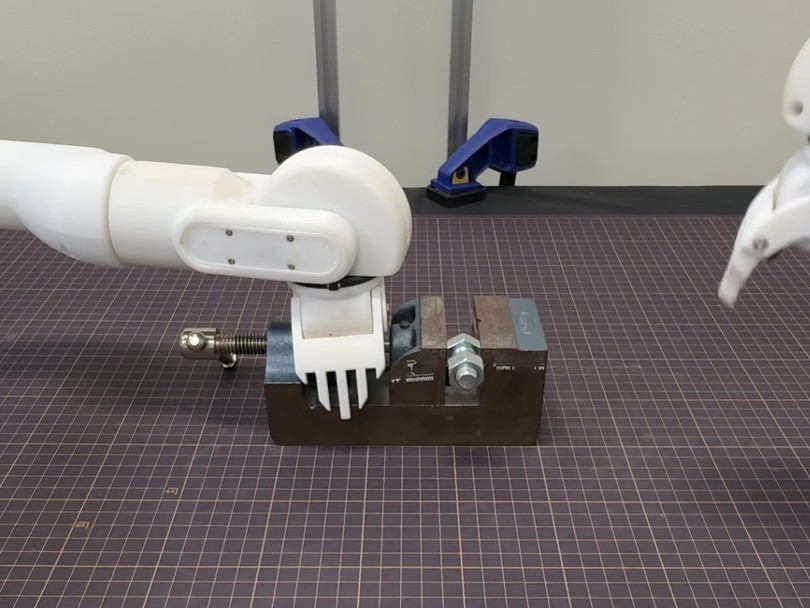}
        \end{minipage}
        \subcaption{Nut Turning}
        \label{fig:snapshot_nut_turning}
    \end{minipage}
    \begin{minipage}[b]{0.99\linewidth}
        \centering
        \begin{minipage}[b]{0.24\linewidth}
            \centering
            \includegraphics[width=\linewidth]{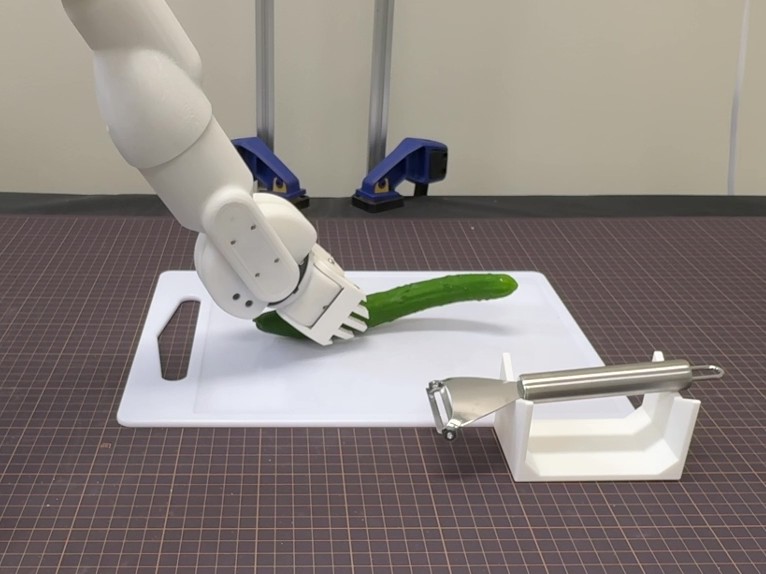}
        \end{minipage}
        \begin{minipage}[b]{0.24\linewidth}
            \centering
            \includegraphics[width=\linewidth]{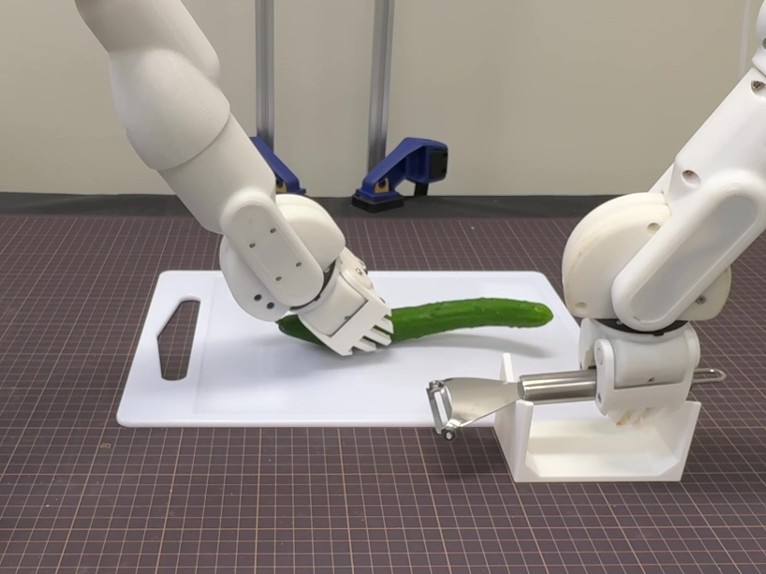}
        \end{minipage}
        \begin{minipage}[b]{0.24\linewidth}
            \centering
            \includegraphics[width=\linewidth]{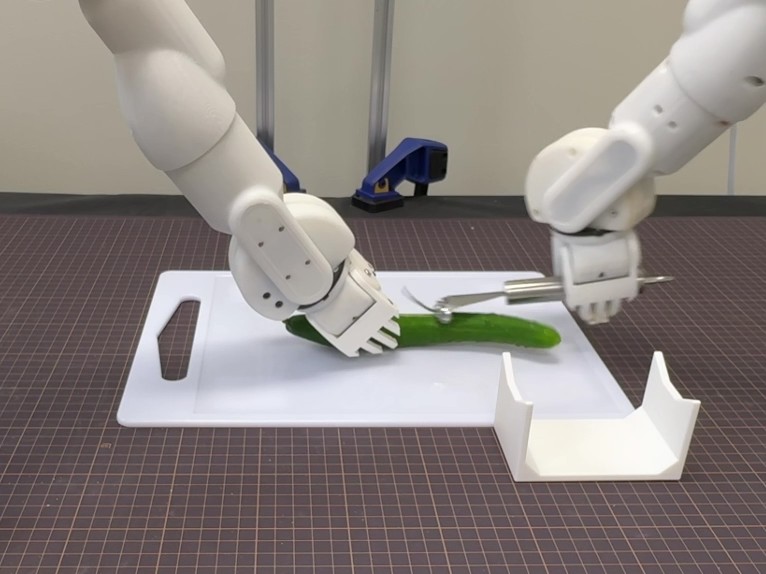}
        \end{minipage}
        \begin{minipage}[b]{0.24\linewidth}
            \centering
            \includegraphics[width=\linewidth]{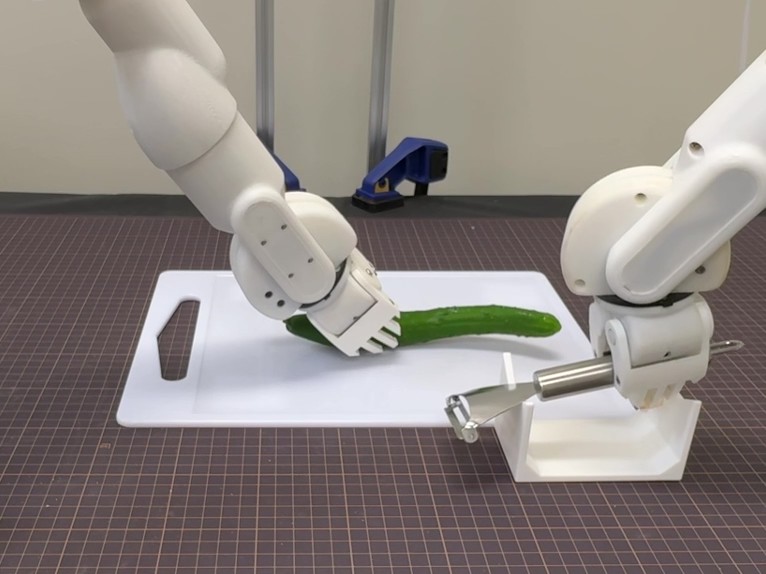}
        \end{minipage}
        \subcaption{Cucumber Peeling}
        \label{fig:snapshot_cucumber_peeling}
    \end{minipage}
    \caption{Imitation learning task snapshots}
    \label{fig:snapshot_IL}
\end{figure*}

\begin{table}[t]
    \caption{\textbf{Controller parameters}. Since weighting was performed based on identified inertia, the same parameters were used for each joint. $\bm{K}_{p}$ and $\bm{K}_{d}$ are set to double pole on the difference coordinate, and $\bm{K}_{f}$ is set so that the operator feels the original inertia.}
    \label{table:controller_parameters}
    \centering
    \begin{tabular}{clc}
        \toprule
         & Parameter & Value \\
        \midrule
        $\bm{K}_{p}$ & P gain for position control & 800$\bm{I}$ \\
        $\bm{K}_{d}$ & D gain for position control & 40$\bm{I}$ \\
        $\bm{K}_{f}$ & Gain for force control & $\{2\bm{M}(\bm{\theta})\}^{-1}$ \\
        $\bm{\Omega}_c$ & Cutoff frequency parameter of the observer [rad/s] & 50.0$\bm{I}$ \\
        $\bm{Z}$ & Damping ratio of the observer & $\bm{I}$ \\
        $f$ & Sampling frequency [Hz] & 1000 \\
        \bottomrule
    \end{tabular}
\end{table}

\begin{table}[t]
    \caption{
    \textbf{Mean Absolute Error (MAE) of joint angle, angular velocity, and torque.}
    Boldface indicates the lowest MAE among the non-shaded methods.
    The shaded cells indicate the oscillatory phase-lagged velocity case, which is excluded from the boldface comparison.
    For the fixed-inertia 4ch case, the torque MAE was computed using the corresponding fixed-inertia force estimate.
    }
    \label{table:MAE}
    \centering
    \begin{tabular}{l|lrr}
        \toprule
        & Teleoperation method & \makecell{Angle \\ $[$deg$]$} & \makecell{Torque \\ $[$Nm$]$} \\
        \midrule
        \multirow{6}{*}{\rotatebox{90}{free motion}}
        & Unilateral & 2.59 & 0.57 \\
        & Symmetric position type & 1.63 & 0.60 \\
        & Force feedback type & 1.55 & 0.36 \\
        & 4ch using fixed inertia & 3.19 & \textbf{0.28} \\
        & 4ch using phase-lagged velocity & \cellcolor{gray!20}0.66 & \cellcolor{gray!20}0.69 \\
        & 4ch (proposed) & \textbf{0.82} & 0.68 \\
        \midrule
        \multirow{6}{*}{\rotatebox{90}{wiping}}
        & Unilateral & 1.11 & 1.63 \\
        & Symmetric position type & 1.18 & 0.12 \\
        & Force feedback type & 1.05 & \textbf{0.09} \\
        & 4ch using fixed inertia & 4.50 & 0.14 \\
        & 4ch using phase-lagged velocity & \cellcolor{gray!20}0.45 & \cellcolor{gray!20}0.12 \\
        & 4ch (proposed) & \textbf{0.59} & 0.11 \\
        \bottomrule
    \end{tabular}
\end{table}

\section{Experiments}

\subsection{Velocity and External Force Estimation}
We implemented the velocity and disturbance observer and the bilateral teleoperation controller on CRANE-X7, a low-cost 7-DoF manipulator (RT Corporation, Tokyo, Japan). To evaluate the estimation performance and the effect of observer parameters, we conducted two types of experiments: free motion of the gripper and pushing against a force sensor along the sensor's $z$-axis. Several cutoff frequencies and damping ratios were compared to verify whether the observer can be practically tuned by fixing the damping ratio and adjusting only the cutoff frequency.

\subsubsection{Free motion}
The free-motion results are shown in Fig.~\ref{fig:obs_param_comp_free_motion}. The upper plot compares different cutoff frequencies under critical damping ($\zeta=1.0$). With the low cutoff frequency ($\omega_c=20~\mathrm{rad/s}$), position tracking is poor, and the velocity estimate overshoots the offline-differentiated velocity. This is because the acceleration control layer cannot sufficiently compensate for low-frequency disturbances, and the velocity predicted by integrating the acceleration reference becomes larger than the velocity obtained from the offline differentiation. With the middle cutoff frequency ($\omega_c=50~\mathrm{rad/s}$), the velocity estimate closely matches the offline-differentiated velocity, indicating that the phase lag of the filtered backward difference is effectively compensated. With the high cutoff frequency ($\omega_c=250~\mathrm{rad/s}$), the position-tracking performance is further improved, but both the velocity and external-torque estimates become noisy. This noise is mainly caused by the quantization of the low-resolution rotary encoder; therefore, in this hardware setup, encoder noise is the dominant factor limiting the cutoff frequency.

The lower plot in Fig.~\ref{fig:obs_param_comp_free_motion} compares different damping ratios with the middle cutoff frequency ($\omega_c=50~\mathrm{rad/s}$). The underdamped case ($\zeta=0.3$) exhibits oscillatory behavior, especially when changing direction. The Butterworth setting ($\zeta=1/\sqrt{2}\simeq0.707$) shows no significant degradation compared with critical damping ($\zeta=1.0$) in free motion. In contrast, the overdamped case ($\zeta=2.0$) shows poorer tracking performance and a higher noise level than the critically damped case.

\subsubsection{Comparison with force sensor}
The force-sensor pushing experiment results are shown in Fig.~\ref{fig:obs_param_comp_force_sensor}. The upper plot compares different cutoff frequencies under critical damping ($\zeta=1.0$). With the low cutoff frequency ($\omega_c=20~\mathrm{rad/s}$), the estimated force responds more slowly than the force sensor measurement, although the overall magnitude is comparable. With the middle cutoff frequency ($\omega_c=50~\mathrm{rad/s}$), the estimated force agrees well with the force sensor measurement over a wide range of the motion. However, when the applied force becomes large, the estimated force exceeds the force sensor measurement. This discrepancy suggests that the accuracy of the identified dynamics parameters and the compensation model is still insufficient, especially in high-force contact conditions.

The lower plot in Fig.~\ref{fig:obs_param_comp_force_sensor} compares different damping ratios with the middle cutoff frequency ($\omega_c=50~\mathrm{rad/s}$). In this experiment, the end-effector was brought into contact with the force sensor and then pressed against it continuously. The underdamped case ($\zeta=0.3$) shows large overshoot and oscillation in the estimated force. The Butterworth setting ($\zeta=1/\sqrt{2}\simeq0.707$) reduces the oscillation but still exhibits a small overshoot. Critical damping ($\zeta=1.0$) suppresses overshoot and provides an estimate that closely follows the force sensor measurement. The overdamped case ($\zeta=2.0$) also avoids overshoot, but its convergence is slower, and the corresponding velocity estimate becomes noticeably noisy.

\subsubsection{Tuning}
These results provide the following tuning guidelines. First, the damping ratio should not be chosen in the underdamped region because it can cause significant overshoot and oscillation, especially during contact. Excessively overdamped settings should also be avoided, as they degrade the position-tracking response and increase noise in the velocity estimate. In practice, reasonable damping ratios are limited to a narrow range between the Butterworth setting and critical damping. For contact-rich tasks, however, overshoot in the external force estimate is undesirable because it can produce an estimate larger than the force sensor measurement. Therefore, critical damping is a practical default choice, as it provides the fastest response without overshoot.

Once the damping ratio is fixed to critical damping, the remaining tuning parameter is the cutoff frequency. The cutoff frequency determines the trade-off between tracking performance and noise sensitivity. This trade-off is particularly important for low-cost manipulators equipped with low-resolution rotary encoders, where increasing the cutoff frequency improves the response but increases sensitivity to quantization noise.

\subsection{Teleoperation}
We implemented the proposed 4-channel bilateral teleoperation on CRANE-X7.
We compared three other teleoperation techniques and conducted two ablation studies.
We compared teleoperation performance on free motion by repeatedly swinging the joint closest to the base as quickly as possible at a 90-degree angle 10 times, and contact-rich motion by repeatedly wiping the whiteboard 5 times.
The controller parameters used in the experiment are shown in Table~\ref{table:controller_parameters}.
The results are shown in Fig.~\ref{fig:teleoperation_comparison}, Fig.~\ref{fig:teleoperation_comparison_wiping}, and the Mean Absolute Error (MAE) is shown in Table~\ref{table:MAE}.

\subsubsection{vs. Unilateral Control}
Unilateral control is a teleoperation method that controls the follower's position using the leader's position as the target.
The biggest drawback of unilateral control is the lack of force feedback to the operator, though position-tracking performance is also an issue.
Compared to the 4-channel type, which uses position and force control for both the leader and follower robots, unilateral control uses only position control for the follower, resulting in limited tracking performance at the same gain.
As a result, the position error exceeded that of the proposed method, especially during high-speed free motion.

\subsubsection{vs. Symmetric Position Type Bilateral Control}
Symmetric position-type bilateral control is a method in which both the leader and the follower control their positions based on each other's positions as targets.
It is known that this method makes the operation feel very heavy due to the force that keeps the current position unchanged.
In this case, since the manipulator was relatively lightweight, the heavy operation problem did not seem so serious.
On the other hand, there was a difference in the position tracking error.
In the 4-channel type, the large operating force applied to the leader when it changes direction is also applied to the follower via force control, enabling rapid directional changes. On the other hand, in a symmetric position type that relies solely on position control, large feedback was generated only when a large position error occurred, making it difficult to follow rapid directional changes.

\subsubsection{vs. Force Feedback Type Bilateral Control}
Force feedback type bilateral control is a method in which the follower side performs position error feedback and the leader side performs force error feedback.
Like unilateral control, this type also relies solely on the follower's position control to reduce position error.
Consequently, its position tracking performance was lower than that of the 4-channel type and symmetric position type at the same gain. 
However, due to the force control on the leader side, its tracking performance is slightly higher than that of unilateral control.

\subsubsection{vs. 4ch Bilateral Control using Fixed Inertia}
We compared the control performance of a system with fixed inertial parameters to that of a system that accounted for inertial variations.
While fixed-inertia (fixed-gain) controllers are simple to stabilize and widely used in manipulator control, their performance is posture-dependent and inconsistent across the workspace.
Especially when the cutoff frequency for external force estimation and its feedback is limited, inertial modeling errors lead to significant performance degradation. Our experimental results demonstrate that incorporating inertial variations significantly reduces position tracking errors.
The force error appeared to be smaller in the fixed-inertia case. However, this is likely a byproduct of the estimation process and does not reflect physical reality. Since the nominal inertias of the leader and follower were kept identical, the inertial modeling errors tended to be similar. Consequently, these errors were largely counterbalanced during the differential calculation, particularly at low frequencies. Meanwhile, the high-frequency components of the modeling error were effectively attenuated by the observer's low-pass filter. This combination results in an apparent suppression of the force error.

\subsubsection{vs. 4ch Bilateral Control using Phase-Lagged Velocity}
Velocity estimation based on numerical differentiation with a first-order low-pass filter exhibited oscillatory behavior. This behavior is attributed to the phase lag of the estimated velocity, which was further increased by the low cutoff frequency required to suppress encoder quantization noise. As indicated by the free-motion plots after task completion and the increased angular-velocity MAE in Table~\ref{table:MAE}, significant high-frequency oscillations occurred.
Although the phase-lagged velocity case yielded the smallest position MAE, this does not necessarily indicate superior practical performance. In the observer-based case, the lead-compensation effect shapes the effective velocity-feedback gain in a frequency-dependent manner, reducing its influence in the low-frequency range where disturbance feedback is active. This frequency-dependent reshaping can increase the position MAE while suppressing frequency-response peaking and high-frequency oscillatory behavior.

\subsection{Imitation Learning}

We collected demonstration data using the proposed 4-channel bilateral teleoperation method and trained imitation learning policies.
The snapshot of the tasks is shown in Fig.~\ref{fig:snapshot_IL}, and the results are shown in Table~\ref{table:success_rate_IL}.
We used Action Chunking with Transformer (ACT)~\cite{zhao2023learning,buamanee2024bi} for the neural network architecture.
We compare the cases of not using force, using force only for input, and using force for both input and output.
All demonstrations were collected by a single operator to reduce inter-operator variability and focus on the effect of force information in the demonstration data.
For each task and input-output condition, ACT was trained with a fixed random seed.
Therefore, the reported success rates should be interpreted as an application-level demonstration of the usefulness of force information, rather than a statistically exhaustive benchmark of imitation-learning performance.

Note that since the control system differs between data collection and autonomous operation, a decrease in success rate was expected when force information is not used in the policy output.
However, differences in control systems or dynamics between data collection and autonomous operation are not uncommon in imitation learning, especially when using data collection methods other than teleoperation.
Additionally, since the dataset's quality can significantly affect imitation learning results, it is preferable to compare using the same dataset.
Considering these points, we conducted this comparison.

\begin{figure}[t]
    \centering
    \includegraphics[width=\linewidth]{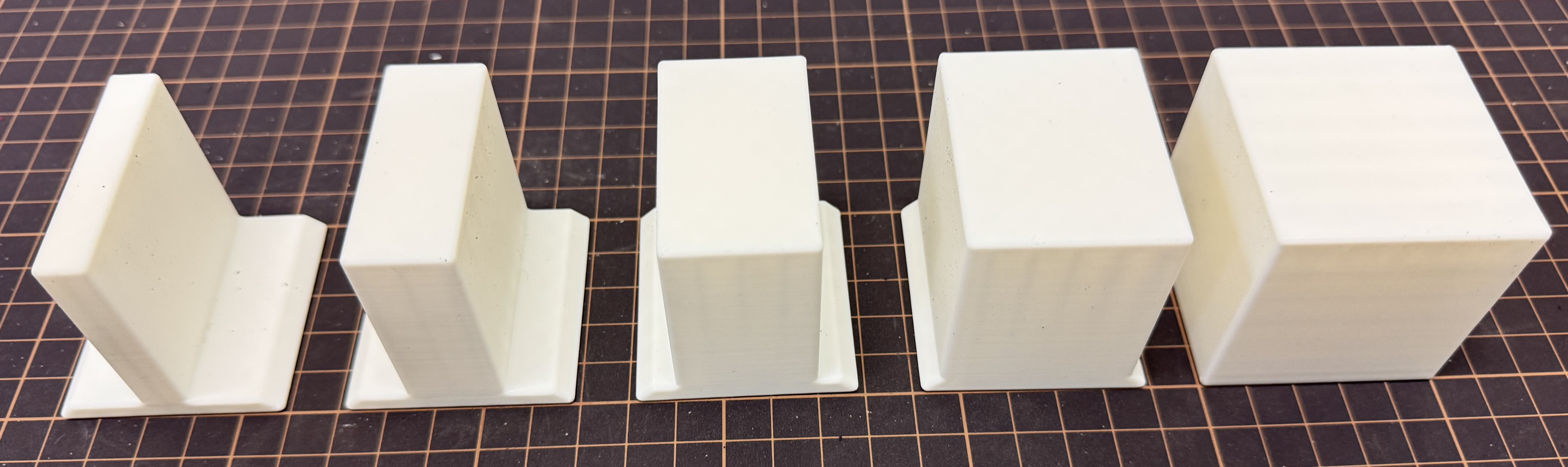}
    \caption{Objects for Dual-arm Pick-and-Place}
    \label{fig:object_pick_and_place}
\end{figure}

\begin{table*}[t]
    \caption{Success rate of imitation learning tasks by Action Chunking with Transformer}
    \label{table:success_rate_IL}
    \centering
    \begin{tabular}{lc|cc|cc|cc}
        \toprule
        & force input & \multicolumn{2}{c|}{-} & \multicolumn{2}{c|}{$\bigcirc$} & \multicolumn{2}{c}{$\bigcirc$} \\
        & force output & \multicolumn{2}{c|}{-} & \multicolumn{2}{c|}{-} & \multicolumn{2}{c}{$\bigcirc$} \\
        \midrule
        \midrule
        \multirow{7}{*}{Dual-arm Pick-and-Place}& block width & pick & place & pick & place & pick & place \\
        \cmidrule(lr){2-8}
        & 10 mm & 0/5 & 0/5 & \textbf{5/5} & 0/5 & \textbf{5/5} & \textbf{5/5} \\
        & 20 mm & 0/5 & 0/5 & \textbf{5/5} & \textbf{5/5} & \textbf{5/5} & \textbf{5/5} \\
        & 30 mm & \textbf{5/5} & 0/5 & \textbf{5/5} & \textbf{5/5} & \textbf{5/5} & \textbf{5/5} \\
        & 40 mm & \textbf{5/5} & 0/5 & \textbf{5/5} & 4/5 & \textbf{5/5} & \textbf{5/5} \\
        & 50 mm & \textbf{5/5} & 0/5 & \textbf{5/5} & 0/5 & \textbf{5/5} & \textbf{5/5} \\
        \cmidrule(lr){2-8}
        & Total & 15/25 & 0/25 & \textbf{25/25} & 14/25 & \textbf{25/25} & \textbf{25/25} \\
        \midrule
        Nut Turning && \multicolumn{2}{c|}{0/5} & \multicolumn{2}{c|}{\textbf{5/5}} & \multicolumn{2}{c}{\textbf{5/5}} \\
        Cucumber Peeling && \multicolumn{2}{c|}{0/5} & \multicolumn{2}{c|}{3/5} & \multicolumn{2}{c}{2/5} \\
        \bottomrule
    \end{tabular}
\end{table*}

\subsubsection{Task1: Dual-arm Pick-and-Place}
We performed a grasping task with blocks of different widths between the two arms.
The snapshot is shown in Fig.~\ref{fig:snapshot_dual_arm_pick_and_place}.
We prepared five types of blocks, ranging from 10 mm to 50 mm in width, and collected a total of 10 demonstration data sets, two for each type.
The figure of the blocks is shown in Fig.~\ref{fig:object_pick_and_place}.
In this experimental environment, the camera captured images from a bird's-eye view, making it difficult to determine the width of blocks based solely on the images.
Regarding success or failure, if the block was completely lifted off the table, the pick was considered successful, and if the block was placed without being dropped or knocked over, the placement was considered successful.

As a result, when no force information was used for either input or output, picking was not possible in cases with small widths of 10 mm and 20 mm. In cases with relatively large widths of 30 mm to 50 mm, picking was possible once, but problems such as immediate dropping of the object or the angle changing while holding it occurred, and the object could not be placed in the correct position.
When force information was used in the input, the success rate of picking improved significantly, with all trials resulting in successful picking. However, instability after picking persisted, and placement failures occurred frequently in cases where the width was farthest from the average (10 mm and 50 mm).
When both the input and the output used force information, both pick-and-place operations were successful in all trials. The stability of holding after picking improved, with no instances of dropping midway or misalignment.

\subsubsection{Task2: Nut Turning}
We tried the nut-turning task, which requires more force and speed.
The snapshot is shown in Fig.~\ref{fig:snapshot_nut_turning}.
We collected 10 demonstration data sets.
This task involved quickly rubbing a nut that was loosely attached to a screw with the robot's finger and turning it with momentum.
If the robot does not apply enough force and speed, the nut will not turn properly.
The motion is repeated until the nut reaches the screw head.
Success was defined as the nut reaching the screw head.

As a result, the policy failed when force information was included in neither the input nor the output. 
In contrast, when force information was included in the input, the task was successful regardless of whether force information was included in the output. 
Without force information in the input, the robot often stopped at a configuration where the gripper visually overlapped with the nut in the bird's-eye-view image. 
In this experimental setup, the bird's-eye-view images just before contact and during contact between the gripper, the nut, and the stage were visually very similar. 
Therefore, the policy could not reliably infer the contact state from visual observations alone and tended not to proceed further. 
This observation indicates that the failure was not simply due to poor position-control accuracy, but rather to partial observability of the contact state when force information was not provided. 
The force input provided essential contact-state information that was difficult to infer from the camera image, especially under the limited temporal context available to the ACT policy.

\subsubsection{Task3: Cucumber Peeling}
For a task involving the handling of irregularly shaped objects, we tried peeling cucumbers with a peeler.
The snapshot is shown in Fig.~\ref{fig:snapshot_cucumber_peeling}.
We collected 10 demonstration data sets.
First, the left arm grasps the cucumber placed in the center of the cutting board, then the right arm grabs the peeler located in the designated spot. Next, the peeler is pressed against the cucumber, and the right arm is quickly moved from left to right while applying downward pressure to peel the cucumber.
Success was defined as the cucumber skin being peeled off.

As a result, when no force information was used for input or output, the experiment was completely unsuccessful.
The failure occurred when the gripper overlapped with the peeler in the bird's-eye view image, similar to the failure of the nut turning task.
Using force information for input yielded success in three out of five cases, while using force information for both input and output yielded success in two out of five cases.
The main failure case occurred when the system transitioned to the finished state before touching the cucumber or peeler.
This is likely because there was not much difference between the initial image and the final image.

\section{Conclusion}

In this study, we demonstrated that practical high-speed bilateral teleoperation with force feedback is achievable on low-cost manipulators without force sensors by employing a sensorless 4-channel bilateral control framework. The proposed approach enables stable and accurate position-force interaction under severe sensing and bandwidth constraints, which are common limitations of low-cost robotic hardware.

The key contribution of this work lies in integrating nonlinear dynamics compensation with a disturbance-observer-based velocity and external force estimation framework, along with a clear frequency-domain interpretation of the observer structure. This analysis revealed the intrinsic coupling between the velocity and force estimation bandwidths and reduced the observer tuning freedom to a single cutoff frequency, providing practical, hardware-oriented parameter tuning guidelines suitable for low-cost implementations.

In the teleoperation experiments, the proposed sensorless 4-channel bilateral teleoperation method improved position-tracking accuracy compared with unilateral control, symmetric position control, force-feedback bilateral control, and conventional 4-channel bilateral control using simplified dynamics models. 
Although the phase-lagged velocity case, which used low-pass-filtered numerical differentiation instead of the observer-based velocity estimate, achieved lower position error than the observer-based case, it exhibited high-frequency oscillatory behavior in the position and velocity responses.
In contrast, the proposed velocity and disturbance observer suppressed this high-frequency oscillatory behavior while maintaining stable force-feedback teleoperation.

As an application, we further demonstrated that the proposed teleoperation system enables the collection of high-fidelity demonstration data and that incorporating force information significantly improves the success rate of imitation learning across multiple manipulation tasks. While force input was consistently beneficial, the effectiveness of force output was task-dependent.

Future work includes improving data collection and parameter identification by adopting more advanced identification techniques, extending the proposed framework to a wider range of manipulation tasks, and investigating more expressive learning architectures that can better exploit force-aware demonstrations.

\section*{APPENDIX}
\subsection{Implementation}
A CRANE-X7 manipulator (RT Corporation, Tokyo, Japan) was used in the experiments.
The manipulator has seven degrees of freedom (DoF), and the gripper adds one additional DoF, resulting in a total of eight DoF.
An Intel RealSense D435i RGB-D camera was used to capture RGB images.

The low-level control system was implemented in C++, with each manipulator controlled by a dedicated thread.
Periodic execution was implemented using timerfd, enabling waking up at an absolute time.
Each control thread was pinned to dedicated CPU cores and executed under a high-priority First-In, First-Out~(FIFO) real-time scheduling policy. The assigned cores were isolated from operating system interrupts, allowing a 1 kHz soft real-time control loop.

The manipulator and the camera were connected to a PC running Ubuntu 22.04 LTS.
The PC was equipped with 32 GB RAM, an AMD Ryzen 9 7950X 16 core CPU, and an NVIDIA GeForce RTX 4080 GPU.
The AMD Ryzen 9 7950X consists of multiple Core Complex Die (CCD) with separated L3 caches~\cite{amd_ryzen_software_optimization_gdc2023}.
By appropriately configuring CPU affinity, OS-related processes and real-time control threads can be physically separated at the cache level, reducing cache interference and execution jitter.

The imitation learning policy was implemented in Python, and inter-process communication between the Python policy and the C++ control system was implemented using shared memory.

\begin{figure*}[!t]
    \begin{minipage}[b]{0.33\linewidth}
        \centering
        \includegraphics[width=\linewidth]{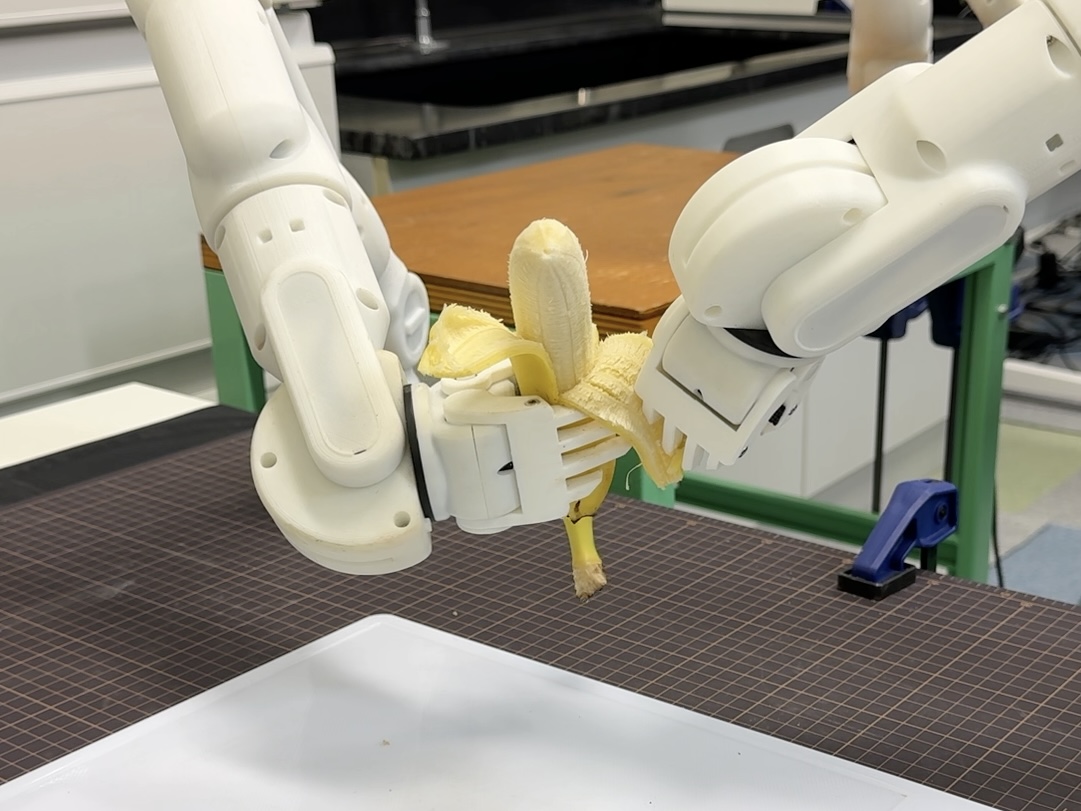}
        \subcaption{Banana Peeling}
    \end{minipage}
    \begin{minipage}[b]{0.33\linewidth}
        \centering
        \includegraphics[width=\linewidth]{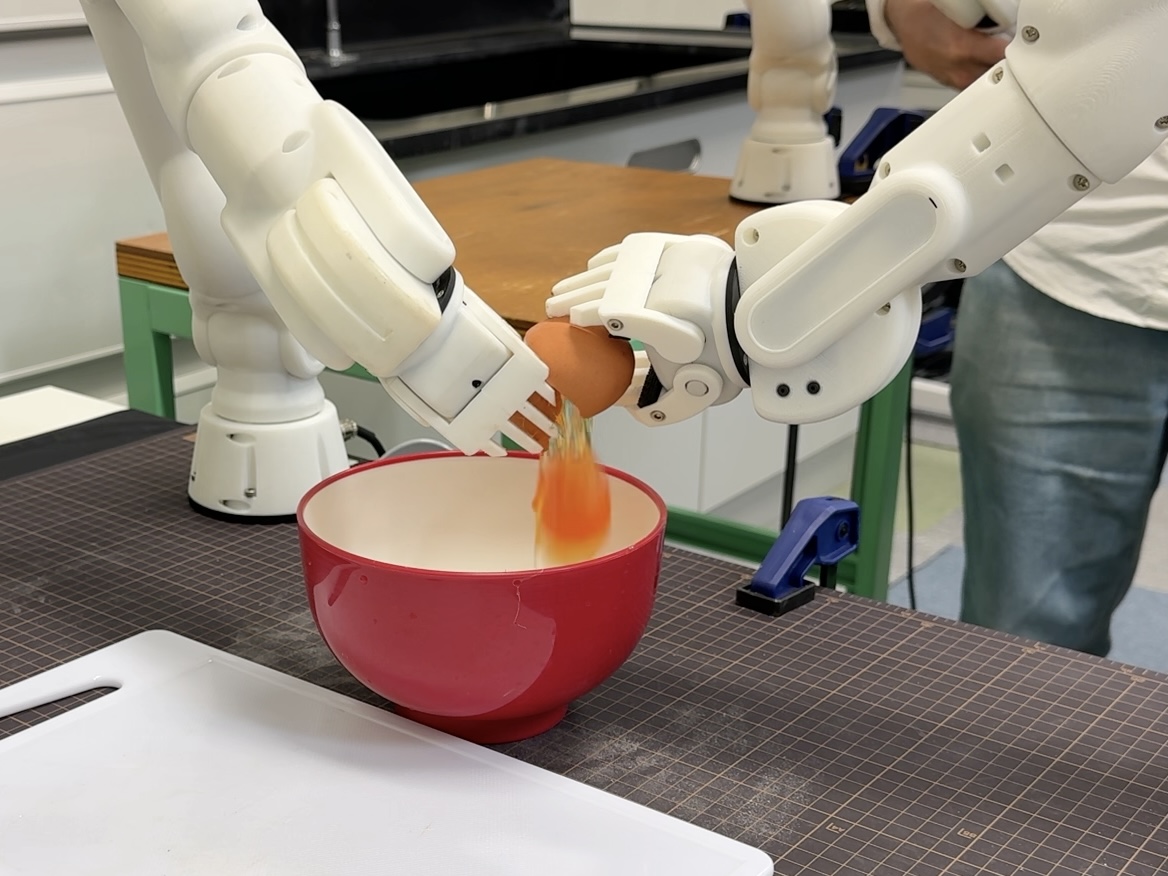}
        \subcaption{Egg Cracking}
    \end{minipage}
    \begin{minipage}[b]{0.33\linewidth}
        \centering
        \includegraphics[width=\linewidth]{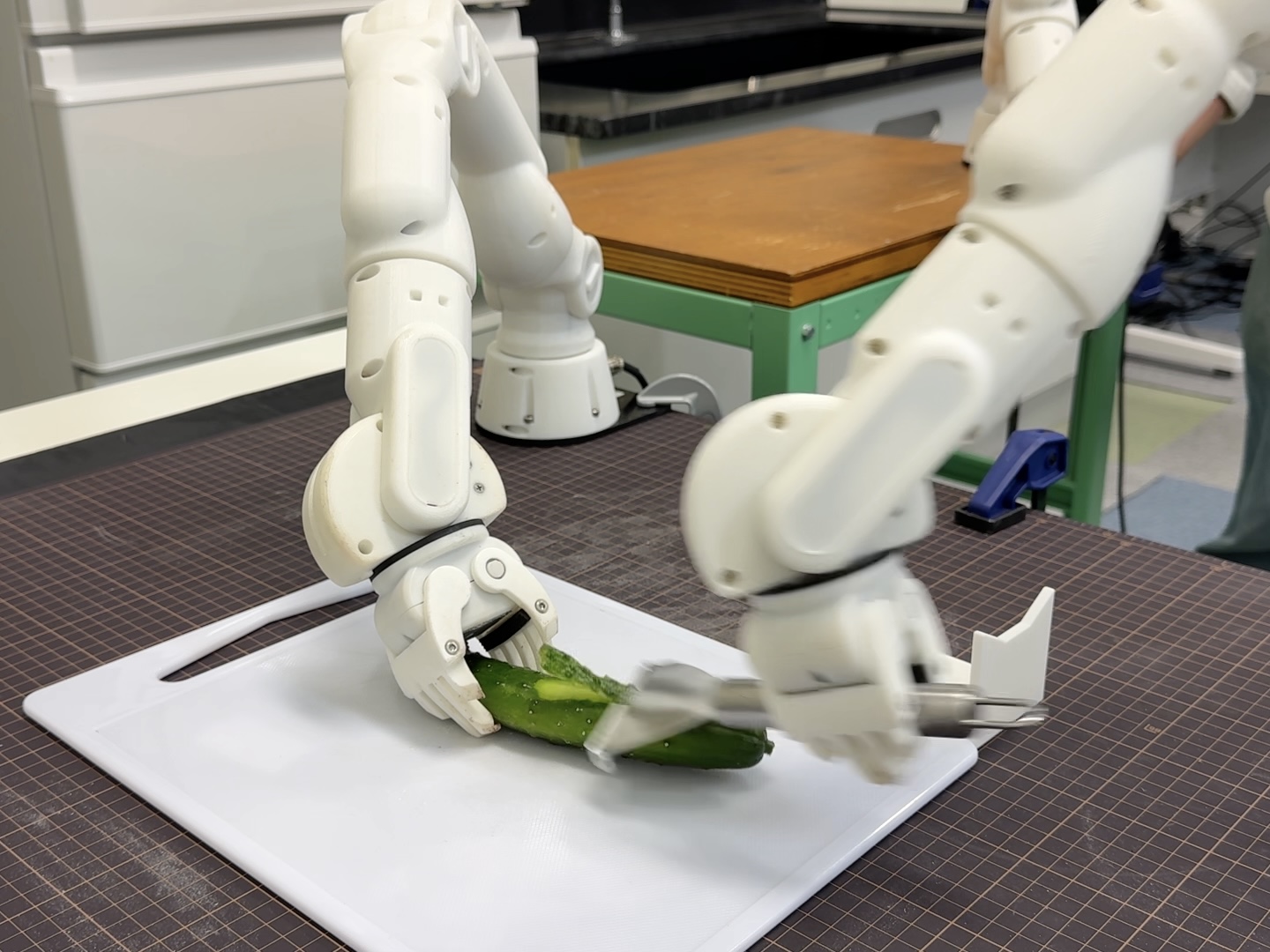}
        \subcaption{Cucumber Peeling}
    \end{minipage}
    \begin{minipage}[b]{0.33\linewidth}
        \centering
        \includegraphics[width=\linewidth]{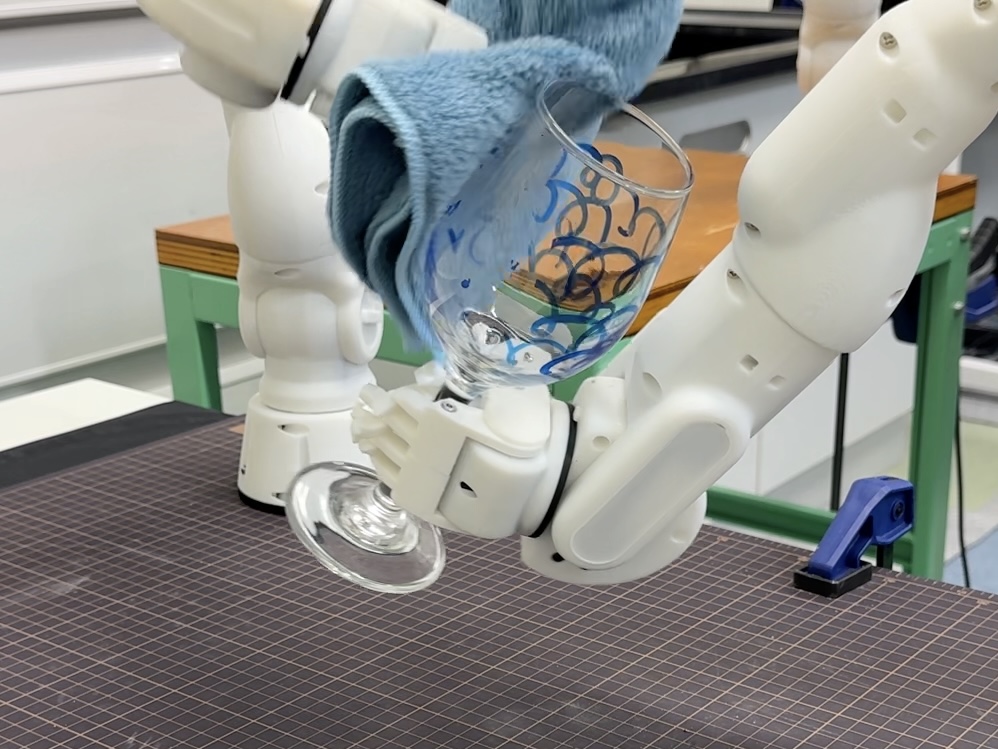}
        \subcaption{Wine Glass Wiping}
    \end{minipage}
    \begin{minipage}[b]{0.33\linewidth}
        \centering
        \includegraphics[width=\linewidth]{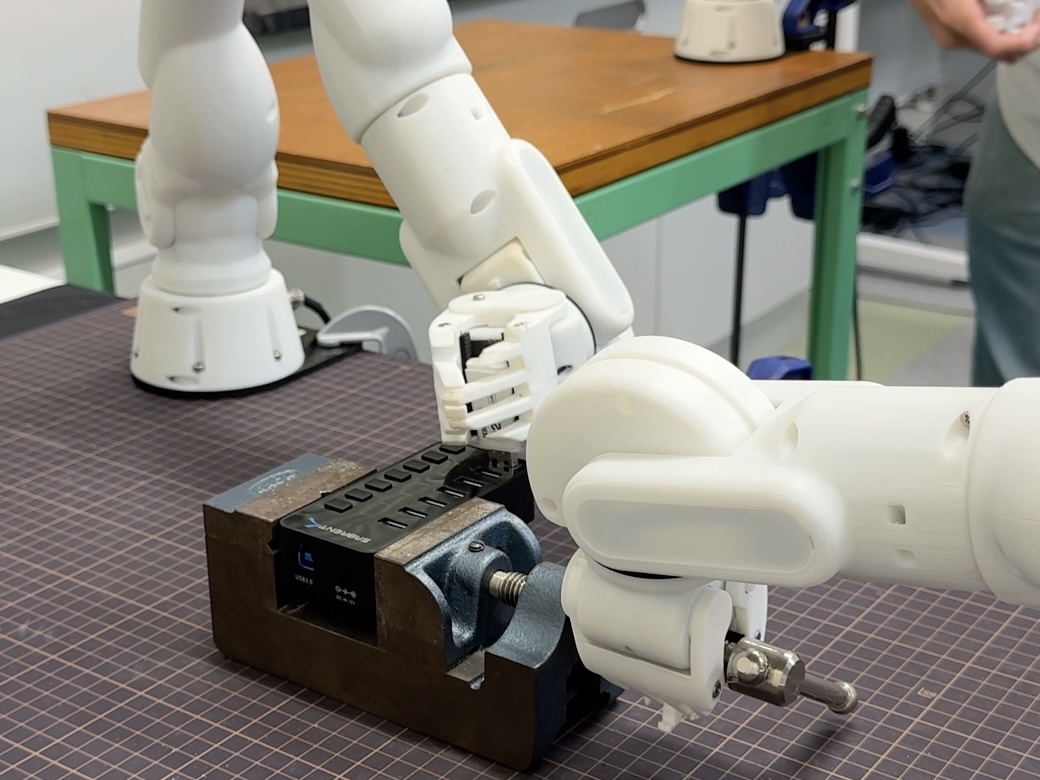}
        \subcaption{USB Insertion}
    \end{minipage}
    \begin{minipage}[b]{0.33\linewidth}
        \centering
        \includegraphics[width=\linewidth]{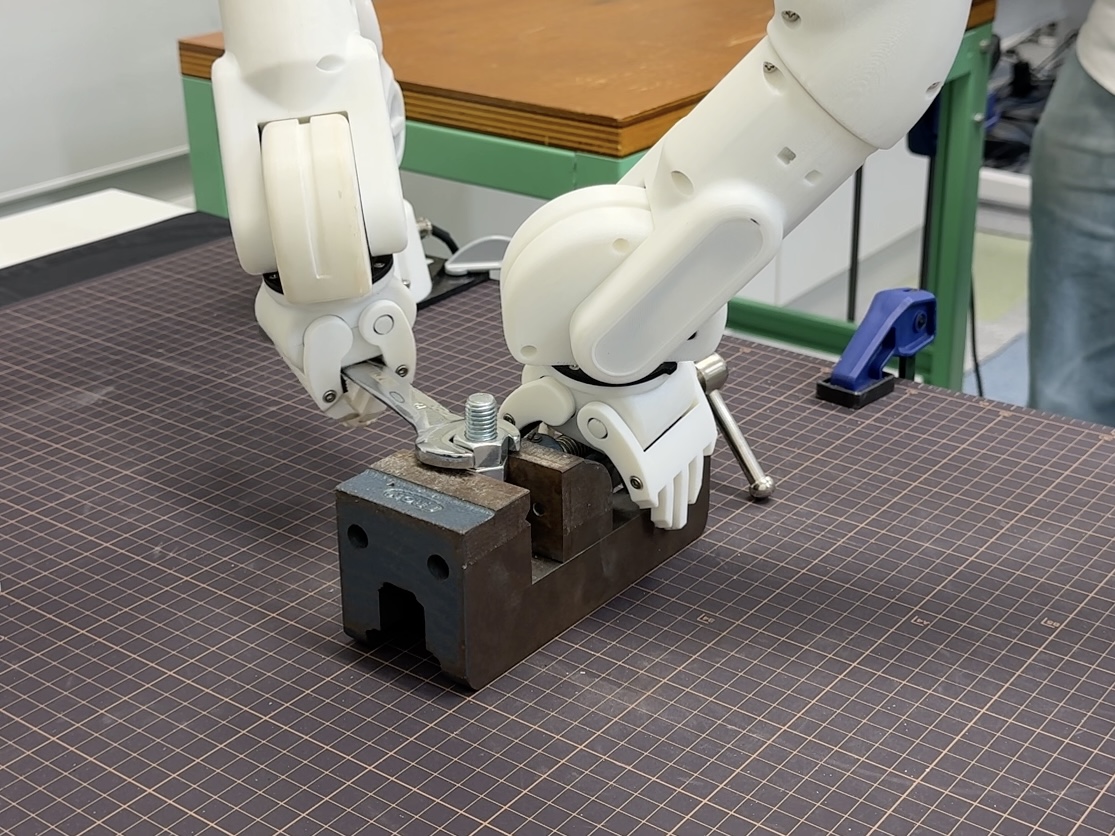}
        \subcaption{Nut Tightening}
    \end{minipage}
    \caption{Teleoperation examples}
    \label{fig:teleoperation_examples}
\end{figure*}

Discretization was performed using a bilinear transformation.
The pseudocode of the observer and controller is shown in algorithm~\ref{alg:vel_obs}, and examples of teleoperation tasks are shown in Fig.~\ref{fig:teleoperation_examples}.

\subsection{Derivation of the Modal Impedance Error}
\label{appendix:modal_impedance_derivation}

For the scalar single-axis case, let
\begin{equation}
Z_{\pm}=
\begin{bmatrix}
Z_{--} & Z_{-+}\\
Z_{+-} & Z_{++}
\end{bmatrix},
\qquad
J_{\pm}=
\begin{bmatrix}
1 & -1\\
1/2 & 1/2
\end{bmatrix}.
\end{equation}
From \(Z_{LF}=J_{\pm}^{\top}Z_{\pm}J_{\pm}\), the leader--follower impedance elements are obtained as
\begin{align}
Z_{ll} &= Z_{--}+\frac{1}{2}(Z_{-+}+Z_{+-})+\frac{1}{4}Z_{++},\\
Z_{lf} &= -Z_{--}+\frac{1}{2}(Z_{-+}-Z_{+-})+\frac{1}{4}Z_{++},\\
Z_{fl} &= -Z_{--}-\frac{1}{2}(Z_{-+}-Z_{+-})+\frac{1}{4}Z_{++},\\
Z_{ff} &= Z_{--}-\frac{1}{2}(Z_{-+}+Z_{+-})+\frac{1}{4}Z_{++}.
\end{align}
For compactness, define the diagonal and off-diagonal modal impedance combinations as
\begin{align}
Z_{diag+} &:= Z_{--}+\frac{1}{4}Z_{++},&
Z_{diag-} &:= -Z_{--}+\frac{1}{4}Z_{++},\\
Z_{off+} &:= \frac{1}{2}(Z_{-+}+Z_{+-}),&
Z_{off-} &:= \frac{1}{2}(Z_{-+}-Z_{+-}).
\end{align}
Here, the subscripts \(d\) and \(o\) denote diagonal and off-diagonal modal impedance combinations, respectively, and \(+\) and \(-\) denote sum- and difference-type combinations. Using these definitions, the leader--follower impedance elements can be written as
\begin{align}
Z_{ll} &= Z_{diag+}+Z_{off+},\\
Z_{lf} &= Z_{diag-}+Z_{off-},\\
Z_{fl} &= Z_{diag-}-Z_{off-},\\
Z_{ff} &= Z_{diag+}-Z_{off+}.
\end{align}
The displayed-impedance error is
\begin{equation}
\Delta Z_{l\leftarrow f}
=
-Z_{env}
+Z_{ll}
-Z_{lf}(Z_{ff}+Z_{env})^{-1}Z_{fl}.
\end{equation}
Because all quantities are scalar in this case, this expression can be rewritten as
\begin{equation}
\Delta Z_{l\leftarrow f}
=
\frac{
(Z_{ll}-Z_{env})(Z_{ff}+Z_{env})
-
Z_{lf}Z_{fl}
}{
Z_{ff}+Z_{env}
}.
\end{equation}
The denominator is
\begin{equation}
\begin{aligned}
D_\Delta
&=
Z_{ff}+Z_{env}
=
Z_{diag+}
-
Z_{off+}
+
Z_{env}
\\&=
Z_{--}
+
\frac{1}{4}Z_{++}
+
Z_{env}
-
\frac{1}{2}(Z_{-+}+Z_{+-})
.
\end{aligned}
\end{equation}
The product \(Z_{lf}Z_{fl}\) is
\begin{align}
Z_{lf}Z_{fl}
&=
(Z_{diag-}+Z_{off-})
(Z_{diag-}-Z_{off-})\\
&=
Z_{diag-}^{2}
-
Z_{off-}^{2}.
\end{align}
Similarly,
\begin{equation}
\begin{aligned}
&(Z_{ll}-Z_{env})(Z_{ff}+Z_{env}) \\
&=
(Z_{diag+}+Z_{off+}-Z_{env})
(Z_{diag+}-Z_{off+}+Z_{env})\\
&=
Z_{diag+}^{2}
-
(Z_{off+}-Z_{env})^2\\
&=
Z_{diag+}^{2}
-
Z_{off+}^{2}
+
2Z_{env}Z_{off+}
-
Z_{env}^2.
\end{aligned}
\end{equation}
Therefore, the numerator becomes
\begin{equation}
\begin{aligned}
N_\Delta
&=
(Z_{ll}-Z_{env})(Z_{ff}+Z_{env})
-
Z_{lf}Z_{fl}
\\
&=
\left(
Z_{diag+}^{2}
-
Z_{diag-}^{2}
\right)
-
\left(
Z_{off+}^{2}
-
Z_{off-}^{2}
\right)
+
2Z_{env}Z_{off+}
-
Z_{env}^2.
\end{aligned}
\end{equation}
The diagonal and off-diagonal combinations satisfy
\begin{align}
Z_{diag+}^{2}-Z_{diag-}^{2}
&=
Z_{--}Z_{++},\\
Z_{off+}^{2}-Z_{off-}^{2}
&=
Z_{-+}Z_{+-},\\
2Z_{env}Z_{off+}
&=
Z_{env}(Z_{-+}+Z_{+-}).
\end{align}
Thus, the numerator is obtained as
\begin{equation}
N_\Delta
=
Z_{--}Z_{++}
-
Z_{env}^2
+
Z_{env}(Z_{-+}+Z_{+-})
-
Z_{-+}Z_{+-}.
\end{equation}

\subsection{Derivation of the Frequency-Domain Observer Representation}
\label{appendix:observer_frequency_derivation}

This appendix derives the frequency-domain representation of the velocity and external force observer.
The Laplace transform is applied to the observer dynamics written in acceleration-domain variables.
Define
\begin{equation}
\bm u_a := \tilde{\bm M}(\bm q)^{-1}\bm\tau_u,
\qquad
\hat{\bm d} := \tilde{\bm M}(\bm q)^{-1}
\hat{\bm\tau}_{ext}.
\end{equation}
Here, \(\bm u_a\) is the nominal acceleration input and
\(\hat{\bm d}\) is the acceleration-domain disturbance estimate.
Thus, the configuration-dependent inertia matrix is included in the transformed signals rather than treated as a constant plant parameter.

Using these variables, the state-space form of the observer dynamics is
\begin{equation}
\begin{aligned}
\frac{d}{dt}
\begin{bmatrix}
\hat{\dot{\bm q}} \\
\hat{\bm d}
\end{bmatrix}
&=
\begin{bmatrix}
\bm 0 & \bm I \\
\bm 0 & \bm 0
\end{bmatrix}
\begin{bmatrix}
\hat{\dot{\bm q}} \\
\hat{\bm d}
\end{bmatrix}
+
\begin{bmatrix}
\bm I \\
\bm 0
\end{bmatrix}
\bm u_a  \\
&\quad+
\begin{bmatrix}
\bm H_1 \\
\bm H_2
\end{bmatrix}
(\dot{\bm q}-\hat{\dot{\bm q}}).
\end{aligned}
\label{eq:vfob_appendix}
\end{equation}
Applying the Laplace transform to~\eqref{eq:vfob_appendix} gives
\begin{align}
s\hat{\dot{\bm q}}
&=
\hat{\bm d}
+
\bm u_a
+
\bm H_1(s\bm q-\hat{\dot{\bm q}}),
\label{eq:obs_1st}\\
s\hat{\bm d}
&=
\bm H_2(s\bm q-\hat{\dot{\bm q}}),
\label{eq:obs_2nd}
\end{align}
where \(\bm H_1\) and \(\bm H_2\) are the observer gain matrices.
From~\eqref{eq:obs_1st},
\begin{align}
(s\bm I+\bm H_1)\hat{\dot{\bm q}}
&=
\hat{\bm d}
+
\bm u_a
+
\bm H_1s\bm q,\\
\hat{\dot{\bm q}}
&=
(s\bm I+\bm H_1)^{-1}
(\hat{\bm d}+\bm u_a)
\nonumber\\
&\quad+
(s\bm I+\bm H_1)^{-1}
\bm H_1s\bm q .
\end{align}
Using
\begin{equation}
\ddot{\bm q}_{ref}
:=
\bm u_a+\hat{\bm d}
=
\tilde{\bm M}(\bm q)^{-1}
(\bm\tau_u+\hat{\bm\tau}_{ext}),
\end{equation}
the velocity estimate is written as
\begin{equation}
\hat{\dot{\bm q}}
=
(s\bm I+\bm H_1)^{-1}
\ddot{\bm q}_{ref}
+
(s\bm I+\bm H_1)^{-1}
\bm H_1s\bm q .
\end{equation}

Next, substituting the velocity estimate into~\eqref{eq:obs_2nd} gives
\begin{align}
&\{s\bm I+\bm H_2(s\bm I+\bm H_1)^{-1}\}\hat{\bm d}
\nonumber\\
&=
\bm H_2
\left[
\{\bm I-(s\bm I+\bm H_1)^{-1}\bm H_1\}
s\bm q
\right.
\nonumber\\
&\qquad\left.
-(s\bm I+\bm H_1)^{-1}\bm u_a
\right].
\end{align}
Therefore,
\begin{align}
\hat{\bm d}
&=
\{s\bm I+\bm H_2(s\bm I+\bm H_1)^{-1}\}^{-1}
\bm H_2
\nonumber\\
&\quad\cdot
\left[
\{\bm I-(s\bm I+\bm H_1)^{-1}\bm H_1\}
s\bm q
\right.
\nonumber\\
&\qquad\left.
-(s\bm I+\bm H_1)^{-1}\bm u_a
\right].
\end{align}

When the observer gains are parameterized by the cutoff angular frequency \(\omega_c\) and damping ratio \(\zeta\) as
\begin{equation}
\bm H_1=2\zeta\omega_c\bm I,
\qquad
\bm H_2=\omega_c^2\bm I,
\end{equation}
the velocity estimate becomes
\begin{equation}
\hat{\dot{\bm q}}
=
\frac{s}{s+2\zeta\omega_c}
\frac{1}{s}
\ddot{\bm q}_{ref}
+
\frac{2\zeta\omega_c}{s+2\zeta\omega_c}
s\bm q .
\end{equation}
The disturbance estimate becomes
\begin{align}
\hat{\bm d}
&=
\left(
s+\frac{\omega_c^2}{s+2\zeta\omega_c}
\right)^{-1}
\omega_c^2
\nonumber\\
&\quad\cdot
\left[
\frac{s}{s+2\zeta\omega_c}s\bm q
-
\frac{1}{s+2\zeta\omega_c}\bm u_a
\right]\\
&=
\frac{\omega_c^2}
{s^2+2\zeta\omega_cs+\omega_c^2}
(s^2\bm q-\bm u_a).
\end{align}
Finally, using
\(\hat{\bm\tau}_{ext}
=\tilde{\bm M}(\bm q)\hat{\bm d}\), we obtain
\begin{equation}
\hat{\bm\tau}_{ext}
=
\tilde{\bm M}(\bm q)
\frac{\omega_c^2}
{s^2+2\zeta\omega_cs+\omega_c^2}
\left(
s^2\bm q
-
\tilde{\bm M}(\bm q)^{-1}\bm\tau_u
\right).
\end{equation}

\subsection{Cross Structure Gripper}
\begin{figure}[t]
    \centering
    \includegraphics[width=\linewidth]{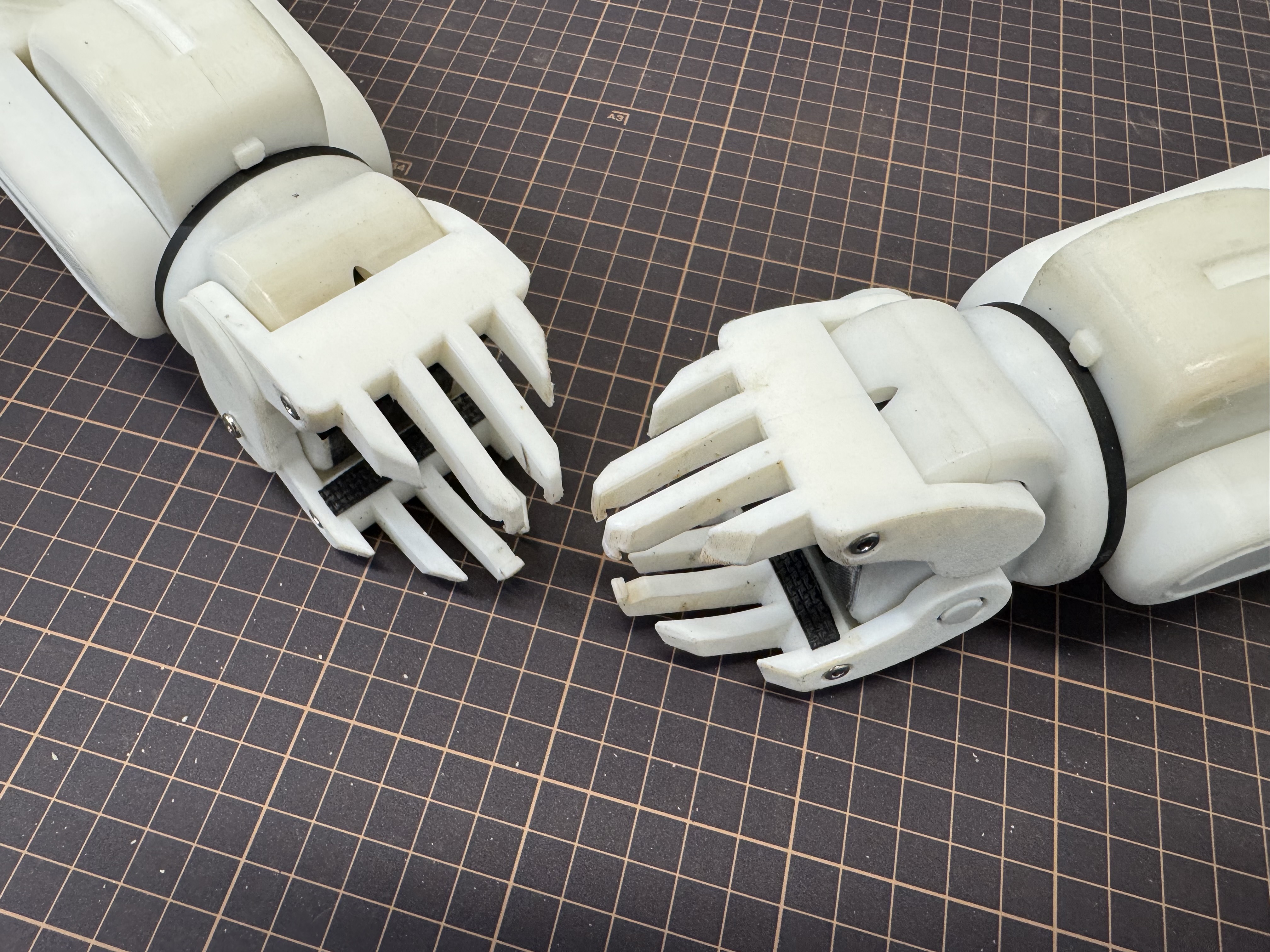}
    \caption{Cross Structure Gripper}
    \label{fig:cross_structure_hand}
\end{figure}

We used an improved version of the Cross Structure Gripper,
which we referred to as a cross-structure hand in our previous work~\cite{yamane2023soft}.
The figure is shown in Fig.~\ref{fig:cross_structure_hand}.
The Cross Structure Gripper is a simple one-degree-of-freedom rotating hand, but thanks to its crossed structure, it can close down to the base of the fingers, enabling powerful grasping of tools, in particular.
The hand used in this experiment has a narrower finger spacing of 5 mm, which allows it to grip even more delicate objects.
In addition, the hand's shape, which narrows towards the tip, allows it to apply pinpoint force when necessary and reduces interference between the hands during dual-arm tasks.

\subsection{Parameter Identification}
The rigid serial link model can be converted to linear equations by performing a variable transformation.
The matrix obtained by transforming position, velocity, and acceleration is referred to as the regressor matrix, and the corresponding constant parameters are termed parameter vectors.
The parameter vector can be estimated using linear least squares.
The regressor matrix~$\bm{Y}(\bm{q}, \dot{\bm{q}}, \ddot{\bm{q}})$ and the parameter vector~$\bm{\phi}$ of the manipulator dynamics are written as follows:
\begin{align}
\bm{\tau}
&= \bm{M}(\bm{q})\ddot{\bm{q}} + \bm{C}(\bm{q}, \dot{\bm{q}})\dot{\bm{q}} + \bm{D}\dot{\bm{q}} + \bm{g}(\bm{q}) \notag \\ 
&= \bm{Y}(\bm{q}, \dot{\bm{q}}, \ddot{\bm{q}}) \bm{\phi}
.
\end{align}
In this paper, we utilize the current control mode of the DYNAMIXEL motor and assume that its bandwidth is sufficiently high.
When the external force $\bm{\tau}_{ext}$ is zero, the estimated values of the parameter vector~$\bm{\phi}$ can be obtained using the linear least squares method as follows:
\begin{align}
\hat{\bm{\phi}} &= (\bm{Y}^{\top}\bm{Y})^{-1}\bm{Y}^{\top} \bm{\tau}
.
\end{align}
Note that the regressor matrix~$\bm{Y}(\bm{q}, \dot{\bm{q}}, \ddot{\bm{q}})$ and parameter vector~$\bm{\phi}$ are arbitrary, and the parameter vector is an abstract parameter that is a combination of various physical parameters, such as weight and link length, and therefore does not have a clear physical meaning.

We used data collected by unilateral teleoperation for parameter identification.
Mathematically, the rank of the regressor matrix should be kept as high as possible.
Therefore, the input is typically designed using a target trajectory that minimizes the drop in rank.
However, in reality, only a portion of the full space of position, velocity, and acceleration is used for teleoperation, and the accuracy of areas rarely used in human operations is not essential.
Teleoperated data encompasses many elements of human operation, making it one of the most efficient methods for identifying parameters of manipulator dynamics in teleoperation.
However, teleoperation data depends on the operator and is difficult to reproduce.
Autonomous data collection for identification is one of our important future works.

We implemented MATLAB scripts for identification based on OpenSYMORO~\cite{khalil2014OpenSYMORO}. We then translated OpenSYMORO's manipulator dynamics code into C++ and implemented the controller in C++ using the identified parameters.
The identified parameters are shown in Table~\ref{table:identified_parameters}.
The data from unilateral teleoperation are sampled at 500 Hz and downsampled to 25 Hz for parameter identification using the ``resample'' function in MATLAB.
Although the inertia matrix has nondiagonal elements in the identification result, they are set to 0 in the controller implementation because they are tiny compared to the diagonal elements in practice.

\begin{table}[t]
    \caption{
        \textbf{Identified parameters of rigid serial link model.}
        These parameters are elements of the parameter vector $\bm{\phi}$ of the regressor matrix form ($\bm{\tau}= \bm{Y}(\bm{q}, \dot{\bm{q}}, \ddot{\bm{q}}) \bm{\phi}$) and include abstract values that are products of multiple physical parameters, such as weights and link lengths.
        Furthermore, since unidentifiable parameters are not included, not all parameters of the original dynamic equation are included.
        The current-to-torque conversion constants used in this study were 2.3179 Nm/A for the XM430-W350 actuators and 3.1317 Nm/A for the XM540-W270 actuator.
    }
    \label{table:identified_parameters}
    \centering
    \begin{tabular}{l|lrc}
        \toprule
         &  & Value & Unit \\
        \midrule
        \multirow{12}{*}{\rotatebox{90}{gravity parameters}}
        & MX2 & -0.0095784 & $\rm{kg \cdot m}$ \\
        & MYR2 & -0.2140494 & $\rm{kg \cdot m}$ \\
        & MX3 & 0.0164795 & $\rm{kg \cdot m}$ \\
        & MYR3 & -0.0015841 & $\rm{kg \cdot m}$ \\
        & MX4 & 0.0112601 & $\rm{kg \cdot m}$ \\
        & MYR4 & -0.1269891 & $\rm{kg \cdot m}$ \\
        & MX5 & 0.0011854 & $\rm{kg \cdot m}$ \\
        & MYR5 & 0.0006837 & $\rm{kg \cdot m}$ \\
        & MX6 & -0.0049209 & $\rm{kg \cdot m}$ \\
        & MYR6 & -0.0051238 & $\rm{kg \cdot m}$ \\
        & MX7 & 0.0003040 & $\rm{kg \cdot m}$ \\
        & MY7 & 0.0002715 & $\rm{kg \cdot m}$ \\
        \midrule
        \multirow{12}{*}{\rotatebox{90}{inertia parameters}}
        & ZZR1 & 0.0040049 & $\rm{kg \cdot m^2}$ \\
        & XXR2 & 0.0447190 & $\rm{kg \cdot m^2}$ \\
        & ZZR2 & 0.0695762 & $\rm{kg \cdot m^2}$ \\
        & XXR3 & 0.0018078 & $\rm{kg \cdot m^2}$ \\
        & ZZR3 & 0.0010000 & $\rm{kg \cdot m^2}$ \\
        & XXR4 & 0.0204158 & $\rm{kg \cdot m^2}$ \\
        & ZZR4 & 0.0160292 & $\rm{kg \cdot m^2}$ \\
        & XXR5 &-0.0006468 & $\rm{kg \cdot m^2}$ \\
        & ZZR5 & 0.0001000 & $\rm{kg \cdot m^2}$ \\
        & XXR6 & 0.0008617 & $\rm{kg \cdot m^2}$ \\
        & ZZR6 & 0.0011530 & $\rm{kg \cdot m^2}$ \\
        & XXR7 & -0.0007504 & $\rm{kg \cdot m^2}$ \\
        & ZZ7 & 0.0001000 & $\rm{kg \cdot m^2}$ \\
        \midrule
        \multirow{6}{*}{\rotatebox{90}{motor inertia}}
        & IA3 & 0.0056659 & $\rm{kg \cdot m^2}$ \\
        & IA4 & 0.0159844 & $\rm{kg \cdot m^2}$ \\
        & IA5 & 0.0044899 & $\rm{kg \cdot m^2}$ \\
        & IA6 & 0.0054869 & $\rm{kg \cdot m^2}$ \\
        & IA7 & 0.0042852 & $\rm{kg \cdot m^2}$ \\
        & IA8 & 0.0053280 & $\rm{kg \cdot m^2}$ \\
        \midrule
        \multirow{8}{*}{\rotatebox{90}{viscous friction}}
        & FV1 & 0.0510939 & $\rm{Nm \cdot s/rad}$ \\
        & FV2 & 0.0888340 & $\rm{Nm \cdot s/rad}$ \\
        & FV3 & 0.0214482 & $\rm{Nm \cdot s/rad}$ \\
        & FV4 & 0.0761949 & $\rm{Nm \cdot s/rad}$ \\
        & FV5 & 0.0290511 & $\rm{Nm \cdot s/rad}$ \\
        & FV6 & 0.0400000 & $\rm{Nm \cdot s/rad}$ \\
        & FV7 & 0.0299360 & $\rm{Nm \cdot s/rad}$ \\
        & FV8 & 0.0000000 & $\rm{Nm \cdot s/rad}$ \\
        \bottomrule
    \end{tabular}
\end{table}

\begin{algorithm*}
\caption{Velocity and Disturbance Observer and Controller}
\label{alg:vel_obs}
\KwIn{Desired states $\bm{q}_{d}[k],\dot{\bm{q}}_{d}[k],\bm{\tau}_{d}[k]$, identified inertia matrix $\tilde{\bm{M}}(\bm{q}[k])$ and nonlinear term $\tilde{\bm{h}}(\bm{q}[k], \hat{\dot{\bm{q}}}[k])$ (including centrifugal, Coriolis, gravity and friction), cutoff angular frequency $\omega_c$, damping ratio $\zeta$, controller gain $\bm{K}_p, \bm{K}_d, \bm{K}_f$, torque limit $\bm{\tau}_{limit}$}
\KwOut{Estimated states $\hat{\dot{\bm{q}}}[k]$, $\hat{\bm{\tau}}_{ext}[k]$, torque command $\bm{\tau}_{cmd}$}

\BlankLine
\tcc{ --- Initialization ---}
$\bm{q}[0] \gets \mathrm{readPosition()}$ \tcp*[r]{Receive joint angle from rotary encoder}

$\ddot{\bm{q}}_{ref}[0], \ddot{\bm{q}}_{ref}^{lpf}[0], \bm{q}^{lpf}[0], \dot{\bm{q}}_{pred}[0] \gets \bm{0}$\;
$k \gets 1$\;

\BlankLine
\While{true}{
    $\bm{q}[k] \gets \mathrm{readPosition()}$ \tcp*[r]{Receive joint angle from rotary encoder}
    $T_s \gets \mathrm{measuredSamplingPeriod()}$ \tcp*[r]{cycle-to-cycle measured sampling period}
    
    \BlankLine
    \BlankLine
    \BlankLine
    \tcc{ --- Velocity Estimation ---}
    \BlankLine
    \tcp{Discrete-time pole of 1st-order LPF}
    \tcp{Calculated each period to handle jitter}
    $\alpha \gets \frac{2 - (2\zeta\omega_c) T_s}{2 + (2\zeta\omega_c) T_s}$  \tcp*[r]{Bilinear transform}
    \BlankLine
    $\ddot{\bm{q}}_{ref}^{lpf}[k] \gets \alpha \cdot \ddot{\bm{q}}_{ref}^{lpf}[k-1] + (1-\alpha) \cdot \left(\ddot{\bm{q}}_{ref}[k] + \ddot{\bm{q}}_{ref}[k-1]\right)/2$  \tcp*[r]{1st-order LPF}
    \BlankLine
    \tcp{Filtered numerical differentiation of joint angle}
    $\bm{q}^{lpf}[k] \gets \alpha \cdot \bm{q}^{lpf}[k-1] + (1-\alpha) \cdot \left(\bm{q}[k] + \bm{q}[k-1]\right)/2$  \tcp*[r]{1st-order LPF}
    $\dot{\bm{q}}^{lpf}[k] \gets 2\zeta\omega_c \cdot ( \bm{q}[k] - \bm{q}^{lpf}[k])$ \tcp*[r]{$\frac{2\zeta\omega_c}{s+2\zeta\omega_c}s = 2\zeta\omega_c\left(1 - \frac{2\zeta\omega_c}{s+2\zeta\omega_c}\right)$}
    \BlankLine
    \tcp{Complementary filter (see Eq.\ref{eq:vel_obs})}
    \tcp{$\hat{\dot{\bm{q}}}=\frac{s}{s+2\zeta\omega_c}\frac{1}{s}\ddot{\bm{q}}_{ref} + \frac{2\zeta\omega_c}{s+2\zeta\omega_c}s\bm{q}=\frac{1}{2\zeta\omega_c}\frac{2\zeta\omega_c}{s+2\zeta\omega_c}\ddot{\bm{q}}_{ref} + \frac{2\zeta\omega_c}{s+2\zeta\omega_c}s\bm{q}$}
    $\hat{\dot{\bm{q}}}[k] \gets \ddot{\bm{q}}_{ref}^{lpf}[k] /(2\zeta\omega_c) + \dot{\bm{q}}^{lpf}[k]$  \tcp*[r]{Velocity estimate}

    \BlankLine
    \BlankLine
    \BlankLine
    \tcc{ --- External Force Estimation ---}
    \BlankLine
    \tcp{Velocity prediction from integral of acceleration reference $\left(\frac{1}{s}\frac{2\zeta\omega_c}{s+2\zeta\omega_c}\ddot{\bm{q}}_{ref}\right)$}
    $\dot{\bm{q}}_{pred}[k] \gets \dot{\bm{q}}_{pred}[k-1] + \left(\ddot{\bm{q}}_{ref}^{lpf}[k]+\ddot{\bm{q}}_{ref}^{lpf}[k-1]\right)/2 \cdot T_s$  \tcp*[r]{Trapezoidal integral}
    \BlankLine
    \tcp{Disturbance estimate in the acceleration domain $\hat{\bm{d}}$ (see Eq.\ref{eq:disturbance},\ref{eq:tau_ext_obs},\ref{eq:acceleration_integral_control})}
    \tcp{Rewriting Eq.\ref{eq:tau_ext_obs} using $\hat{\bm{d}}=\tilde{\bm{M}}(\bm{\theta})^{-1}\hat{\bm{\tau}}_{ext}$ and  $\ddot{\bm{q}}_{ref}=\tilde{\bm{M}}(\bm{\theta})^{-1}(\bm{\tau}_u+\bm{\tau}_{ext})=\tilde{\bm{M}}(\bm{\theta})^{-1}\bm{\tau}_u+\hat{\bm{d}}$:}
    \tcp{$\hat{\bm{d}}=\frac{\omega_c^2}{s^2+2\zeta\omega_cs+\omega_c^2}\left\{s^2\bm{q}-\left(\ddot{\bm{q}}_{ref}-\hat{\bm{d}}\right)\right\}=\frac{s^2+2\zeta\omega_cs+\omega_c^2}{s^2+2\zeta\omega_cs}\frac{\omega_c^2}{s^2+2\zeta\omega_cs+\omega_c^2}\left(s^2\bm{q}-\ddot{\bm{q}}_{ref}\right)=\frac{\omega_c}{2\zeta}\frac{2\zeta\omega_c}{s+2\zeta\omega_c}\left(s\bm{q}-\frac{1}{s}\ddot{\bm{q}}_{ref}\right)$}
    $\hat{\bm{d}}[k] \gets \omega_c/(2\zeta) \cdot \left(\dot{\bm{q}}^{lpf}[k] - \dot{\bm{q}}_{pred}[k]\right)$\;
    \BlankLine
    $\hat{\bm{\tau}}_{ext}[k] \gets \tilde{\bm{M}}(\bm{q}[k])\hat{\bm{d}}$  \tcp*[r]{External force estimate}

    \BlankLine
    \BlankLine
    \BlankLine
    \tcc{ --- Controller ---}
    $\ddot{\bm{q}}_{ref}[k+1] \gets \bm{K}_p(\bm{q}_{d}[k] - \bm{q}[k]) + \bm{K}_d(\dot{\bm{q}}_{d}[k] - \hat{\dot{\bm{q}}}[k]) + \bm{K}_f(\bm{\tau}_{d}[k] + \hat{\bm{\tau}}_{ext}[k])$\;
    \BlankLine
    \tcp{Computed torque method with disturbance compensation}
    $\bm{\tau}_{cmd}[k+1] \gets \tilde{\bm{M}}(\bm{q}[k]) (\ddot{\bm{q}}_{ref}[k+1] - \hat{\bm{d}}) + \tilde{\bm{h}}(\bm{q}[k], \hat{\dot{\bm{q}}}[k])$\;
    
    \BlankLine
    \BlankLine
    \BlankLine
    \tcc{--- Torque limit ---}
    $\bm{\tau}_{cmd}[k+1] \gets \mathrm{sat}(\bm{\tau}_{cmd}[k+1],\bm{\tau}_{limit})$  \tcp*[r]{element-wise symmetric saturation $\left(|\tau_i| \leq \tau_{limit,i}\right)$}
    $\ddot{\bm{q}}_{ref}[k+1] \gets \tilde{\bm{M}}(\bm{q}[k])^{-1} \left ( \bm{\tau}_{cmd}[k+1] - \tilde{\bm{h}}(\bm{q}[k], \hat{\dot{\bm{q}}}[k]) \right) + \hat{\bm{d}}[k]$  \tcp*[r]{Torque limit for observer}
    
    \BlankLine
    \BlankLine
    \BlankLine
    $\mathrm{writeTorque}(\bm{\tau}_{cmd}[k+1])$ \tcp*[r]{Send torque command to motors}

    $k \gets k+1$\;
    $\mathrm{sleep}()$ \tcp*[r]{Periodic execution}
}
\end{algorithm*}

\section*{Acknowledgment}
This work was supported by JSPS KAKENHI Grant Number 24K00905, 24KJ0503, JST PRESTO Grant Number JPMJPR24T3, and JST ALCA-Next Japan, Grant Number JPMJAN24F1.
This study was based on the results obtained from the JPNP20004 project subsidized by the New Energy and Industrial Technology Development Organization (NEDO).
\bibliographystyle{IEEEtran}
\bibliography{ref}

\begin{IEEEbiography}[{\includegraphics[width=1in,height=1.25in,clip,keepaspectratio]{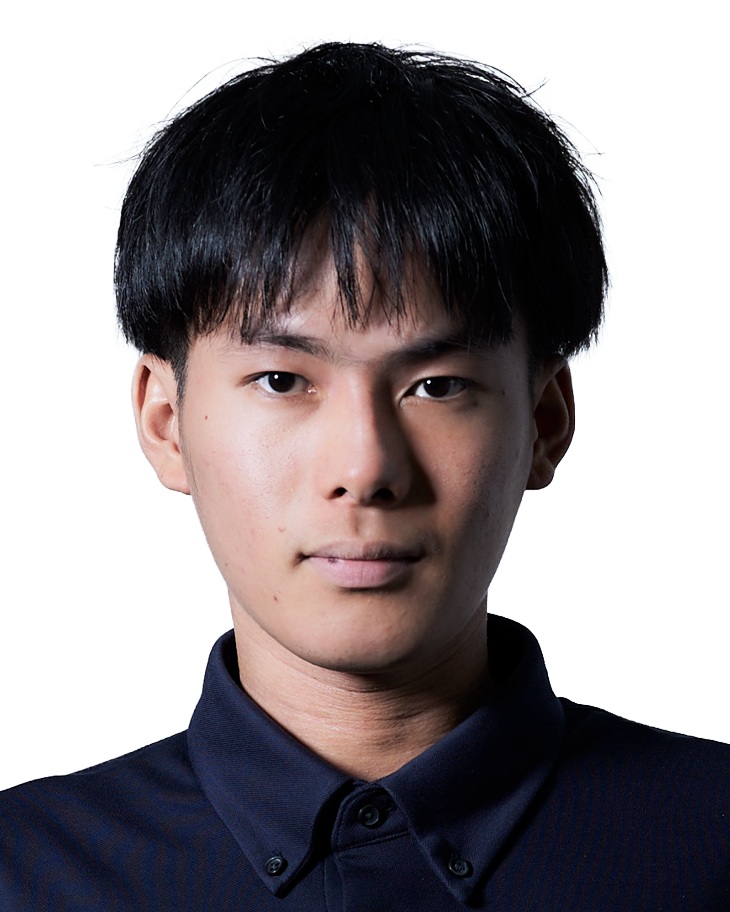}}]{Koki Yamane} 
received the B.E. degree in engineering systems and the M.E. degree in intelligent and mechanical interaction systems from the University of Tsukuba, Tsukuba, Japan, in 2022 and 2024, respectively.
He is currently pursuing the Ph.D. degree in intelligent and mechanical interaction systems at the Graduate School of Science and Technology,  University of Tsukuba.
From October 2025 to March 2026, he was a Visiting Researcher with the Institute of Cognitive Systems, Technical University of Munich, Munich, Germany.
He has also gained industry research experience with Preferred Networks, Inc., and OMRON SINIC X Corporation.
His research interests include force control, bilateral teleoperation, and imitation learning.
He received the Dean’s Award and the Outstanding Master’s Thesis Award from the University of Tsukuba in 2024.
He was awarded a JSPS Research Fellowship for Young Scientists (DC1).
\end{IEEEbiography}

\begin{IEEEbiography}[{\includegraphics[width=1in,height=1.25in,clip,keepaspectratio]{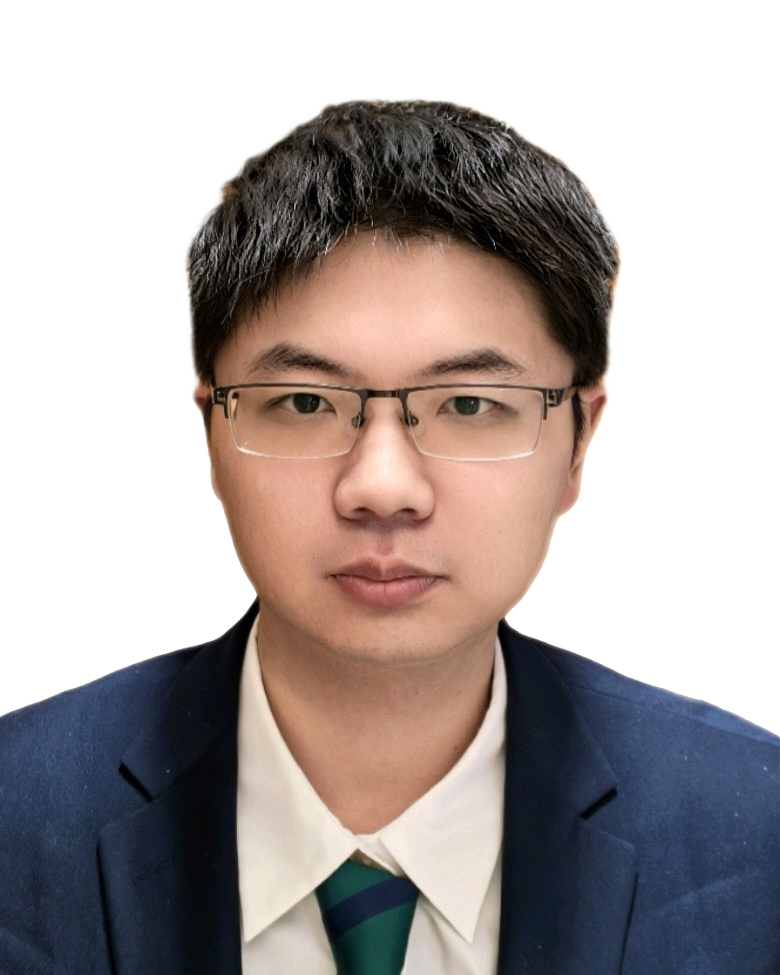}}]
{Yunhan Li} received the B.E. degree in electronic information engineering from SWJTU-Leeds Joint School, Southwest Jiaotong University, Chengdu, China, in 2022.
He received the M.E. degree in intelligent and mechanical interaction systems from the  University of Tsukuba, Tsukuba, Japan, in 2026.
He is currently pursuing the Ph.D. degree in intelligent and mechanical interaction systems at the Graduate School of Science and Technology,  University of Tsukuba.
His research interests include robot learning, imitation learning, and motion control.
He received the IEEJ Industry Applications Society Excellent Presentation Award from the Institute of Electrical Engineers of Japan in 2025.
\end{IEEEbiography}

\begin{IEEEbiography}[{\includegraphics[width=1in,height=1.25in,clip,keepaspectratio]{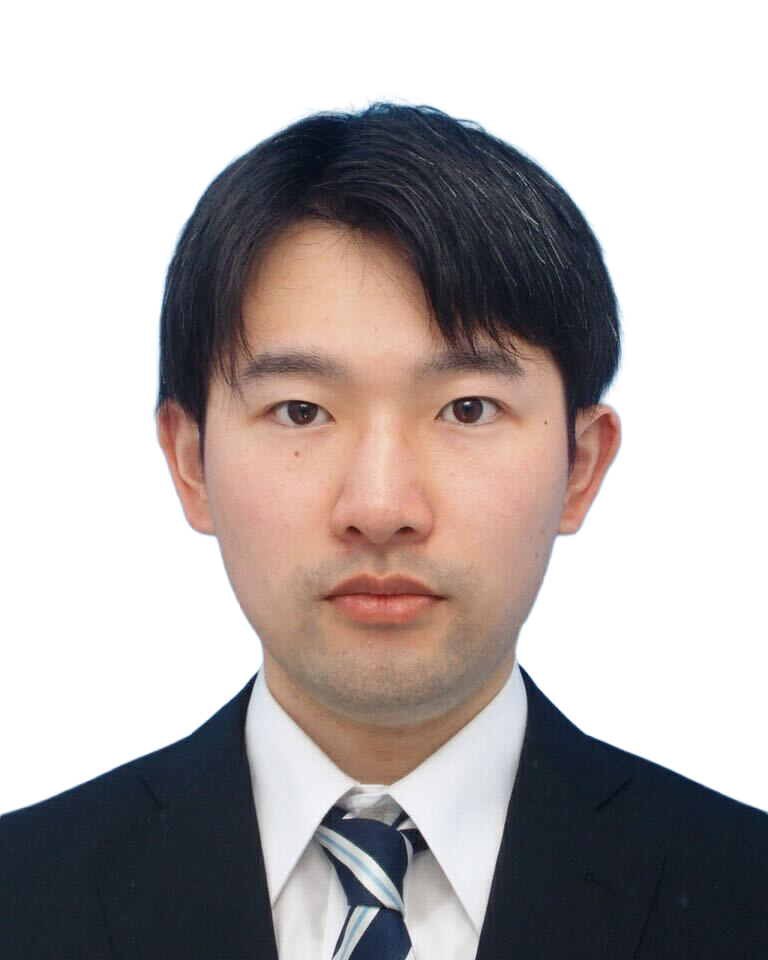}}]{Masashi Konosu} received the B.E. degree in engineering systems and the M.E. degree in intelligent and mechanical interaction systems from the University of Tsukuba, Tsukuba, Japan, in 2024 and 2026, respectively.
His research interests include motion control, force control, and imitation learning.
He received the Outstanding Master’s Thesis Award from the University of Tsukuba in 2026.
\end{IEEEbiography}

\begin{IEEEbiography}[{\includegraphics[width=1in,height=1.25in,clip,keepaspectratio]{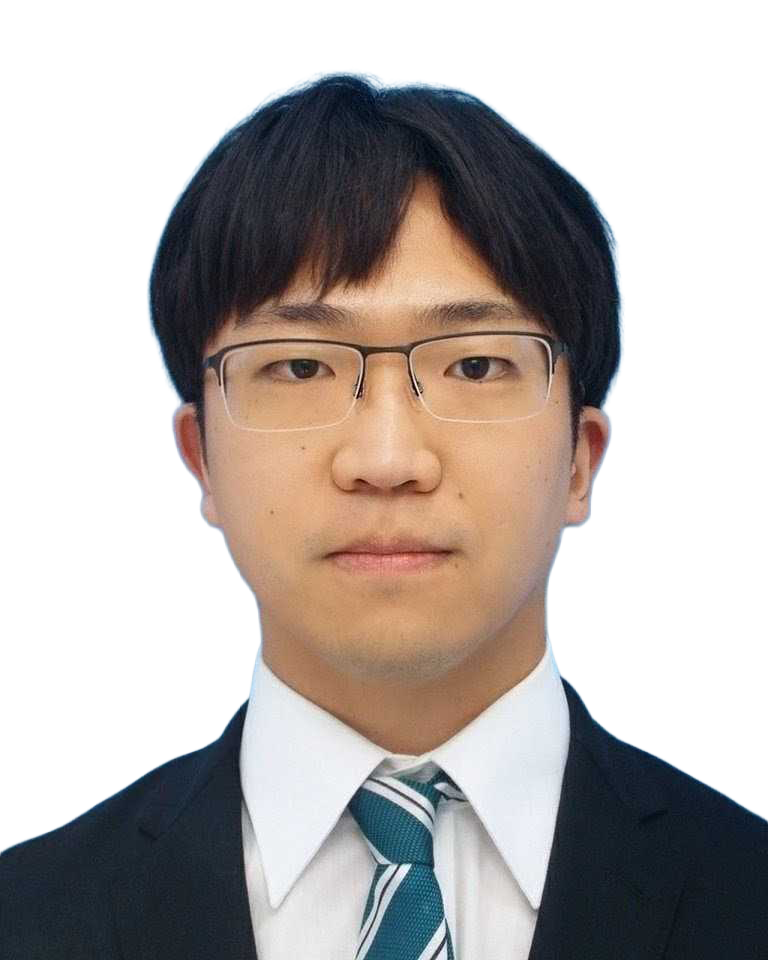}}]{Koki Inami} 
received the B.E. degree in policy and planning sciences and the M.E. degree in intelligent and mechanical interaction systems from the University of Tsukuba, Tsukuba, Japan, in 2024 and 2026, respectively.
He is currently pursuing the Ph.D. degree in intelligent and mechanical interaction systems at the Graduate School of Science and Technology, University of Tsukuba.
He has also been involved in industry research on reinforcement learning for humanoid robots.
His research interests include motion control, force control, and imitation learning.
He received the Dean’s Award and the Outstanding Master’s Thesis Award from the University of Tsukuba in 2026.
He also received the Japan Society of Mechanical Engineers Miura Award in 2026.
He was awarded a JSPS Research Fellowship for Young Scientists (DC1).
\end{IEEEbiography}

\begin{IEEEbiography}[{\includegraphics[width=1in,height=1.25in,clip,keepaspectratio]{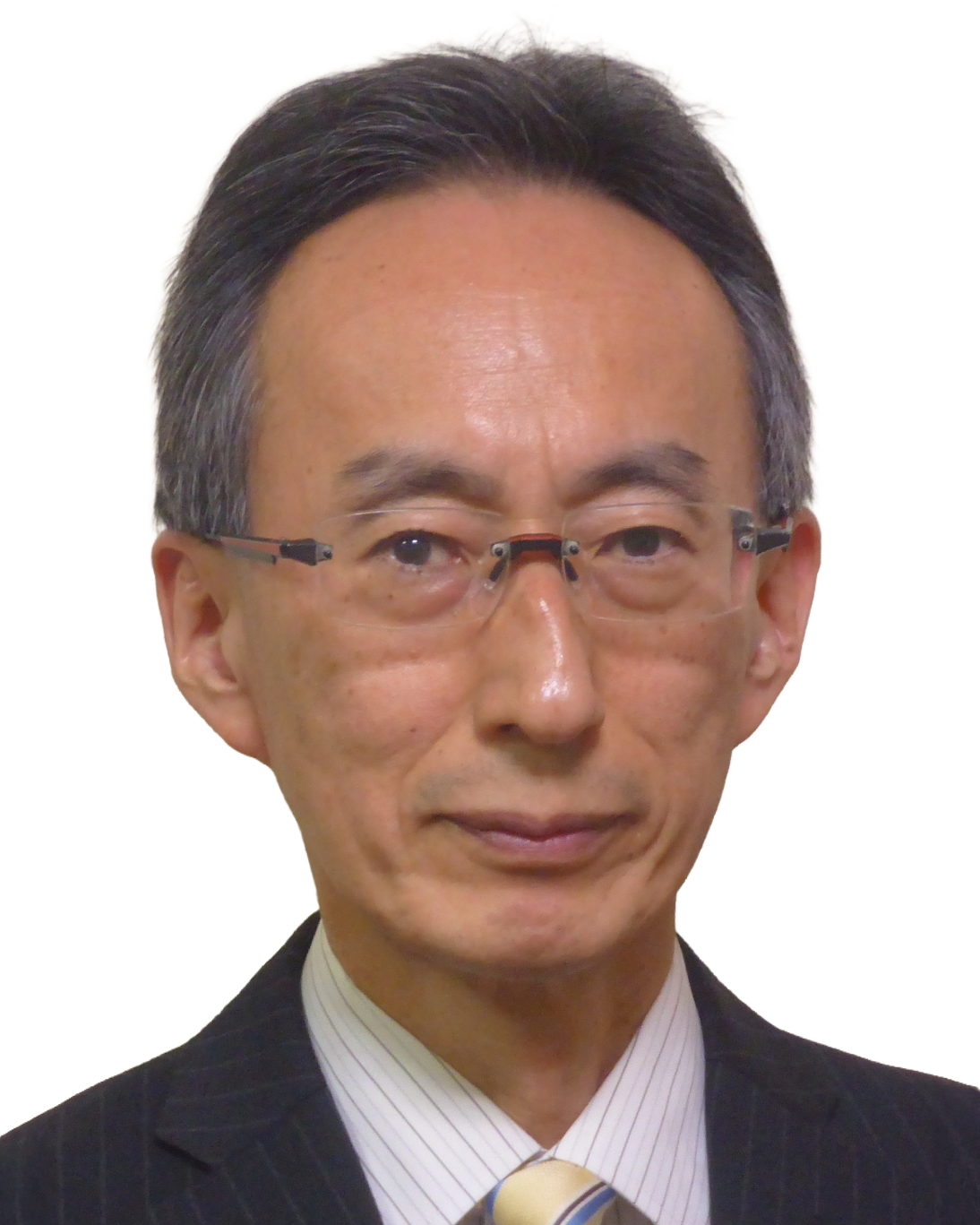}}]{Junji Oaki} received the B.E. and M.E. degrees in electrical engineering from Chiba University, Chiba, Japan, in 1983 and 1985, respectively, and the Ph.D. degree in fundamental science and technology from Keio University, Yokohama, Japan, in 2010. 
From 1985 to 2024, he was with the Corporate Research and Development Center, Toshiba Corporation, Kawasaki, Japan. 
Since 2024, he has been a full-time researcher at the University of Tsukuba, Tsukuba, Japan. 
His research interests include system identification, motion control, and real-time control systems. 
He received the SICE Control Division Technology Award in 2012, the IEEJ Industry Applications Society Distinguished Transaction Paper Award in 2016, and the METI Minister's Award for Excellence at the World Robot Summit 2018. 
He is a Fellow of the Robotics Society of Japan (RSJ). 
\end{IEEEbiography}

\begin{IEEEbiography}[{\includegraphics[width=1in,height=1.25in,clip,keepaspectratio]{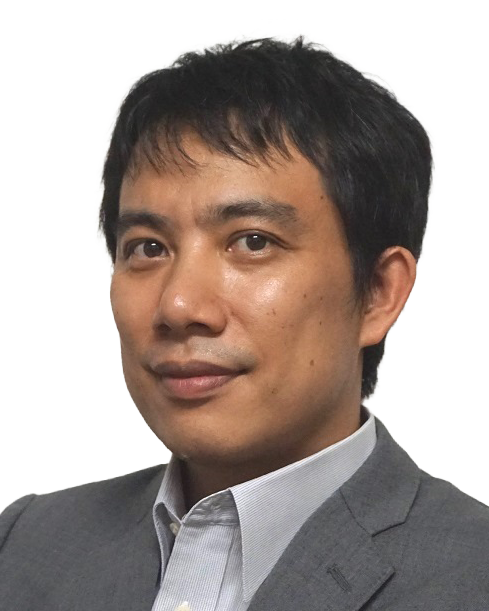}}]{Toshiaki Tsuji} received the B.E. degree in system design engineering and the M.E. and Ph.D. degrees in integrated design engineering from Keio University, Yokohama, Japan, in 2001, 2003, and 2006, respectively. He was a research associate with the Department of Mechanical Engineering, Tokyo University of Science, from 2006 to 2007. He is currently an associate professor with the Department of Electrical and Electronic Systems, Saitama University, Saitama, Japan. His research interests include motion control, haptics, and rehabilitation robots. He received the FANUC FA and Robot Foundation Original Paper Award in 2007 and 2008, respectively. He also received the RSJ Advanced Robotics Excellent Paper Award and the IEEJ Industry Application Society Distinguished Transaction Paper Award in 2020.
\end{IEEEbiography}

\begin{IEEEbiography}[{\includegraphics[width=1in,height=1.25in,clip,keepaspectratio]{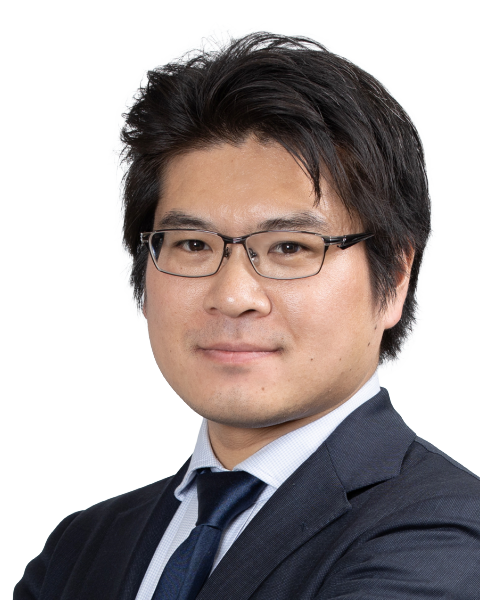}}]{Sho Sakaino} received the B.E. degree in system design engineering and the M.E. and Ph.D. degrees in integrated design engineering from Keio University, Yokohama, Japan, in 2006, 2008, and 2011, respectively.
He was an assistant professor at Saitama University between 2011 and 2019.
Since 2019, he has been an associate professor at the University of Tsukuba.
His research interests include mechatronics, motion control, robotics, and haptics.
He received the IEEE IES Best Conference Paper Award in 2022.
He also received the IEEJ Industry Application Society Distinguished Transaction Paper Award in 2011 and 2020 and the RSJ Advanced Robotics Excellent Paper Award in 2020.
\end{IEEEbiography}

\EOD

\end{document}